%% file: visexmem_iclr2027.tex
\RequirePackage{fix-cm}
\documentclass{article}

\usepackage{iclr2027_conference,times}

\input{math_commands.tex}

\usepackage{amsmath}

\usepackage{graphicx}
\usepackage{svg}
\usepackage{xspace}

\usepackage[T1]{fontenc}

\usepackage{booktabs}
\usepackage{multirow}
\usepackage{array}
\usepackage{xltabular}
\usepackage{threeparttable}
\usepackage{ragged2e}
\usepackage{float}
\usepackage{fancyvrb}

\usepackage{xcolor}
\usepackage{colortbl}

\definecolor{PromptBlue}{HTML}{123F79}
\definecolor{PromptBG}{HTML}{F4F6FB}

\definecolor{VQABG}{RGB}{248,236,246}
\definecolor{GPTBG}{RGB}{252,247,232}

\usepackage{tikz}
\usetikzlibrary{calc,shadows,positioning,arrows.meta,fit}

\usepackage{fvextra}

\usepackage{caption}

\usepackage{enumitem}

\usepackage[
    disable,
    textsize=tiny,
    colorinlistoftodos
]{todonotes}

\usepackage[most]{tcolorbox}

\usepackage{tikz}
\usetikzlibrary{positioning,arrows.meta,fit,calc}
\usepackage{tabularx}
\usepackage{url}
\usepackage{hyperref}

\newcommand{\visexmem}{\textsc{VisExMEM}\xspace}

\title{
\texorpdfstring{
\raggedright
\raisebox{-0.16\height}{%
  \includegraphics[
    height=0.70cm,
    trim=35 25 45 25,
    clip
  ]{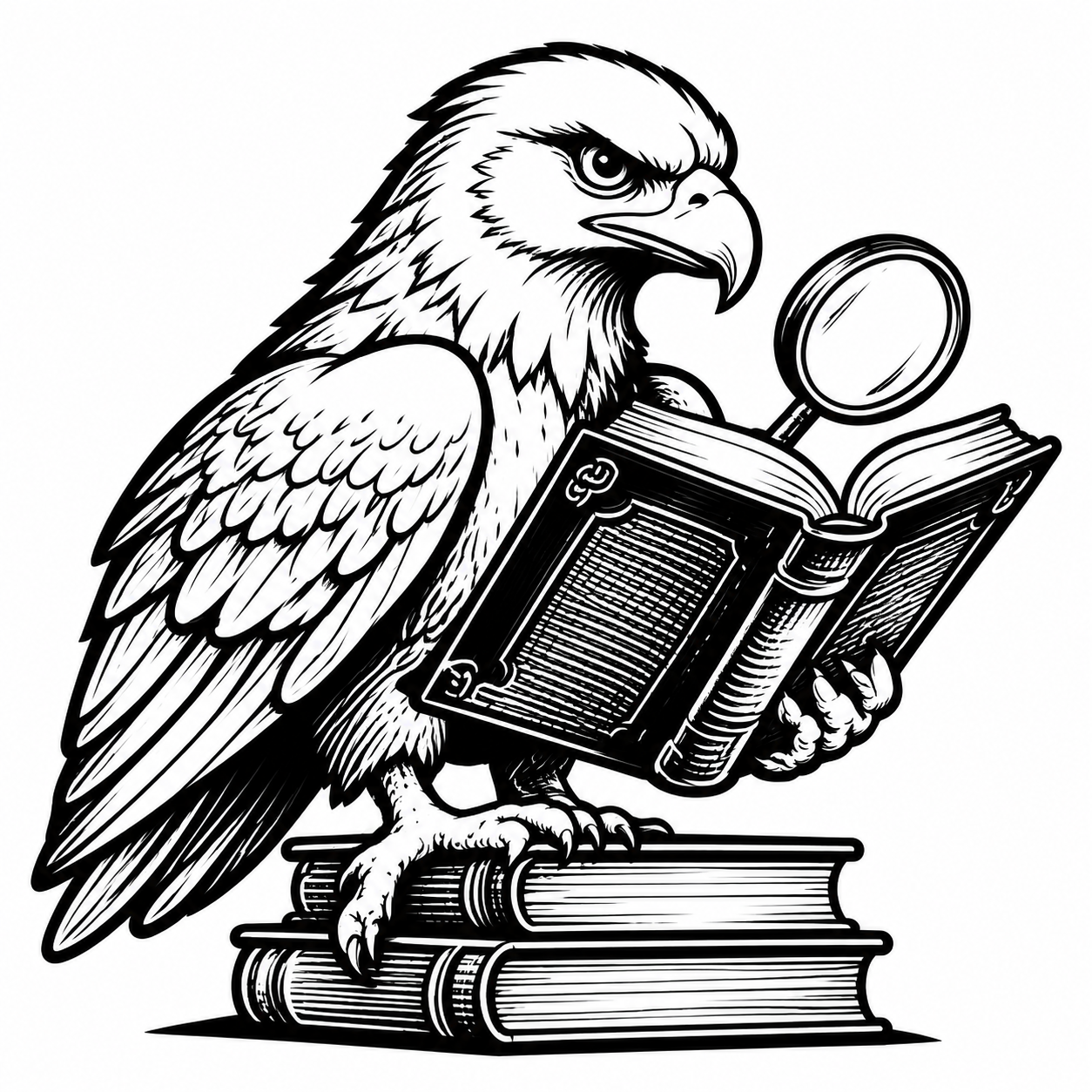}%
}%
\hspace{0.5em}%
\textsc{VisExMEM}: Visual Evidence-Based Explainable Multimodal Evaluation Metric
}{
VisExMEM: Visual Evidence-Based Explainable Multimodal Evaluation Metric
}
}

\author{
Subhadeep Roy, Christoph Leiter, Steffen Eger \\
{\hypersetup{urlbordercolor={1 1 1}}%
\href{https://nl2g.github.io/}{Natural Language Learning \& Generation Lab}}, University of Technology Nuremberg \\
\texttt{\{subhadeep.roy, christoph.leiter, steffen.eger\}@utn.de}
}

\iclrfinalcopy 

\begin{document}

\maketitle
\lhead{Preprint}

\input{strucuture/0_abstract}
\input{strucuture/1_intro}
\input{strucuture/2_related_work}
\input{strucuture/3_visexmem}
\input{strucuture/4_agita_bench}
\input{strucuture/5_experiments}
\input{strucuture/6_results_analysis}
\input{strucuture/7_conclusion}
\input{strucuture/ai_ethich}

\bibliography{iclr2027_conference}
\bibliographystyle{iclr2027_conference}

\appendix
\section{Appendix}
\input{strucuture/appendix}

\end{document}

%% file: math_commands.tex
\usepackage{amsmath,amsfonts,bm}

\def\eqref#1{equation~\ref{#1}}

\def\1{\bm{1}}

\DeclareMathAlphabet{\mathsfit}{\encodingdefault}{\sfdefault}{m}{sl}
\SetMathAlphabet{\mathsfit}{bold}{\encodingdefault}{\sfdefault}{bx}{n}



%% file: strucuture/0_abstract.tex
\begin{abstract}
Current image--text evaluation often relies on a global score, which gives
limited insight into which semantic details are supported by the image.
Explainable metrics provide rationales, but the visual evidence supporting
those rationales can still be difficult to verify. We introduce \textsc{VisExMEM}, a reference-free evaluator that decomposes captions into
atomic claims, localizes visual evidence, verifies each claim, and averages
claim-level support into a global score. Explanations use the same evidence
and decisions. We train Qwen3.5-based 4B and 9B models on 100K samples with
multi-task SFT, counterfactual refinement, and evidence-aware GDPO. Across eight benchmarks, \textsc{VisExMEM}-9B achieves the highest descriptive composite among open-source evaluators in our comparison, exceeding VQAScore by 11.7 points while remaining competitive with GPT-5.6 Sol. We also introduce \textsc{AGITA-Bench}, containing 600 human-annotated image--text pairs and 8.7K atomic claims for
global and claim-level evaluation. For untuned Qwen3.5 and GPT-5.6 Sol controls, decomposed prompting raises the same composite by 3.6--5.1 points relative to holistic prompting. Human evaluation shows that \textsc{VisExMEM} explanations outperform prior explainable metrics and approach GPT-5.6 Sol. \footnote{{\tiny VisExMEM and
AGITA-Bench will be released at: \url{https://github.com/NL2G/VisExMEM}.}}
\end{abstract}

%% file: strucuture/1_intro.tex
\section{Introduction}

The rapid progress of multimodal generative models has made reliable
image--text evaluation increasingly important for benchmarking and model
development
\citep{hartwig2025surveyqualitymetricstexttoimage,
berger2025surveyinglandscapeimagecaptioning,
sarto2025imagecaptioningevaluationage}.
Automatic evaluation has progressed from global cross-modal similarity
\citep{hessel2022clipscorereferencefreeevaluationmetric}
to learned preference models
\citep{kirstain2023pickapicopendatasetuser,xu2023imagereward,
wu2023humanpreferencescorev2}, structured VQA-based evaluation
\citep{hu2023tifaaccurateinterpretabletexttoimage,
cho2024davidsonianscenegraphimproving,
lin2024evaluatingtexttovisualgenerationimagetotext},
VLM-as-a-judge approaches
\citep{chen2024mjbenchmultimodalrewardmodel,
pu2025judgeanythingmllmjudge,
li2026ksortevalefficientpreference},
and structured alignment functions
\citep{chen2026hyperalignhyperbolicentailmentcones,
koyama2026rigelselfdistilledscoreadaptation}.
However, a high global score can still hide a local semantic error. An image
may look broadly correct while getting a count, attribute binding, action,
spatial relation, or subject--object role wrong. Benchmarks such as
Winoground, TIFA, GenAI-Bench, MMIG-Bench, and ImagenWorld are designed to
test these fine-grained and compositional errors
\citep{thrush2022winogroundprobingvisionlanguage,
hu2023tifaaccurateinterpretabletexttoimage,
li2024genaibenchevaluatingimprovingcompositional,
hua2025mmigbenchcomprehensiveexplainableevaluation,
sani2026imagenworldstresstestingimagegeneration}.

Strong evaluation performance also does not make a metric easy to inspect.
VIEScore and FLEUR provide natural-language rationales
\citep{ku2024viescoreexplainablemetricsconditional,
lee2024fleurexplainablereferencefreeevaluation}, while recent work adds
criterion-level explanations
\citep{kim2025expertexplainableimagecaptioning},
multi-criteria aggregation
\citep{ohi2026multimodalmultitaskmulticriteriaautomatic},
and element-level grounding
\citep{shi2026revealerreinforcementguidedvisualreasoning}.
A remaining issue is whether the explanation reflects the visual evidence and
the decisions that produced the score. A fluent explanation can still sound
reasonable even when it is based on the wrong evidence or decision.
Similarly, global score correlation does not show whether a metric responds
to the specific visual or linguistic constraint that is wrong
\citep{ross2024makesgoodmetricevaluating}.

\begin{figure}[t]
    \centering
    \setlength{\abovecaptionskip}{0pt}
    \setlength{\belowcaptionskip}{-3pt}
    \includegraphics[width=\textwidth]{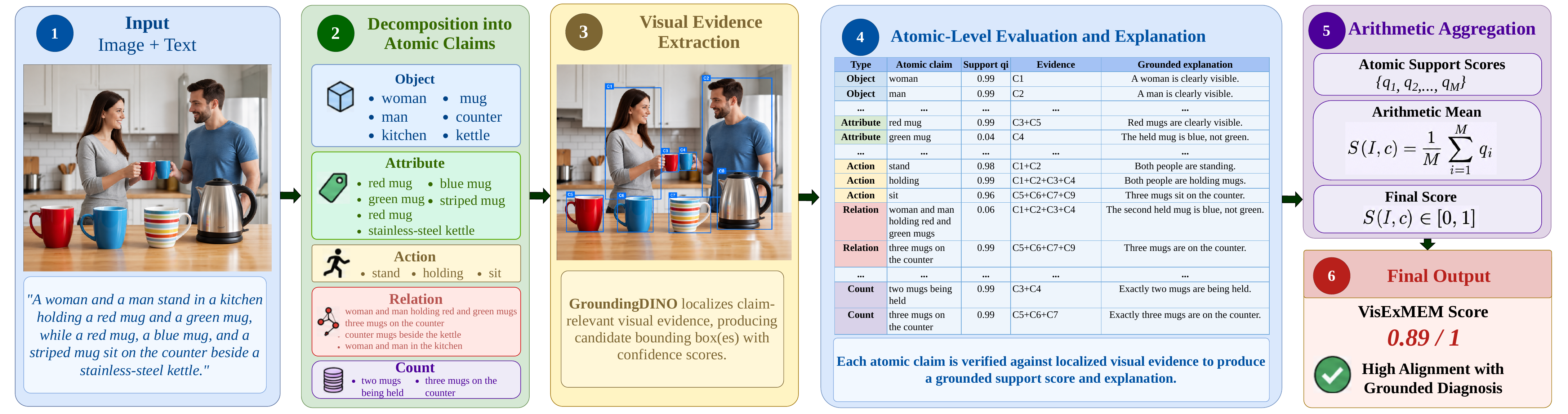}
    \caption{\textbf{Overview of VisExMEM.}
    VisExMEM decomposes captions into atomic claims, localizes visual evidence
    for each claim, computes claim support scores, and averages them into a global score. Explanations are generated afterward from the same evidence and fixed claim decisions.}
    \label{fig:visexmem_overview}
    \vspace{-2mm}
\end{figure}

We introduce \textsc{VisExMEM}
(\emph{\textbf{Vis}ual Evidence-Based \textbf{Ex}plainable
\textbf{M}ultimodal \textbf{E}valuation \textbf{M}etric}), a reference-free
metric that decomposes text into atomic claims and localizes visual evidence
(Figure~\ref{fig:visexmem_overview}). For each claim, the localized evidence
is used to compute a Yes/No log-probability margin, which is mapped to a claim
support score. The arithmetic mean of these scores gives the global score.
Explanations are generated afterward from the same evidence and fixed claim
decisions, so they do not determine the score. We train Qwen3.5-based~\citep{qwen35blog}
4B and 9B models in three stages (Figure~\ref{fig:visexmem_training_overview}).
Multi-task SFT establishes the evaluation pipeline and its structured outputs.
Counterfactual refinement uses minimal semantic changes so the model responds
to a single changed constraint while keeping the remaining content fixed.
Finally, evidence-aware GDPO~\citep{liu2026gdpogrouprewarddecouplednormalization}
improves semantic verification and encourages predictions to rely on localized visual evidence.

To evaluate both global and claim-level agreement with humans, we introduce
\textsc{AGITA-Bench}
(\emph{\textbf{A}tomic and \textbf{G}lobal
\textbf{I}mage--\textbf{T}ext \textbf{A}lignment Benchmark}), a 600-case
benchmark with human annotations covering image-to-text (I2T) and
text-to-image (T2I) evaluation. In I2T, captions are evaluated against real
images. In T2I, generation prompts are evaluated against generated images.
Each case belongs to an easy, medium, or hard compositional setting.
Annotators provide an overall image--text consistency rating and claim-level
support judgments. This allows us to evaluate both global and claim-level
scores. Our contributions are:

\begin{itemize}[
    leftmargin=2.5em,
    labelsep=0.5em,
    itemsep=4pt,
    topsep=2pt,
    parsep=0pt,
    partopsep=0pt
]

    \item \textbf{Evidence-Grounded Explainable Evaluation:}
    We introduce \textsc{VisExMEM}, which links atomic claim scores to
    localized visual evidence and generates explanations from the same
    evidence and claim decisions.

    \item \textbf{Progressive Multimodal Verification Training:}
    We develop a three-stage training strategy for learning structured
    evaluation, sensitivity to minimal semantic changes, and dependence on
    visual evidence.

    \item \textbf{AGITA-Bench:}
    We provide a benchmark that evaluates global agreement and individual
    claim decisions in both I2T and T2I settings across different levels of
    compositional difficulty.
\end{itemize}


%% file: strucuture/2_related_work.tex
\section{Related Work}
\label{sec:related_work}

\paragraph{Image--text alignment evaluation.}
Reference-free metrics include embedding-based methods that score an
image--text pair as a whole, without decomposing it into smaller semantic units
\citep{hessel2022clipscorereferencefreeevaluationmetric,li2023blip2}
and learned human-preference models
\citep{kirstain2023pickapicopendatasetuser,xu2023imagereward,
wu2023humanpreferencescorev2,zhang2024learningmultidimensionalhumanpreference}.
More recent methods break the evaluation into smaller semantic units.
TIFA~\citep{hu2023tifaaccurateinterpretabletexttoimage} and
DSG~\citep{cho2024davidsonianscenegraphimproving} decompose prompts into
verifiable questions,
VQAScore~\citep{lin2024evaluatingtexttovisualgenerationimagetotext}
uses generative VQA likelihood, and T2I-FineEval
\citep{hosseini2025t2ifineeval} decomposes both text and image.
Recent approaches also explore hyperbolic alignment
\citep{chen2026hyperalignhyperbolicentailmentcones} and task-adapted caption scoring~\citep{koyama2026rigelselfdistilledscoreadaptation}.
\textsc{VisExMEM} follows this fine-grained setting by decomposing the caption into atomic claims for evaluation.

\paragraph{Explainable and evidence-grounded evaluation.}
FLEUR~\citep{lee2024fleurexplainablereferencefreeevaluation} and
VIEScore~\citep{ku2024viescoreexplainablemetricsconditional} pair scores with
natural-language rationales, while EXPERT
\citep{kim2025expertexplainableimagecaptioning} and HarmonicEval
\citep{ohi2026multimodalmultitaskmulticriteriaautomatic} expose structured
criteria. Grounding-oriented work includes
GLaMM~\citep{rasheed2024glammpixelgroundinglarge},
Grounding-IQA~\citep{chen2026groundingiqagroundingmultimodallanguage}, and
multimodal-rationale VQA~\citep{li2025multimodalrationalesexplainablevisual}.
Most closely, REVEALER~\citep{shi2026revealerreinforcementguidedvisualreasoning}
performs element-level alignment evaluation through a
grounding--reasoning--scoring trajectory. \textsc{VisExMEM} instead uses localized visual evidence to score each atomic claim. These claim scores determine the global score, while the explanation is generated afterward from the same evidence and fixed claim decisions.

\paragraph{Diagnostic benchmarks and metric meta-evaluation.}
Benchmarks test different aspects of image--text evaluation.
Winoground~\citep{thrush2022winogroundprobingvisionlanguage} and
EqBen~\citep{wang2023equivariantsimilarityvisionlanguagefoundation}
probe compositional sensitivity; TIFA
\citep{hu2023tifaaccurateinterpretabletexttoimage},
GenAI-Bench~\citep{li2024genaibenchevaluatingimprovingcompositional},
T2I-CompBench++~\citep{huang2025t2icompbenchenhancedcomprehensivebenchmark}, and
CountBench~\citep{paiss2023teachingclipcount} assess fine-grained
faithfulness, compositionality, and counting, while
Pick-a-Pic~\citep{kirstain2023pickapicopendatasetuser} provides human
preference comparisons. Broader diagnostic resources include EvalMuse-40K
\citep{han2024evalmuse40kreliablefinegrainedbenchmark},
MMIG-Bench~\citep{hua2025mmigbenchcomprehensiveexplainableevaluation},
FineGRAIN~\citep{hayes2025finegrain}, and
ImagenWorld~\citep{sani2026imagenworldstresstestingimagegeneration}.
For metric meta-evaluation, \textsc{ProtoBias}
\citep{roy2026prototypicalitybiasrevealsblindspots} tests prototypicality
bias, T2IScoreScore~\citep{saxon2024evaluatesevaluationsobjectivelyscoring}
evaluates ordering and separation over semantic error graphs, and
CROC~\citep{leiter2026crocevaluatingtrainingt2i} uses pseudo- and
human-labeled contrastive robustness checks. \textsc{AGITA-Bench} complements
these benchmarks by pairing a global rating with claim-level support
judgments for the same image--text case, allowing overall agreement and
individual claim decisions to be evaluated together.

%% file: strucuture/3_visexmem.tex
\section{VisExMEM}
\label{sec:method}

\textsc{VisExMEM} is a reference-free image--caption evaluation metric that
grounds claim-level verification in localized visual evidence. In this section, we describe
its scoring formulation, training procedure, and grounded explanation mechanism.

\subsection{Evidence-Grounded Verification and Scoring}
\label{subsec:visexmem_inference}

Given image $I$ and caption $c$, \textsc{VisExMEM} decomposes $c$ into
$M$ typed atomic claims
$\mathcal{A}(c)=\{(a_i,t_i)\}_{i=1}^{M}$, where $a_i$ is the claim and $t_i$
its type (\emph{object, attribute, action, relation, count or negation}).
GroundingDINO~\citep{liu2024groundingdinomarryingdino} localizes
claim-relevant regions as visual evidence. As shown in
Figure~\ref{fig:visexmem_overview}, the model evaluates each claim using this
evidence with the full image to retain scene context.

For each claim, the model assigns mean token log-probabilities $s_i^{Y}$ and
$s_i^{N}$ to \texttt{yes} and \texttt{no}. We define
$m_i=s_i^{Y}-s_i^{N}$ and $q_i=\sigma(m_i)$, where
$\sigma(x)=1/(1+e^{-x})$. Applying sigmoid after subtraction maps the Yes/No
margin to $[0,1]$; applying it first would compare compressed
scores. The global score is $S(I,c)=M^{-1}\sum_{i=1}^{M}q_i$.
We retain each claim's type, support score, localized box(es), grounding
confidence, and decision. Explanations are generated afterward from these
fixed records and cannot alter the score.

\subsection{Training}
\label{subsec:training}

\begin{figure}[t]
    \centering
    \setlength{\abovecaptionskip}{0pt}
    \setlength{\belowcaptionskip}{-3pt}
    \includegraphics[width=\linewidth]{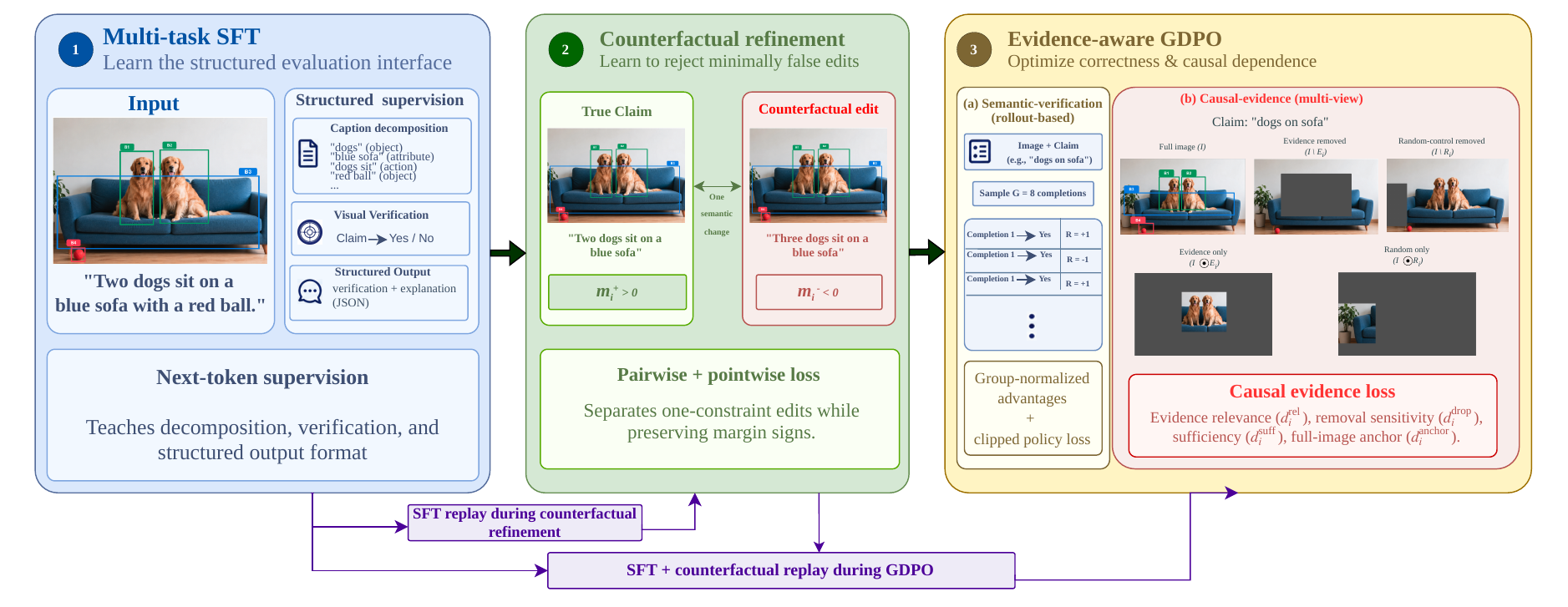}
    \caption{\textbf{VisExMEM training pipeline.}
    Multi-task SFT learns the structured evaluation tasks, counterfactual
    refinement targets semantic changes, and evidence-aware GDPO improves
    verification and evidence dependence. Earlier supervision is replayed
    in later stages.}
    \label{fig:visexmem_training_overview}
    \vspace{-2mm}
\end{figure}

We train Qwen3.5-4B and Qwen3.5-9B~\citep{qwen35blog}
\textsc{VisExMEM} models with our novel three-stage training procedure
(Figure~\ref{fig:visexmem_training_overview}):
(1) multi-task SFT directly supervises decomposition, claim verification,
evidence selection, and explanation,
(2) counterfactual refinement teaches the model to distinguish minimally
incorrect semantic edits from supported claims, and
(3) evidence-aware GDPO~\citep{liu2026gdpogrouprewarddecouplednormalization}
optimizes claim verification to depend on localized visual evidence.

\paragraph{Training data.}
Training data come from MS COCO~\citep{lin2014coco}, Visual
Genome~\citep{krishna2017visualgenome}, VSR~\citep{liu2023vsr},
Flickr30k Entities~\citep{plummer2015flickr30kentities}, Open
Images~\citep{kuznetsova2020openimages} with Localized
Narratives~\citep{ponttuset2020localizednarratives},
DOCCI~\citep{onoe2024docci}, PixMo-Count~\citep{deitke2024molmo},
CROC$_{\mathrm{syn}}$~\citep{leiter2026crocevaluatingtrainingt2i}, and the
training split of EvalMuse-40K~\citep{han2024evalmuse40kreliablefinegrainedbenchmark}. We convert the text into typed atomic claims,
using native annotations and human judgments when available and an offline
Qwen3.6-27B teacher for missing decomposition or structure; uncertain outputs
are filtered out. GroundingDINO then localizes candidate boxes as visual
evidence cues, which are provided as evidence rather than SFT targets. We use
boxes instead of pixel masks because they localize the relevant region without
revealing its exact silhouette, reducing mask-shape shortcuts. We construct 40K SFT examples, a 20K counterfactual source pool, and 40K Stage~3 examples equally split between semantic-verification and causal-evidence supervision (see Appendix~Figure~\ref{fig:processed_training_examples}).
Counterfactual data combine CROC$_{\mathrm{syn}}$ image pairs with text-side
minimal pairs that change one verified semantic constraint while keeping the
image fixed. Explanations are constructed only after claims, decisions, and
evidence are fixed. Full construction, filtering, source composition, and
replay schedules are in Appendix~\ref{app:training-data}.

\paragraph{Stage 1: Multi-task SFT.}
Stage~1 teaches the structured evaluation tasks used by
\textsc{VisExMEM}. We jointly supervise caption decomposition, claim-level
verification, evidence selection, and explanation generation through
next-token supervision. For image $I$, caption $c$, and structured target
$y=(y_1,\ldots,y_T)$, we minimize
\begin{equation}
\mathcal{L}_{\mathrm{SFT}}
=
-\mathbb{E}
\sum_{t=1}^{T}
\log \pi_\theta
\left(
y_t \mid y_{<t}, I, c
\right).
\label{eq:sft}
\end{equation}
where $\pi_\theta$ is the trainable model. Candidate boxes are supplied as
visual evidence; evidence selection predicts candidate IDs rather than box
coordinates. Representative processed records are shown in
Appendix~Figure~\ref{fig:processed_training_examples}; exact target formats,
task mixtures, and optimization settings are given in
Appendix~\ref{app:sft_details}.

\paragraph{Stage 2: Counterfactual refinement.}
Stage~2 makes claim verification sensitive to controlled semantic changes.
Each supported image--claim pair is paired with a contradicted counterpart.
Text-side pairs change one constraint while keeping the image fixed, e.g.,
``two dogs'' to ``three dogs,'' while image-side pairs provide the
corresponding visual contrast. Let
\begin{equation}
m_\theta(I,a)
=
\ell_\theta(\texttt{yes}\mid I,a)
-
\ell_\theta(\texttt{no}\mid I,a),
\label{eq:verification_margin}
\end{equation}
where $\ell_\theta$ denotes the decision-token logit under model parameters
$\theta$. Training requires a positive margin for the supported example, a
negative margin for the counterfactual, and their difference to exceed a
target margin. This is supervised margin training, not reinforcement
learning. Stage~1 decomposition and verification examples are replayed to
retain earlier supervision~\citep{luo2026rlforgetscontinualpolicy}; the full
loss, filtering rules, and replay settings are given in
Appendix~\ref{app:cf_details}.

\paragraph{Stage 3: Evidence-aware GDPO.}
Stage~3 improves semantic verification while making predictions depend on
localized visual evidence. Building on GRPO-style group-relative optimization
\citep{shao2024deepseekmathpushinglimitsmathematical}, we use an
evidence-aware GDPO variant inspired by
GDPO~\citep{liu2026gdpogrouprewarddecouplednormalization}: semantic
correctness provides the group-relative policy signal, while evidence
dependence is optimized through a separate causal objective. For semantic
input $x_i$, we sample $G$ completions
$o_{i,g}\sim\pi_{\theta_{\mathrm{old}}}(\cdot\mid x_i)$, where $g$ indexes
rollouts and $\pi_{\theta_{\mathrm{old}}}$ is the rollout policy. For semantic
reward $r_{i,g}$, let $\mu_i$ and $\sigma_i$ be the group mean and standard
deviation, and define $h_{i,g,t}=(x_i,o_{i,g,<t})$. The normalized advantage
and policy ratio are
\begin{equation}
A_{i,g}
=
\frac{r_{i,g}-\mu_i}{\sigma_i},
\qquad
\rho_{i,g,t}
=
\frac{\pi_\theta(o_{i,g,t}\mid h_{i,g,t})}
     {\pi_{\theta_{\mathrm{old}}}(o_{i,g,t}\mid h_{i,g,t})}.
\label{eq:gdpo_advantage_ratio}
\end{equation}
Here, $A_{i,g}$ compares each rollout with its group, while
$\rho_{i,g,t}$ measures the token-level change from the rollout policy to the
updated policy. Numerical stabilization and zero-variance handling are given
in Appendix~\ref{app:gdpo_details}. Following the clipped GRPO objective, we
optimize
\begin{equation}
\mathcal{J}_{\mathrm{sem}}
=
\mathbb{E}\!\left[
\frac{1}{G}\sum_{g=1}^{G}\frac{1}{|o_{i,g}|}\sum_t
\left(
\min\!\left\{
\rho_{i,g,t}A_{i,g},
\operatorname{clip}(\rho_{i,g,t},1-\epsilon,1+\epsilon)A_{i,g}
\right\}
-\beta d^{\mathrm{KL}}_{i,g,t}
\right)
\right],
\label{eq:gdpo}
\end{equation}
where $\epsilon$ is the clipping radius, $\beta$ weights KL regularization,
and $d^{\mathrm{KL}}_{i,g,t}$ penalizes deviation from the frozen Stage~2
reference.

Semantic reward alone does not ensure evidence dependence. For each eligible
supported claim, selected evidence $E_i$ is paired with a matched random
region $R_i$, and $s_i(V)$ denotes the verification margin under image view
$V$. We encourage
$s_i(I)-s_i(I\setminus E_i)>
s_i(I)-s_i(I\setminus R_i)$ and
$s_i(I\odot E_i)>s_i(I\odot R_i)$, so selected evidence is more necessary
and sufficient than its control; $I\setminus E$ masks region $E$, while
$I\odot E$ retains only it. Stage~3 minimizes
$\mathcal{L}_{\mathrm{Stage3}}
=-\mathcal{J}_{\mathrm{sem}}
+\lambda_{\mathrm c}\mathcal{L}_{\mathrm{causal}}$, where
$\lambda_{\mathrm c}$ weights the causal loss, using a $7{:}2{:}1$
GDPO:counterfactual:SFT update mixture. Further details are in
Appendix~\ref{app:gdpo_details}; the Verification-GRPO and Causal-GRPO
ablations branch from Stage~2 and are described in Appendix~\ref{app:grpo}.

%% file: strucuture/4_agita_bench.tex
\section{AGITA-Bench}
\label{sec:agita}

\begin{figure}[t]
    \centering
    \setlength{\abovecaptionskip}{0pt}
    \setlength{\belowcaptionskip}{-3pt}
    \includegraphics[width=.8\linewidth]{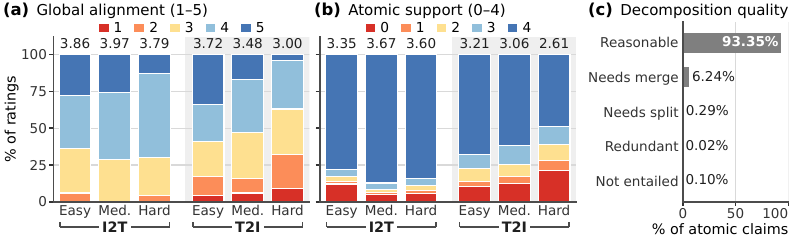}
    \caption{\textbf{AGITA-Bench annotation summary.}
    (a) Global-alignment and (b) atomic-support distributions by direction and
    difficulty; means are shown above bars. (c) Decomposition quality.}
    \label{fig:agita_human_scores}
    \vspace{-2mm}
\end{figure}

To evaluate both global and claim-level agreement with humans, we introduce
\textsc{AGITA-Bench}
(\emph{\textbf{A}tomic and \textbf{G}lobal
\textbf{I}mage--\textbf{T}ext \textbf{A}lignment}), a 600-pair benchmark with
human judgments at both levels. It contains 300 I2T real photographs, with
100 each from ADE20K~\citep{Zhou_2017_CVPR},
HICO-DET~\citep{chao2018learningdetecthumanobjectinteractions}, and
Objects365~\citep{Shao_2019_ICCV}, and 300 T2I pairs from DiffusionDB
(181)~\citep{wang2023diffusiondblargescalepromptgallery} and MJHQ-30K
(119)~\citep{li2024playgroundv25insightsenhancing}. I2T captions cover high,
partial, and low alignment, while T2I uses the original generation prompts.
We balance 200 easy, 200 medium, and 200 hard cases by \emph{semantic
complexity}, not alignment severity. GPT-5.6 Sol decomposes the text into
atomic claims, and fixed rules assign difficulty from claim number and
composition, easy cases contain few simple claims, medium cases add relations
or attribute/count assignments, and hard cases require more constraints or
correct binding of attributes, counts, roles, or spatial relations. This
produces 8,735 claims spanning objects, attributes, actions, relations
(including spatial relations), counts, and negation. Representative examples
are shown in Figure~\ref{fig:agita_qual_examples}; AGITA-Bench is excluded
from training, calibration, and model selection.

Annotators rate atomic support from 0 (contradicted/not supported) to
4 (fully supported) and global alignment from 1 (not aligned) to 5
(fully aligned), with \emph{cannot judge} excluded from atomic statistics.
Decomposition quality is rated separately; \emph{reasonable as written}
means a clear, non-redundant atomic claim requiring neither splitting nor
merging, independent of image support. Mean global alignment is 3.64/5 and
atomic support 3.35/4, while 93.35\% of claims are judged reasonable and
2.07\% receive \emph{cannot judge}. Figure~\ref{fig:agita_human_scores}
shows that I2T ratings remain relatively high across complexity levels,
whereas T2I shifts toward lower global and atomic ratings as complexity
increases. All seven annotators independently rate a shared set of 30 pairs
(460 unique claims), giving mean pairwise quadratic-weighted Cohen's
$\kappa$ of $.602$ for global alignment and $.564$ for atomic support; the
remaining 570 pairs are rated once. Full annotation instructions, agreement
results, annotator details, and the decomposition prompt are given in
Appendix~\ref{app:agita}.

\input{figures/agita_examples_table}

%% file: figures/agita_examples_table.tex

\begin{figure*}[t!]
\centering

\begingroup
\setlength{\fboxsep}{1.2pt}

\newcommand{\agglobal}[2]{%
    \colorbox{blue!6}{%
        \parbox{\dimexpr\linewidth-2\fboxsep\relax}{%
            \centering
            \fontsize{5.4}{5.8}\selectfont
            \textbf{Global H/V:} #1 / #2
        }%
    }%
}

\newcommand{\agclaim}[4]{%
    \noindent
    \textbf{#1} #2
    \textbf{(#3)}:
    \textit{#4}\par
}


\begin{tcolorbox}[
    enhanced,
    width=\textwidth,
    colback=green!2,
    colframe=green!40!black,
    colbacktitle=green!8,
    coltitle=black,
    title=\textbf{Easy $\boldsymbol{\cdot}$ I2T},
    fonttitle=\small,
    boxrule=0.4pt,
    arc=0.8mm,
    boxsep=0.4mm,
    left=0.8mm,
    right=0.8mm,
    top=0.45mm,
    bottom=0.45mm,
    before skip=0pt,
    after skip=0.55mm
]

\begin{minipage}[t]{0.13\linewidth}
\vspace{0pt}
\centering
{\fontsize{5.6}{5.9}\selectfont\textbf{Input}}\par
\vspace{0.25mm}

\includegraphics[
    width=\linewidth,
    height=1.90cm,
    keepaspectratio
]{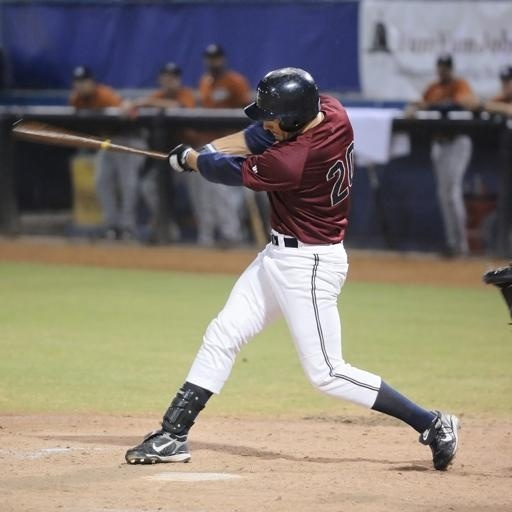}
\end{minipage}
\hfill
%
\begin{minipage}[t]{0.13\linewidth}
\vspace{0pt}
\centering
{\fontsize{5.6}{5.9}\selectfont\textbf{Evidence}}\par
\vspace{0.25mm}

\includegraphics[
    width=\linewidth,
    height=1.90cm,
    keepaspectratio
]{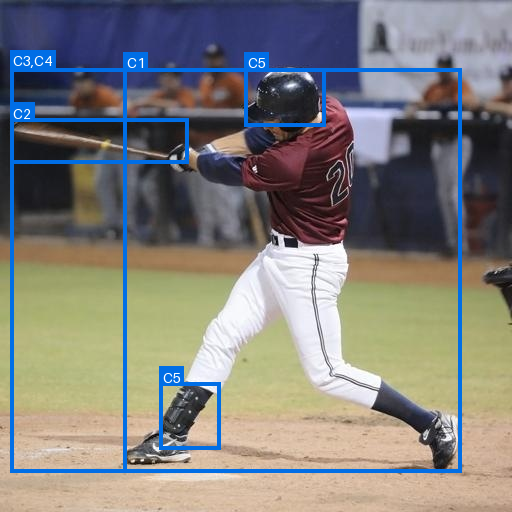}
\end{minipage}
\hfill
%
\begin{minipage}[t]{0.19\linewidth}
\vspace{0pt}
\raggedright

{\fontsize{4.9}{5.3}\selectfont
\textbf{Caption.}
A baseball player swings a bat while wearing neither a helmet nor a shin guard.

\vspace{0.55mm}

\agglobal{0.75}{79}
}
\end{minipage}
\hfill
%
\begin{minipage}[t]{0.52\linewidth}
\vspace{0pt}
\raggedright

{\fontsize{4.55}{4.85}\selectfont
\textbf{Atomic H/V + Explanation}\par
\vspace{0.20mm}

\agclaim{C1}{baseball player}{1.00/0.98}
{person in baseball uniform and batting gear.}
\agclaim{C2}{bat}{1.00/0.96}
{bat clearly visible in the player's hands.}
\agclaim{C3}{swings}{1.00/0.97}
{body and bat position depict a swing.}
\agclaim{C4}{baseball player swings a bat}{1.00/0.95}
{player is visibly holding and swinging the bat.}
\agclaim{C5}{baseball player wearing neither a helmet nor a shin guard}{0.00/0.10}
{both a helmet and shin guard are visible.}
}

\end{minipage}

\end{tcolorbox}


\begin{tcolorbox}[
    enhanced,
    width=\textwidth,
    colback=orange!2,
    colframe=orange!55!black,
    colbacktitle=orange!8,
    coltitle=black,
    title=\textbf{Medium $\boldsymbol{\cdot}$ T2I},
    fonttitle=\small,
    boxrule=0.4pt,
    arc=0.8mm,
    boxsep=0.4mm,
    left=0.8mm,
    right=0.8mm,
    top=0.45mm,
    bottom=0.45mm,
    before skip=0pt,
    after skip=0pt
]

\begin{minipage}[t]{0.13\linewidth}
\vspace{0pt}
\centering
{\fontsize{5.6}{5.9}\selectfont\textbf{Input}}\par
\vspace{0.25mm}

\includegraphics[
    width=\linewidth,
    height=1.90cm,
    keepaspectratio
]{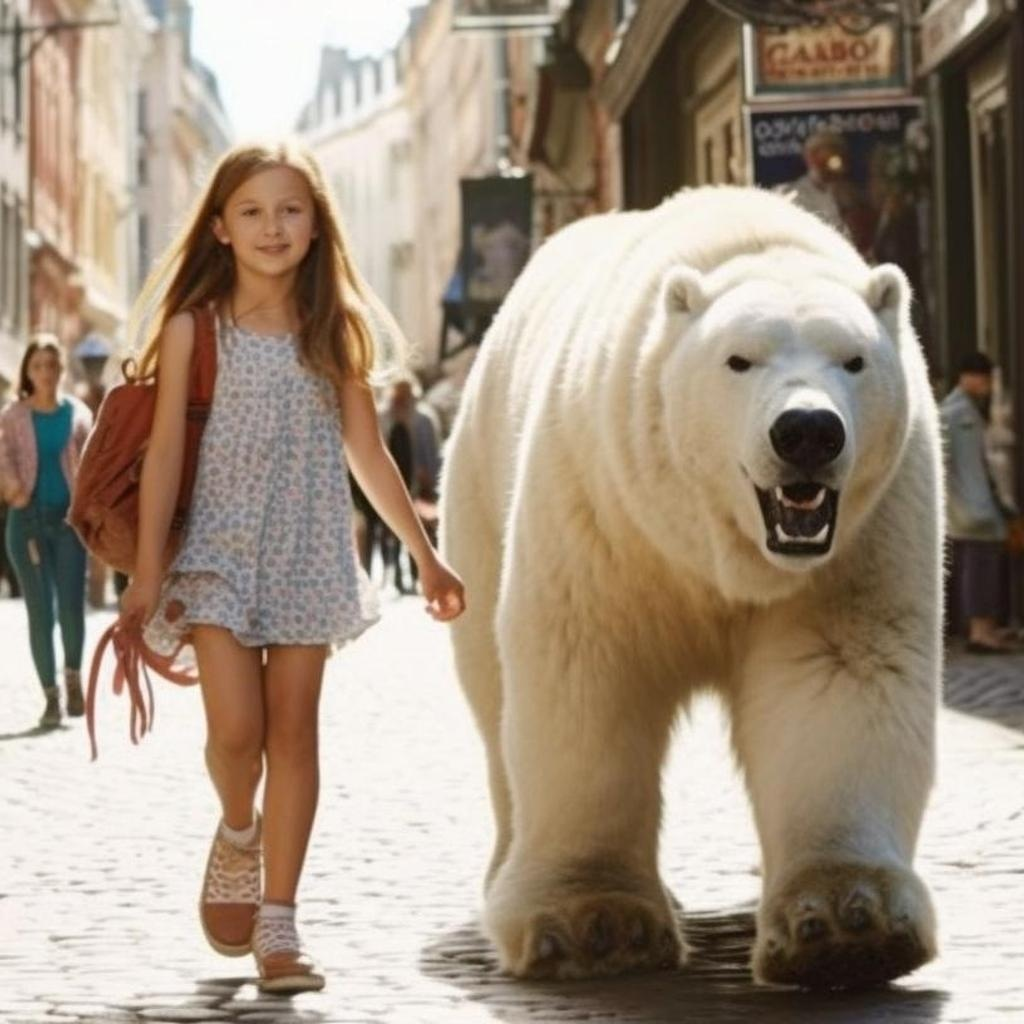}
\end{minipage}
\hfill
%
\begin{minipage}[t]{0.13\linewidth}
\vspace{0pt}
\centering
{\fontsize{5.6}{5.9}\selectfont\textbf{Evidence}}\par
\vspace{0.25mm}

\includegraphics[
    width=\linewidth,
    height=1.90cm,
    keepaspectratio
]{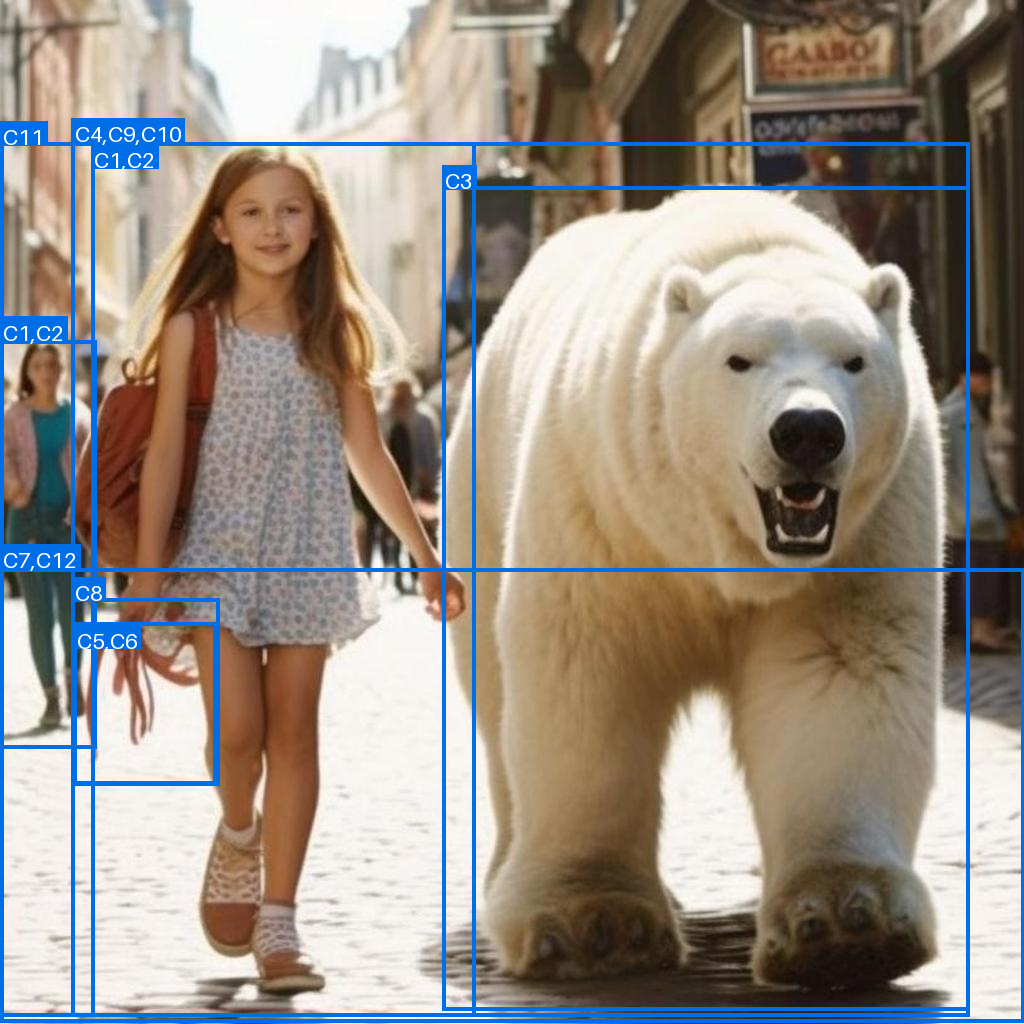}
\end{minipage}
\hfill
%
\begin{minipage}[t]{0.19\linewidth}
\vspace{0pt}
\raggedright

{\fontsize{4.50}{4.90}\selectfont
\textbf{Prompt.}
two swedish girls, 9 and 11 years old, holding their gigantic pet male polar
bear in tight leash while walking the streets of stockholm, summertime, sunny
and warm life style stock image popular no text prompt trend. pinterest
contest winner

\vspace{0.45mm}

\agglobal{0.75}{0.72}
}
\end{minipage}
\hfill
%
\begin{minipage}[t]{0.52\linewidth}
\vspace{0pt}
\raggedright

{\fontsize{4.10}{4.40}\selectfont
\textbf{Atomic H/V + Explanation}\par
\vspace{0.20mm}

\begin{minipage}[t]{0.488\linewidth}
\vspace{0pt}

\agclaim{C1}{girls}{1.00/0.89}
{a foreground girl and another female-presenting pedestrian are visible.}
\agclaim{C2}{two girls}{1.00/0.63}
{two female-presenting people are visible, but only one is clearly a girl.}
\agclaim{C3}{polar bear}{1.00/0.98}
{a large white bear is clearly visible.}
\agclaim{C4}{gigantic polar bear}{1.00/0.91}
{its scale relative to the girl supports ``gigantic.''}
\agclaim{C5}{leash}{0.00/0.67}
{straps are visible, but no attachment to the bear is shown.}
\agclaim{C6}{tight leash}{0.00/0.07}
{the visible strap is slack rather than tight.}

\end{minipage}
\hfill
\begin{minipage}[t]{0.488\linewidth}
\vspace{0pt}

\agclaim{C7}{streets}{1.00/0.95}
{the figures are on a paved city street.}
\agclaim{C8}{holding}{1.00/0.93}
{the girl's hand visibly holds the strap-like object.}
\agclaim{C9}{walking}{1.00/0.92}
{leg positions depict forward walking.}
\agclaim{C10}{girls holding their polar bear in a tight leash}{0.46/0.10}
{only one girl holds slack straps, with no tight attachment to the bear.}
\agclaim{C11}{girls walking the streets}{1.00/0.62}
{two female-presenting figures walk in the street, though age is unclear.}
\agclaim{C12}{sunny}{1.00/0.91}
{direct light and pronounced shadows indicate sunny conditions.}

\end{minipage}
}

\end{minipage}

\end{tcolorbox}

\vspace{0.5mm}


\begin{tcolorbox}[
    enhanced,
    width=\textwidth,
    colback=blue!1,
    colframe=blue!40!black,
    colbacktitle=blue!6,
    coltitle=black,
    title=\textbf{Hard $\boldsymbol{\cdot}$ I2T},
    fonttitle=\small,
    boxrule=0.4pt,
    arc=0.8mm,
    boxsep=0.4mm,
    left=0.8mm,
    right=0.8mm,
    top=0.45mm,
    bottom=0.45mm,
    before skip=0pt,
    after skip=0.55mm
]

\begin{minipage}[t]{0.13\linewidth}
\vspace{0pt}
\centering
{\fontsize{5.6}{5.9}\selectfont\textbf{Input}}\par
\vspace{0.25mm}

\includegraphics[
    width=\linewidth,
    height=1.90cm,
    keepaspectratio
]{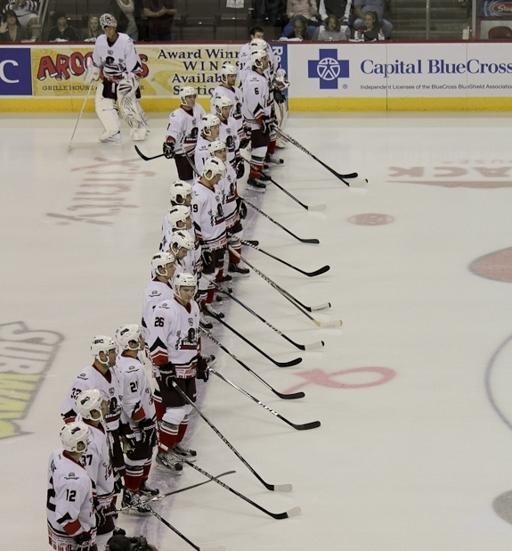}
\end{minipage}
\hfill
%
\begin{minipage}[t]{0.13\linewidth}
\vspace{0pt}
\centering
{\fontsize{5.6}{5.9}\selectfont\textbf{Evidence}}\par
\vspace{0.25mm}

\includegraphics[
    width=\linewidth,
    height=1.90cm,
    keepaspectratio
]{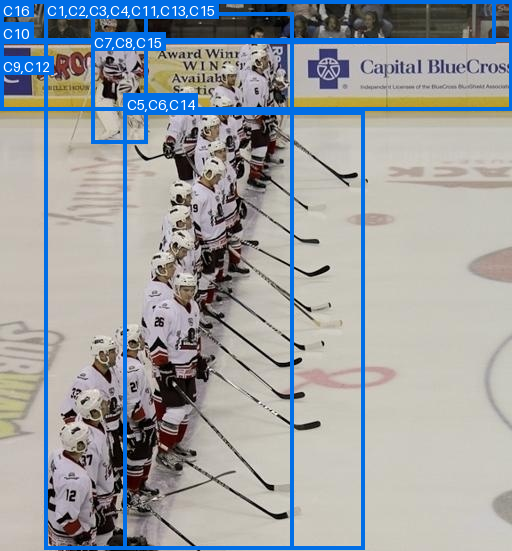}
\end{minipage}
\hfill
%
\begin{minipage}[t]{0.19\linewidth}
\vspace{0pt}
\raggedright

{\fontsize{4.55}{4.95}\selectfont
\textbf{Caption.}
A team in predominantly white jerseys with black and red trim stands in a
diagonal line stretching from the upper left toward the lower right, with
mostly dark sticks angled to the right; the padded goaltender occupies the
middle of the line as spectators watch from behind the advertising boards.

\vspace{0.45mm}

\agglobal{0.75}{0.81}
}
\end{minipage}
\hfill
%
\begin{minipage}[t]{0.52\linewidth}
\vspace{0pt}
\raggedright

{\fontsize{4.10}{4.40}\selectfont
\textbf{Atomic H/V + Explanation}\par
\vspace{0.20mm}

\begin{minipage}[t]{0.488\linewidth}
\vspace{0pt}

\agclaim{C1}{team}{1.00/0.98}
{uniforms and formation indicate a team.}
\agclaim{C2}{jerseys}{1.00/0.97}
{numbered jerseys are visible.}
\agclaim{C3}{predominantly white jerseys}{1.00/0.95}
{jerseys are predominantly white.}
\agclaim{C4}{jerseys with black and red trim}{1.00/0.89}
{black and red trim is visible.}
\agclaim{C5}{sticks}{1.00/0.93}
{multiple hockey sticks are visible.}
\agclaim{C6}{mostly dark sticks}{1.00/0.83}
{most visible sticks are dark.}
\agclaim{C7}{goaltender}{1.00/0.97}
{mask and large pads identify a goaltender.}
\agclaim{C8}{padded goaltender}{1.00/0.94}
{large protective pads are visible.}

\end{minipage}
\hfill
\begin{minipage}[t]{0.488\linewidth}
\vspace{0pt}

\agclaim{C9}{spectators}{1.00/0.90}
{spectators are visible above the boards.}
\agclaim{C10}{advertising boards}{1.00/0.92}
{text and logos appear on the boards.}
\agclaim{C11}{stands}{1.00/0.96}
{players are standing upright.}
\agclaim{C12}{watch}{1.00/0.62}
{spectators face the ice, though individual gaze is unclear.}
\agclaim{C13}{team stands in a diagonal line from upper left to lower right}{0.50/0.18}
{the visible diagonal has the opposite orientation.}
\agclaim{C14}{sticks angled to the right}{1.00/0.95}
{most visible sticks project to the right.}
\agclaim{C15}{goaltender in the middle of the line}{0.75/0.11}
{the goaltender is near an end rather than the middle.}
\agclaim{C16}{spectators behind the advertising boards}{1.00/0.92}
{spectators are visibly behind the boards.}

\end{minipage}
}

\end{minipage}

\end{tcolorbox}

\endgroup

\vspace{1mm}

\captionsetup{skip=1pt}
\setlength{\abovecaptionskip}{0pt}  
\setlength{\belowcaptionskip}{-3pt}  
\caption{
\textbf{Qualitative AGITA-Bench examples with \textsc{VisExMEM}-9B.}
Easy--Hard cases show the input, localized evidence, caption/prompt, and
Human/\textsc{VisExMEM}-9B global and atomic scores (H/V); human ratings are
normalized to $[0,1]$, with short claim-level explanations.
}
\vspace{-2mm}
\label{fig:agita_qual_examples}
\end{figure*}

%% file: strucuture/5_experiments.tex
\section{Experiments}
\label{sec:experiments}

\paragraph{Benchmarks and Protocols.}
We evaluate ten benchmarks, comprising eight established benchmarks,
T2IScoreScore (TS2), and AGITA-Bench. We report Group scores on
Winoground (WG)~\citep{thrush2022winogroundprobingvisionlanguage} and
EqBen~\citep{wang2023equivariantsimilarityvisionlanguagefoundation},
pairwise accuracy (PAcc) on
TIFA160~\citep{hu2023tifaaccurateinterpretabletexttoimage} and
GenAI-Bench~\citep{li2024genaibenchevaluatingimprovingcompositional}, and
preference accuracy (Pref.\ Acc.) on
Pick-a-Pic~\citep{kirstain2023pickapicopendatasetuser}.
For CROC$_{\mathrm{hum}}$~\citep{leiter2026crocevaluatingtrainingt2i}, we use
the mean over its forward and inverse text-based and image-based evaluations.
We report SRCC on the T2I-CompBench++ (T2I-CB++) human
subset~\citep{huang2025t2icompbenchenhancedcomprehensivebenchmark} and top-1
counting accuracy on CountBench~\citep{paiss2023teachingclipcount}.
TS2~\citep{saxon2024evaluatesevaluationsobjectivelyscoring} organizes images
into semantic error graphs (SEGs). We report ordering $\mathrm{rank}_m$ and
separation $\mathrm{sep}_m$ over all SEGs and the Synth, Nat, and Real subsets.
Synth uses generated images for predefined error graphs, Nat uses natural
images in constructed graphs, and Real uses naturally occurring T2I errors.
AGITA-Bench (Sec.~\ref{sec:agita}) reports Spearman's $\rho$ for global and
atomic human agreement across Easy, Medium, and Hard. Full protocols are in
Appendix~\ref{app:benchmark_protocols}. We also evaluate OOD generalization on
VisPlotBench~\citep{ni2025viscoder2},
CulturalFrames~\citep{nayak2025culturalframes}, and
ProtoBias~\citep{roy2026prototypicalitybiasrevealsblindspots} in
Appendix~\ref{app:ood_generalization}.


\paragraph{Evaluation Metrics.}
We compare \textsc{VisExMEM}-4B and \textsc{VisExMEM}-9B with
\textbf{embedding- and matching-based}
CLIPScore~\citep{hessel2022clipscorereferencefreeevaluationmetric},
BLIPv2Score~\citep{li2023blip2}, and
ALIGNScore~\citep{saxon2024evaluatesevaluationsobjectivelyscoring},
\textbf{preference-trained}
PickScore~\citep{kirstain2023pickapicopendatasetuser},
ImageReward~\citep{xu2023imagereward},
HPSv2~\citep{wu2023humanpreferencescorev2}, and
MPS~\citep{zhang2024learningmultidimensionalhumanpreference},
\textbf{general-purpose or explainable evaluators}
GPT4V-Eval~\citep{zhang2023gpt4visiongeneralistevaluatorvisionlanguage},
VIEScore~\citep{ku2024viescoreexplainablemetricsconditional},
FLEUR, and EXPERT, and
\textbf{structured and VQA-based}
VQ$^2$~\citep{yarom2023readimprovingtextimagealignment},
Davidsonian Scene Graph~\citep{cho2024davidsonianscenegraphimproving},
our DSG + GPT-5.6 Sol setup, and
VQAScore~\citep{lin2024evaluatingtexttovisualgenerationimagetotext}.
We also evaluate GroundingDINO~\citep{liu2024groundingdinomarryingdino}
directly as a grounding-based metric.
Controlled \textbf{VLM-as-a-Judge} experiments use GPT-5.6 Sol and untuned
Qwen3.5~\citep{qwen35blog} with \emph{Holistic (Hol.)} prompting on the full
image-caption pair or \emph{Decomposed (Dec.)} prompting with a decompose,
verify, and aggregate procedure, without \textsc{VisExMEM}-specific training,
learned evidence conditioning, or external grounding cues. On AGITA-Bench,
global agreement uses native scores, while atomic agreement is reported only
for methods directly scoring the fixed benchmark claims. Prompts are given in
Appendix~\ref{app:vlm_judge_protocol}. The omission of
REVEALER~\citep{shi2026revealerreinforcementguidedvisualreasoning} and
T2I-FineEval~\citep{hosseini2025t2ifineeval} is discussed in
Appendix~\ref{app:baseline_coverage}.

\paragraph{Explanation Evaluation.}
We compare EXPERT, VIEScore, GPT-5.6 Sol (Dec.), and
\textsc{VisExMEM}-9B. Annotators rate each explanation from 1 to 5 for
diagnostic correctness, visual grounding, clarity, and diagnostic usefulness.
Correctness measures whether the explanation identifies the relevant matches
or errors, grounding whether its statements are supported by the image and
caption, clarity how easy it is to understand, and usefulness how well it
explains why the image and caption agree or disagree. Annotators also flag
unsupported factual content. For systems with localized evidence, they rate
localization accuracy and sufficiency from 1 to 5. Localization measures
whether the highlighted region identifies the relevant visual evidence, while
sufficiency measures whether that evidence is enough to judge the claim.
Full instructions and agreement statistics are given in
Appendix~\ref{app:explanation_annotation}.

%% file: strucuture/6_results_analysis.tex
\section{Results \& Analysis}
\label{sec:results}

\input{figures/results_main}
\paragraph{Comparison with Existing Metrics.}
Table~\ref{tab:main_results} compares performance across the eight established
benchmarks. \textsc{VisExMEM}-9B has the highest composite among open-source
evaluators in our comparison. It exceeds VQAScore on all eight, with an
11.7-point composite difference. \textsc{VisExMEM}-4B improves the composite
by 6.8 points over VQAScore and exceeds all prior open-source baselines.
For 9B, the largest gains are on Winoground (+22.9), EqBen (+22.2), and
CountBench (+29.0), with a further +6.1 points on GenAI-Bench.
\textsc{VisExMEM}-9B also exceeds GPT-5.6 Sol (Hol.) on five benchmarks and
matches it on EqBen. Its eight-benchmark composite is 72.2, compared with
74.0 for GPT-5.6 Sol (Hol.) and 77.6 for GPT-5.6 Sol (Dec.).

\paragraph{Structured Evaluation Helps Across Backbones.} Replacing holistic judgment with a \visexmem-style procedure that decomposes, verifies, and aggregates improves the average performance of Qwen3.5-4B, Qwen3.5-9B, and GPT-5.6 Sol by +5.0, +5.1, and +3.6 points, respectively, with GPT-5.6 Sol increasing from 74.0 (Hol.) to 77.6 (Dec.). These controlled comparisons show that structured inference improves average performance across all three VLM backbones, although gains vary by benchmark. Together with dedicated \textsc{VisExMEM} training, the 4B and 9B models achieve the strongest open-source results in our comparison.

\begin{figure}[t]
    \centering
    \includegraphics[width=.95\linewidth]{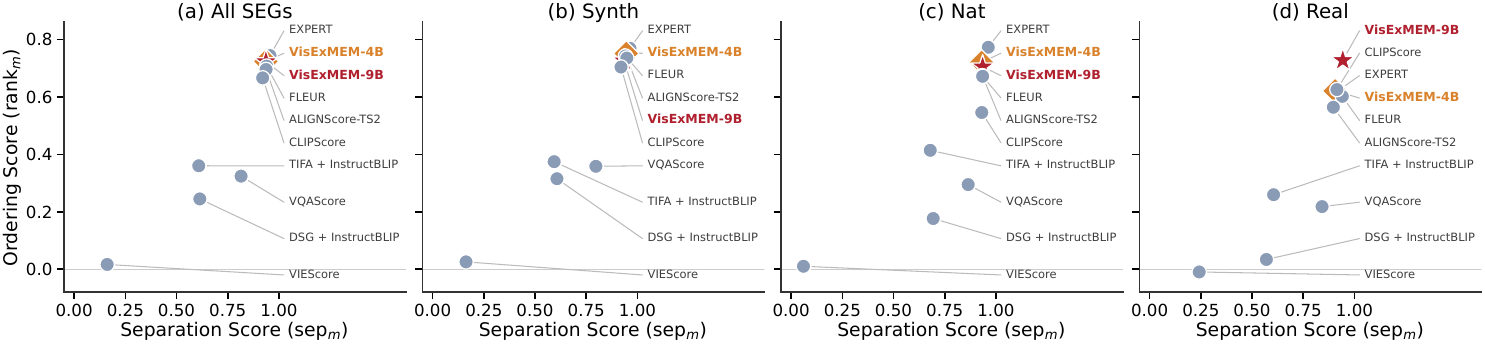}
    \caption{\textbf{T2IScoreScore performance.}
    Ordering ($\mathrm{rank}_m$) versus separation ($\mathrm{sep}_m$) for (a) all SEGs, (b) Synth, (c) Nat, and (d) Real. Higher is better on both axes.}
    \label{fig:ts2}
\end{figure}

\paragraph{T2IScoreScore Performance.}
Figure~\ref{fig:ts2} shows that both \textsc{VisExMEM} variants remain in the high-ordering, high-separation region across \emph{All, Synth, Nat, and Real}. \textsc{VisExMEM}-4B performs strongly on All, Synth, and Nat, while \textsc{VisExMEM}-9B shows its clearest advantage on Real, where it achieves the highest ordering score while maintaining strong separation. This indicates strong performance on naturally occurring T2I errors as well as the constructed TS2 settings.

\input{figures/results_agita_ablation}

\paragraph{Fine-Grained Human Alignment on AGITA-Bench.}
Table~\ref{tab:agita_results} reports agreement with human judgments at both
the global and atomic levels. \textsc{VisExMEM}-9B achieves the strongest
open-source global average at 54.7 and improves over the corresponding untuned
Qwen3.5 controls across both evaluation levels. Among methods that directly
score the fixed benchmark claims, it reaches an atomic average of 44.9,
compared with 64.7 for GPT-5.6 Sol (Dec.). The higher correlations on Hard
than Medium should not be read as Hard being easier. In T2I, difficulty
reflects semantic complexity, and Hard prompts contain more claims
(15.72 vs.\ 8.58) and lower human alignment (2.99 vs.\ 3.49), which changes
the within-tier score distribution. This pattern is not observed in I2T,
where Medium and Hard have comparable claim counts.


\paragraph{Explanation Quality Analysis.}
\label{sec:explanation_results}

\begin{figure}[t]
    \centering
    \includegraphics[width=.90\linewidth]{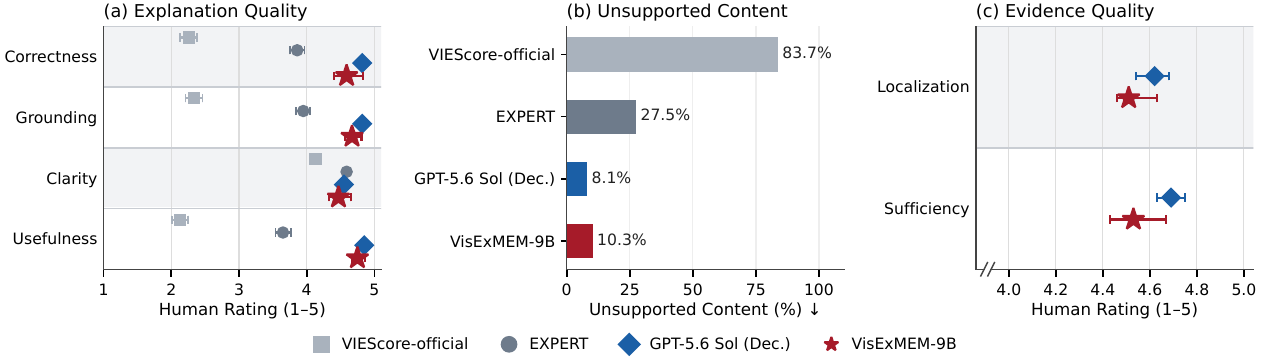}
    \caption{\textbf{Human evaluation of explanation quality.}
    (a) Correctness, visual grounding, clarity, and diagnostic usefulness.
    (b) Unsupported-content rate. (c) Localization accuracy and evidence
    sufficiency. Error bars show 95\% confidence intervals.}
    \label{fig:explanation_quality}
\end{figure}

Figure~\ref{fig:explanation_quality} shows that \textsc{VisExMEM}-9B is close
to GPT-5.6 Sol on explanation quality, reaching 4.75 diagnostic usefulness
with 10.3\% unsupported content, compared with 4.85 and 8.1\% for
GPT-5.6 Sol. It exceeds EXPERT and VIEScore in usefulness
(3.65 and 2.13) while producing much less unsupported content
(27.5\% and 83.7\%). Its localized evidence also receives high ratings for
localization accuracy and sufficiency, showing that the highlighted regions
are relevant and provide useful evidence for judging the associated claims.

\paragraph{Training-Stage Ablation.}
\label{sec:stage_ablation}

Table~\ref{tab:stage_ablation} shows that SFT improves the five-benchmark
average over Base by +1.9/+9.8 points for 4B/9B, although the gains vary across
benchmarks. Counterfactual refinement then gives a larger and more consistent
gain over SFT of +8.8/+9.0 points. From the counterfactual checkpoint,
Verification-GRPO and Causal-GRPO provide smaller, less consistent gains,
while joint GDPO performs best on all five benchmarks at both scales and adds
a further +3.3/+4.5 points on average. This supports the progression from
structured supervision to counterfactual refinement and joint semantic and
evidence-aware optimization.

\paragraph{Efficiency, Cost, and Reproducibility.}
Although GPT-5.6 Sol benefits from the decomposed evaluation procedure,
evaluating the eight benchmarks and AGITA-Bench cost approximately \$800
(\(\sim\)\$150 holistic and \(\sim\)\$650 decomposed). With vLLM-based~\citep{vLLM2023kwon}
inference, \textsc{VisExMEM}-9B requires 8.9\,s per image-text pair including
explanation generation, compared with about 30\,s for GPT-5.6 Sol. At 9B
parameters, \textsc{VisExMEM}-9B provides a compact, locally deployable
alternative to proprietary frontier judges. Its fixed checkpoint,
deterministic aggregation, and explicit atomic decisions and visual evidence
support interpretability and reproducibility without reliance on a
proprietary API.

%% file: figures/results_main.tex

\begin{table*}[t]
\centering

\caption{\textbf{Main results.}
Comparison of prior image--text alignment metrics, controlled VLM-as-a-Judge
baselines, and \textsc{VisExMEM} across eight established benchmarks.
Higher is better. \textbf{Bold}, \underline{underlined}, and
\textit{italicized} values mark the best, second-best, and third-best results.}
\label{tab:main_results}

\scriptsize
\setlength{\tabcolsep}{2.2pt}
\renewcommand{\arraystretch}{0.94}

\resizebox{0.9\textwidth}{!}{%
\begin{tabular}{lccccccccc}
\toprule
\textbf{Method}
& \textbf{WG}
& \textbf{EqBen}
& \textbf{TIFA160}
& \textbf{Pick-a-Pic}
& \textbf{GenAI}
& \textbf{CROC$_{\mathrm{hum}}$}
& \textbf{T2I-CB++}
& \textbf{CountBench}
& \textbf{Avg.}
\\[-0.5mm]

& \textbf{Group}
& \textbf{Group}
& \textbf{PAcc}
& \textbf{Pref. Acc.}
& \textbf{PAcc}
& \textbf{Score}
& \textbf{SRCC}
& \textbf{Acc.}
& \\
\midrule


CLIPScore
& 7.8
& 25.0
& 54.1
& 76.0
& 52.2
& 69.8
& 28.5
& 30.7
& 43.0
\\

BLIPv2Score
& 17.5
& 35.0
& 57.5
& 80.0
& 55.1
& 76.1
& 37.0
& 47.3
& 50.7
\\

ALIGNScore
& 8.3
& 28.6
& 59.6
& 80.0
& 53.4
& 71.8
& 36.1
& 34.5
& 46.5
\\

\midrule


PickScore
& 6.8
& 23.6
& 59.4
& 70.0
& 57.7
& 74.7
& 30.3
& 56.5
& 47.4
\\

ImageReward
& 12.8
& 26.4
& 67.3
& 75.0
& 58.7
& 75.4
& 50.6
& 33.8
& 50.0
\\

HPSv2
& 4.0
& 17.1
& 55.2
& 69.0
& 49.8
& 72.3
& 42.4
& \textit{66.6}
& 47.0
\\

MPS
& 1.8
& 10.0
& 59.0
& 69.0
& 56.1
& 67.0
& 24.2
& 33.8
& 40.1
\\

\midrule


GPT4V-Eval
& 36.3
& 35.0
& 64.0
& 74.0
& 61.4
& --
& --
& --
& 54.1\textsuperscript{$\dagger$}
\\

VIEScore
& 34.5
& 32.9
& 63.9
& 78.0
& 60.3
& 86.5
& 54.9
& 47.3
& 57.3
\\

EXPERT
& 29.3
& 27.1
& 68.0
& 83.0
& 64.6
& 83.0
& 54.2
& 43.9
& 56.6
\\

FLEUR
& 28.5
& 35.7
& 67.2
& 82.0
& 63.5
& 81.7
& 52.2
& 41.0
& 56.5
\\

\midrule


VQ$^{2}$
& 10.0
& 20.0
& 48.7
& 73.0
& 51.9
& 78.1
& 48.6
& 51.1
& 47.7
\\

Davidsonian
& 15.5
& 20.0
& 54.3
& 70.0
& 54.6
& 77.7
& \textit{60.6}
& 37.9
& 48.8
\\

DSG + GPT-5.6 Sol
& 49.0
& 40.6
& 66.8
& 81.0
& 61.4
& 72.1
& 49.8
& 51.3
& 59.0
\\

\rowcolor{VQABG}
VQAScore
& 46.0
& 47.9
& 71.2
& \textit{84.0}
& 63.3
& 86.1
& 53.2
& 31.7
& 60.4
\\

\midrule


GroundingDINO
& 9.3
& 12.9
& 58.2
& 57.0
& 51.1
& 56.8
& 18.5
& 20.6
& 35.6
\\

\midrule


Qwen3.5-4B (Hol.)
& 28.2
& 44.0
& 64.3
& 72.5
& 62.2
& 78.1
& 56.7
& 36.6
& 55.3
\\

Qwen3.5-4B (Dec.)
& 49.0
& 47.1
& 66.0
& 80.0
& 62.5
& 76.5
& 57.2
& 44.2
& 60.3
\\

\rowcolor{black!4}
\textit{$\Delta$ Dec.--Hol.}
& \textcolor{gray!70!black}{(+20.8)}
& \textcolor{gray!70!black}{(+3.1)}
& \textcolor{gray!70!black}{(+1.7)}
& \textcolor{gray!70!black}{(+7.5)}
& \textcolor{gray!70!black}{(+0.3)}
& \textcolor{gray!70!black}{(-1.6)}
& \textcolor{gray!70!black}{(+0.5)}
& \textcolor{gray!70!black}{(+7.6)}
& \textcolor{gray!70!black}{(+5.0)}
\\

\addlinespace[0.4mm]

Qwen3.5-9B (Hol.)
& 40.4
& 35.7
& 50.1
& 74.2
& 50.3
& 76.2
& 55.8
& 25.5
& 51.0
\\

Qwen3.5-9B (Dec.)
& 41.2
& 45.7
& 62.2
& 81.0
& 54.6
& 74.4
& 54.8
& 35.0
& 56.1
\\

\rowcolor{black!4}
\textit{$\Delta$ Dec.--Hol.}
& \textcolor{gray!70!black}{(+0.8)}
& \textcolor{gray!70!black}{(+10.0)}
& \textcolor{gray!70!black}{(+12.1)}
& \textcolor{gray!70!black}{(+6.8)}
& \textcolor{gray!70!black}{(+4.3)}
& \textcolor{gray!70!black}{(-1.8)}
& \textcolor{gray!70!black}{(-1.0)}
& \textcolor{gray!70!black}{(+9.5)}
& \textcolor{gray!70!black}{(+5.1)}
\\

\addlinespace[0.4mm]

\rowcolor{GPTBG}
GPT-5.6 Sol (Hol.)
& 58.5
& \textbf{70.1}
& \textit{71.8}
& 82.4
& \textit{66.9}
& \textit{91.0}
& \textbf{63.8}
& \textbf{87.6}
& \underline{74.0}
\\

\rowcolor{GPTBG}
GPT-5.6 Sol (Dec.)
& \textbf{80.0}
& \underline{70.0}
& \textbf{74.3}
& \textbf{86.0}
& \textbf{70.5}
& \textbf{94.0}
& \underline{61.1}
& \underline{85.1}
& \textbf{77.6}
\\

\rowcolor{black!4}
\textit{$\Delta$ Dec.--Hol.}
& \textcolor{gray!70!black}{(+21.5)}
& \textcolor{gray!70!black}{(-0.1)}
& \textcolor{gray!70!black}{(+2.5)}
& \textcolor{gray!70!black}{(+3.6)}
& \textcolor{gray!70!black}{(+3.6)}
& \textcolor{gray!70!black}{(+3.0)}
& \textcolor{gray!70!black}{(-2.7)}
& \textcolor{gray!70!black}{(-2.5)}
& \textcolor{gray!70!black}{(+3.6)}
\\

\midrule


\rowcolor{green!10}
\textbf{VisExMEM-4B (Ours)}
& \textit{65.6}
& \textit{59.1}
& 70.2
& 80.2
& 66.3
& 87.0
& 57.8
& 51.9
& 67.3
\\

\rowcolor{blue!4}
\textit{vs.\ VQAScore}
& \textcolor{blue!70!black}{(+19.6)}
& \textcolor{blue!70!black}{(+11.2)}
& \textcolor{blue!70!black}{(-1.0)}
& \textcolor{blue!70!black}{(-3.8)}
& \textcolor{blue!70!black}{(+3.0)}
& \textcolor{blue!70!black}{(+0.9)}
& \textcolor{blue!70!black}{(+4.6)}
& \textcolor{blue!70!black}{(+20.2)}
& \textcolor{blue!70!black}{(+6.8)}
\\

\rowcolor{green!10}
\textbf{VisExMEM-9B (Ours)}
& \underline{68.9}
& \textbf{70.1}
& \underline{73.1}
& \underline{85.6}
& \underline{69.4}
& \underline{91.7}
& 57.7
& 60.7
& \textit{72.2}
\\

\rowcolor{blue!4}
\textit{vs.\ VQAScore}
& \textcolor{blue!70!black}{(+22.9)}
& \textcolor{blue!70!black}{(+22.2)}
& \textcolor{blue!70!black}{(+1.9)}
& \textcolor{blue!70!black}{(+1.6)}
& \textcolor{blue!70!black}{(+6.1)}
& \textcolor{blue!70!black}{(+5.6)}
& \textcolor{blue!70!black}{(+4.5)}
& \textcolor{blue!70!black}{(+29.0)}
& \textcolor{blue!70!black}{(+11.7)}
\\

\bottomrule
\end{tabular}%
}

\vspace{0.6mm}

\parbox{0.9\textwidth}{%
\scriptsize
\textsuperscript{$\dagger$} GPT4V-Eval's Avg. uses its five available
benchmark results; all other Avg. values use all eight. T2I-CB++ reports
$100\rho$ (theoretical range $[-100,100]$). Avg. is a descriptive composite
of different benchmark statistics. Blue rows show signed differences
relative to VQAScore in the displayed units; T2I-CB++ differences are
$100\Delta\rho$, not percentage points.
}

\end{table*}

%% file: figures/results_agita_ablation.tex

\begin{table*}[t]
\centering

\begin{tabular}{
    @{}
    p{0.505\textwidth}
    @{\hspace{0.018\textwidth}}
    p{0.457\textwidth}
    @{}
}


\captionof{table}{
\textbf{Human Agreement on AGITA-Bench.}
Global and atomic agreement across Easy, Medium, and Hard difficulty tiers.
\textbf{Bold}, \underline{underline}, and \textit{italics} denote
first, second, and third best, respectively.
}
\label{tab:agita_results}
&
\captionof{table}{
\textbf{Training-Stage Ablation.}
Performance of \textsc{VisExMEM} across successive training stages at
4B and 9B scales. Blue rows show gains relative to the indicated reference
stage; \textbf{bold} denotes the best result within each model scale.
}
\label{tab:stage_ablation}
\\[-8.5mm]


\begin{minipage}[t]{\linewidth}
\vspace{0pt}
\centering

\fontsize{6.4}{6.9}\selectfont
\setlength{\tabcolsep}{1.35pt}
\renewcommand{\arraystretch}{0.91}

\resizebox{\linewidth}{!}{%
\begin{tabular}{@{}lcccccccc@{}}
\toprule
&
\multicolumn{4}{c}{\textbf{Global}}
&
\multicolumn{4}{c}{\textbf{Atomic}}
\\
\cmidrule(lr){2-5}
\cmidrule(lr){6-9}

\textbf{Method}
& \textbf{Easy}
& \textbf{Med.}
& \textbf{Hard}
& \textbf{Avg.}
& \textbf{Easy}
& \textbf{Med.}
& \textbf{Hard}
& \textbf{Avg.}
\\
\midrule

CLIPScore
& 19.6 & -3.6 & -8.6 & 2.5
& 5.4 & -4.8 & -0.2 & 0.1
\\

BLIPv2Score
& 21.1 & 3.1 & -0.8 & 7.8
& 3.3 & 2.1 & 4.5 & 3.3
\\

AlignScore
& 24.5 & 3.2 & 6.2 & 11.3
& 5.8 & -0.5 & 0.5 & 1.9
\\

PickScore
& 33.5 & 29.1 & 34.7 & 32.4
& 11.1 & -3.2 & -2.0 & 2.0
\\

ImageReward
& 38.8 & 21.9 & 16.8 & 25.8
& -- & -- & -- & --
\\

HPSv2
& 17.5 & 14.6 & 13.5 & 15.2
& 14.8 & 5.8 & 8.0 & 9.5
\\

EXPERT
& 50.6 & 33.8 & 45.5 & 43.3
& -- & -- & -- & --
\\

FLEUR
& 52.2 & 31.5 & 43.7 & 42.5
& -- & -- & -- & --
\\

VIEScore
& 51.2 & 31.2 & 42.6 & 41.7
& -- & -- & -- & --
\\

VQAScore
& 54.6 & 31.7 & 43.0 & 43.1
& \textit{52.7} & 32.3 & 41.1 & \textit{42.0}
\\

\midrule

Qwen3.5-4B (Hol.)
& 48.5 & 39.3 & 40.3 & 42.7
& -- & -- & -- & --
\\

Qwen3.5-4B (Dec.)
& 50.5 & 37.1 & 44.1 & 43.9
& 41.2 & 31.7 & 39.1 & 37.3
\\

Qwen3.5-9B (Hol.)
& 53.7 & 42.2 & 40.4 & 45.4
& -- & -- & -- & --
\\

Qwen3.5-9B (Dec.)
& 59.0 & 39.5 & 41.2 & 46.6
& 40.6 & 30.8 & 38.2 & 36.5
\\

GPT-5.6 Sol (Hol.)
& \underline{72.1}
& \underline{54.6}
& \underline{60.6}
& \underline{62.4}
& -- & -- & -- & --
\\

GPT-5.6 Sol (Dec.)
& \textbf{76.9}
& \textbf{60.7}
& \textbf{62.5}
& \textbf{66.7}
& \textbf{70.1}
& \textbf{55.6}
& \textbf{68.3}
& \textbf{64.7}
\\

\midrule

\rowcolor{green!10}
\textbf{VisExMEM-4B}
& 56.8
& 43.1
& 50.6
& 50.2
& 46.6
& \underline{34.1}
& \textit{43.2}
& 41.3
\\

\rowcolor{green!10}
\textbf{VisExMEM-9B}
& \textit{63.3}
& \textit{45.2}
& \textit{55.6}
& \textit{54.7}
& \underline{56.2}
& \textit{33.0}
& \underline{45.5}
& \underline{44.9}
\\

\bottomrule
\end{tabular}%
}

\end{minipage}
&


\begin{minipage}[t]{\linewidth}
\vspace*{3mm}
\centering

\fontsize{6.3}{6.7}\selectfont
\setlength{\tabcolsep}{1.20pt}
\renewcommand{\arraystretch}{0.88}

\resizebox{\linewidth}{!}{%
\begin{tabular}{@{}llccccc@{}}
\toprule

\textbf{Model}
& \textbf{Variant}
& \textbf{WG}
& \textbf{EqB}
& \textbf{TIFA}
& \textbf{Pick}
& \textbf{GenAI}
\\
\midrule


\multirow{8}{*}{\textbf{4B}}
& Base (Hol.)
& 28.2 & 44.0 & 64.3 & 72.5 & 62.2
\\

\cmidrule(lr){2-7}

& SFT
& 50.8 & 48.9 & 56.8 & 70.0 & 54.1
\\

&
\cellcolor{blue!6}
\quad\textcolor{blue!70!black}{\textit{vs.\ Base}}
&
\cellcolor{blue!6}
\textcolor{blue!70!black}{(+22.6)}
&
\cellcolor{blue!6}
\textcolor{blue!70!black}{(+4.9)}
&
\cellcolor{blue!6}
\textcolor{blue!70!black}{(-7.5)}
&
\cellcolor{blue!6}
\textcolor{blue!70!black}{(-2.5)}
&
\cellcolor{blue!6}
\textcolor{blue!70!black}{(-8.1)}
\\

& \quad + CF
& 61.6 & 56.2 & 64.9 & 77.2 & 64.9
\\

&
\cellcolor{blue!6}
\quad\textcolor{blue!70!black}{\textit{vs.\ SFT}}
&
\cellcolor{blue!6}
\textcolor{blue!70!black}{(+10.8)}
&
\cellcolor{blue!6}
\textcolor{blue!70!black}{(+7.3)}
&
\cellcolor{blue!6}
\textcolor{blue!70!black}{(+8.1)}
&
\cellcolor{blue!6}
\textcolor{blue!70!black}{(+7.2)}
&
\cellcolor{blue!6}
\textcolor{blue!70!black}{(+10.8)}
\\

\cmidrule(lr){2-7}

& \qquad + Verif.-GRPO
& 62.5 & 57.9 & 66.4 & 73.9 & 64.3
\\

& \qquad + Causal-GRPO
& 64.1 & 58.6 & 67.6 & 72.8 & 65.2
\\

& \qquad + \textbf{GDPO}
& \textbf{65.6}
& \textbf{59.1}
& \textbf{70.2}
& \textbf{80.2}
& \textbf{66.3}
\\

&
\cellcolor{blue!6}
\qquad\textcolor{blue!70!black}{\textit{vs.\ CF}}
&
\cellcolor{blue!6}
\textcolor{blue!70!black}{(+4.0)}
&
\cellcolor{blue!6}
\textcolor{blue!70!black}{(+2.9)}
&
\cellcolor{blue!6}
\textcolor{blue!70!black}{(+5.3)}
&
\cellcolor{blue!6}
\textcolor{blue!70!black}{(+3.0)}
&
\cellcolor{blue!6}
\textcolor{blue!70!black}{(+1.4)}
\\

\midrule


\multirow{8}{*}{\textbf{9B}}
& Base (Hol.)
& 40.4 & 35.7 & 50.1 & 74.2 & 50.3
\\

\cmidrule(lr){2-7}

& SFT
& 61.9 & 58.9 & 59.8 & 61.8 & 57.2
\\

&
\cellcolor{blue!6}
\quad\textcolor{blue!70!black}{\textit{vs.\ Base}}
&
\cellcolor{blue!6}
\textcolor{blue!70!black}{(+21.5)}
&
\cellcolor{blue!6}
\textcolor{blue!70!black}{(+23.2)}
&
\cellcolor{blue!6}
\textcolor{blue!70!black}{(+9.7)}
&
\cellcolor{blue!6}
\textcolor{blue!70!black}{(-12.4)}
&
\cellcolor{blue!6}
\textcolor{blue!70!black}{(+6.9)}
\\

& \quad + CF
& 66.3 & 64.9 & 68.9 & 77.5 & 67.2
\\

&
\cellcolor{blue!6}
\quad\textcolor{blue!70!black}{\textit{vs.\ SFT}}
&
\cellcolor{blue!6}
\textcolor{blue!70!black}{(+4.4)}
&
\cellcolor{blue!6}
\textcolor{blue!70!black}{(+6.0)}
&
\cellcolor{blue!6}
\textcolor{blue!70!black}{(+9.1)}
&
\cellcolor{blue!6}
\textcolor{blue!70!black}{(+15.7)}
&
\cellcolor{blue!6}
\textcolor{blue!70!black}{(+10.0)}
\\

\cmidrule(lr){2-7}

& \qquad + Verif.-GRPO
& 67.4 & 67.2 & 71.5 & 78.0 & 67.9
\\

& \qquad + Causal-GRPO
& 68.5 & 66.7 & 71.0 & 79.1 & 67.1
\\

& \qquad + \textbf{GDPO}
& \textbf{68.9}
& \textbf{70.1}
& \textbf{73.1}
& \textbf{85.6}
& \textbf{69.4}
\\

&
\cellcolor{blue!6}
\qquad\textcolor{blue!70!black}{\textit{vs.\ CF}}
&
\cellcolor{blue!6}
\textcolor{blue!70!black}{(+2.6)}
&
\cellcolor{blue!6}
\textcolor{blue!70!black}{(+5.2)}
&
\cellcolor{blue!6}
\textcolor{blue!70!black}{(+4.2)}
&
\cellcolor{blue!6}
\textcolor{blue!70!black}{(+8.1)}
&
\cellcolor{blue!6}
\textcolor{blue!70!black}{(+2.2)}
\\

\bottomrule
\end{tabular}%
}

\end{minipage}
\\

\end{tabular}

\vspace{-3mm}
\end{table*}

%% file: strucuture/7_conclusion.tex
\section{Conclusion, Limitations, and Future Work}

We introduced \textsc{VisExMEM}, an evidence-grounded, reference-free
image--text evaluator that derives a global score from localized atomic
verification, together with \textsc{AGITA-Bench} for evaluating global and
fine-grained human agreement. Across established benchmarks, controlled
VLM-as-a-Judge experiments, and human evaluation, structured
decomposed prompting improves average performance across the tested backbones,
while task-specific training further strengthens sensitivity to semantic errors
and produces grounded explanations approaching those of proprietary judges.
Overall, our results show that compact, open evaluators can combine strong
alignment judgment with interpretable evidence and reproducible local
inference.

\textsc{VisExMEM} remains limited by errors in fine-grained verification and
evidence localization, while its multi-stage inference is still more expensive
than lightweight embedding-based metrics. Our prompting experiments also do
not exhaust alternatives such as self- or other-feedback, and AGITA-Bench
focuses on compositional semantic alignment rather than perceptual, aesthetic,
or domain-specific quality. Future work should improve binding-sensitive
verification and grounding, reduce inference cost, explore stronger
verification strategies, and extend atomic evaluation to broader domains and
modalities. More broadly, these directions could support increasingly capable,
open, and auditable evaluators that reduce reliance on proprietary VLM judges.

%% file: strucuture/ai_ethich.tex
\subsection*{AI use statement}

We used large language models for literature retrieval and discovery, improving
the clarity and presentation of the manuscript, and providing feedback during
research development. We also used Claude Code for implementation, debugging,
and routine engineering tasks. Model-based components used in the research,
including the offline teacher for training-data construction and model-assisted
AGITA-Bench decomposition, are described in the paper. All research questions,
methodological choices, experimental design, training objectives, evaluation
protocols, analyses, and conclusions were determined by the authors.
AI-generated suggestions, code, and text were reviewed, tested, and revised
before use. We take full responsibility for the work presented in this paper.

\subsection*{Ethics statement}

\textsc{VisExMEM} is designed to evaluate semantic alignment between images
and text. Its scope does not extend to image quality, social acceptability, or
the correctness of culturally or demographically sensitive representations.
Like other learned multimodal evaluators, its behavior can reflect biases and
representation gaps in the backbone models, grounding model, teacher models,
and training data. This is especially relevant for content involving people,
demographic attributes, or culturally specific concepts. Our
out-of-distribution experiments on CulturalFrames and ProtoBias therefore
characterize metric behavior under distribution shifts, with conclusions
limited to the evaluated benchmarks. Localized visual evidence and
natural-language explanations make predictions easier to inspect, but they do
not guarantee correctness. Errors can arise during decomposition, grounding,
or claim verification, so consequential uses should include independent human
review and other sources of evidence.

Human annotation is used in AGITA-Bench and in the explanation and evidence
evaluation. These tasks focus on image-text alignment, decomposition quality,
explanation quality, and visual evidence. In the explanation study, system
identities are hidden and outputs are shown under randomized labels to reduce
presentation bias. Annotator demographics are reported only in aggregate.
Our experiments rely mainly on existing public research datasets, whose
collection procedures, licensing conditions, geographic and cultural
coverage, and representation of people and concepts can affect the resulting
evaluation. We therefore interpret the reported results within the scope of
the datasets and benchmarks used in this study.

%% file: strucuture/appendix.tex
\subsection{Training Data Construction}
\label{app:training-data}

\paragraph{Data sources.}
Table~\ref{tab:training_sources} summarizes the public datasets used to
construct the training corpus and the supervision extracted from each source.
We combine dense region annotations, grounded captions, spatial-relation
labels, count annotations, natural long-form descriptions, human judgments,
and contrastive image pairs to obtain complementary coverage of the semantic
skills required by \textsc{VisExMEM}.

\paragraph{Atomic claim construction.}
We represent each caption as a set of typed, visually verifiable claims drawn
from six categories: object, attribute, action, relation, count, and negation.
When the source dataset provides explicit structure--e.g., Visual Genome
objects and attributes, Flickr30k phrase--region correspondences, VSR spatial
labels, or PixMo-Count annotations--we convert this supervision
programmatically. For sources dominated by free-form descriptions, including
DOCCI, CROC, and Localized Narratives, we use
\texttt{Qwen3.6-27B}\footnote{
\href{https://openai.com/index/gpt-5-6/}{GPT-5.6 Sol},
\href{https://qwen.ai/blog?id=qwen3.6-27b}{Qwen3.6-27B}
} as an offline decomposition teacher. The teacher is
not used to overwrite reliable source annotations.

\paragraph{Atomic verification.}
Each claim receives a supported or contradicted decision. Labels are derived
preferentially from native dataset annotations or human judgments. When a
reliable decision cannot be obtained directly from the source, the offline
teacher may provide a missing decision or structural link; low-confidence or
unresolved outputs are discarded. Thus, teacher supervision supplements rather
than replaces source-grounded labels.

\paragraph{Counterfactual supervision.}
We construct counterfactual examples through two complementary mechanisms.
First, CROC provides image-side contrastive pairs in which the same textual
description is evaluated against independently generated images with known
match labels. Second, for verified positive claims on natural images, we
programmatically create text-side counterfactuals by modifying a single
semantic constraint---object, attribute, action, relation, count, or
negation---while preserving the image. This produces minimal pairs in which
the intended semantic change is explicitly known.

\paragraph{Visual evidence.}
Evidence is taken directly from native annotations whenever available.
For claims without native bounding boxes, we use GroundingDINO to localize
the already-labeled semantic content. GroundingDINO therefore serves only as
an evidence-localization mechanism and never provides the Yes/No correctness
label. A failed detection is treated as a localization failure rather than
evidence that the claim is absent. Evidence is represented according to claim
type: object and attribute claims use the relevant regions, relations and
actions may involve multiple participating regions, and counting claims retain
the set of candidate instances. Pure absence claims are not assigned
fabricated evidence regions.

\paragraph{Causal-evidence construction.}
To distinguish genuine visual evidence from arbitrary image regions, each
eligible evidence region $E_i$ is paired with a random control region $R_i$
from the same image. These selected-versus-random pairs provide the controlled
interventions used by the Stage~3 causal-evidence objective. Exact random-region
matching and image-view construction are described in
Appendix~\ref{app:gdpo_details}.

\paragraph{Explanation supervision.}
Explanation supervision is constructed only after the atomic claims,
decisions, and evidence have been fixed. Explanation targets are generated
from these structured records rather than used to determine them. In
particular, the explanation teachers do not infer new bounding boxes or
redefine correctness labels. This separation ensures that explanation
learning cannot modify the supervision from which the numerical metric score
is derived.

\paragraph{Processed training examples.}
Figure~\ref{fig:processed_training_examples} shows representative
trainer-facing records after preprocessing. Stage~1 uses separate SFT views
for decomposition, verification, evidence selection, and explanation.
GroundingDINO boxes are supplied as candidate evidence regions; the
evidence-selection task predicts candidate IDs rather than box coordinates.
Stage~2 uses paired supported and counterfactual claims and optimizes their
decision margins without a generated target sequence. Stage~3 combines
semantic examples, where groups of sampled completions are scored by
Yes/No correctness, with causal-evidence examples that compare selected
evidence against a matched random region using multiple image views.
Stage~3 also replays earlier SFT and counterfactual examples. The examples
below are taken from the processed training files; unused bookkeeping fields
are omitted and long coordinates are rounded only for readability.
\input{figures/appendix_training_data}

\paragraph{Data isolation.}
All final evaluation benchmarks are excluded from the training corpus and
model-selection data. Training and development groups are separated before
optimization to prevent the same underlying example from appearing across
splits. Only the training split of EvalMuse is used for training.

\begin{table}[t]
\centering
\small
\setlength{\tabcolsep}{3.5pt}
\begin{tabular}{p{0.23\linewidth} p{0.42\linewidth} p{0.27\linewidth}}
\toprule
\textbf{Source} & \textbf{Supervision used} & \textbf{Primary skills} \\
\midrule
MS COCO &
Captions and object boxes &
Object, count \\

Visual Genome &
Region descriptions, object boxes, attributes &
Object, attribute, relation \\

VSR &
Image--caption pairs with spatial-relation labels &
Spatial relation \\

Flickr30k Entities &
Phrase--entity--region correspondences &
Object, grounding \\

Open Images + Localized Narratives &
Grounded free-form narratives &
Object, attribute, action, relation \\

DOCCI &
Detailed connected and contrasting descriptions &
Attribute, relation, negation \\

PixMo-Count &
Exact counts with point-level evidence &
Count \\

CROCsyn &
Contrastive image pairs with match labels &
Counterfactual verification \\

EvalMuse-40K &
Element-level human judgments from the training split &
Atomic verification \\
\bottomrule
\end{tabular}
\caption{Public data sources used to construct the \textsc{VisExMEM}
training corpus.}
\label{tab:training_sources}
\end{table}

\subsection{Training and Implementation Details}
\label{app:method}

\subsubsection{Models and Parameter-Efficient Training}
\label{app:models}

We train Qwen3.5-4B and Qwen3.5-9B using LoRA~\citep{hu2022lora} with rank $16$, scaling
factor $\alpha=32$, dropout $0.05$, and no bias. LoRA is applied to the
$q$, $k$, $v$, $o$, gate, up, and down projections. All training uses
bfloat16 precision and disables the Qwen3.5 thinking mode. The main training hyperparameters are summarized in
Table~\ref{tab:training_hyperparameters}; construction of the underlying
training corpus is described separately in Appendix~\ref{app:training-data}.

\subsubsection{Verification Prompt and Decision Scoring}
\label{app:verification_training}

Counterfactual refinement and Stage~3 use the same verification prompt
template to avoid prompt drift between supervised and policy optimization.
Given an image, caption, claim type, and atomic claim, the model is instructed
to return exactly one JSON decision,
\texttt{\{"decision":"yes"\}} or
\texttt{\{"decision":"no"\}}. Outputs that do not produce either valid
decision are treated as invalid during Stage~3 policy optimization.

During counterfactual refinement, no free-form generation is performed.
For each supported/counterfactual pair, the same image is evaluated with the
corresponding claim, and the verification margin in
Equation~\ref{eq:verification_margin} is obtained directly from the
\texttt{yes} and \texttt{no} decision logits after their shared output
prefix. In Stage~3 semantic optimization, the same prompt is instead used to
sample $G=8$ decision completions. For causal-evidence training, the textual
prompt is held fixed while only the image view changes. Thus, differences
between causal views arise from the image intervention rather than from
changes to the textual instruction.

\subsubsection{Multi-task SFT}
\label{app:sft_details}

Stage~1 uses a four-part curriculum over decomposition, verification,
evidence selection, and joint supervision. The task-specific views contain
13.5K decomposition, 15K verification, and 7K evidence-selection examples.
The final joint stage contains 40K examples, including 4.5K explanation
examples, and samples decomposition, verification, evidence-selection, and
explanation tasks using an $8{:}9{:}4{:}3$ schedule. Each curriculum stage
is traversed once.

\paragraph{SFT target formats.}
Each SFT example has a single task-specific JSON target $y$, and loss is
computed only over the target tokens using the next-token objective in
Equation~\ref{eq:sft}. Table~\ref{tab:sft_targets} summarizes the exact
target formats. Candidate bounding boxes are provided only as inputs for
evidence selection and are never prediction targets.

\begin{table}[t]
\centering
\small
\setlength{\tabcolsep}{3pt}
\renewcommand{\arraystretch}{1.08}
\begin{tabular}{@{}p{0.18\linewidth}p{0.74\linewidth}@{}}
\toprule
\textbf{Task} & \textbf{Target format} \\
\midrule

Decomposition &
\texttt{\{"parts": [...] \}}; each part contains
\texttt{part\_id}, \texttt{text}, \texttt{type},
\texttt{parent\_part\_id}, \texttt{head\_part\_id}, and
\texttt{target\_part\_id}. \\

Verification &
\texttt{\{"decision": "yes"\}} or
\texttt{\{"decision": "no"\}}. \\

Evidence selection &
\texttt{\{"selected\_evidence\_ids": ["E1", ...]\}}. \\

Explanation &
\texttt{\{"explanation": "..."\}}. \\

\bottomrule
\end{tabular}
\caption{Task-specific target formats used during Stage~1 SFT. Candidate
bounding boxes are inputs to evidence selection and are not prediction
targets.}
\label{tab:sft_targets}
\end{table}

\paragraph{Optimization.}
We use AdamW with cosine learning-rate decay, $3\%$ warmup, and gradient
clipping at $1.0$. The learning rate is $10^{-5}$ for the first three
task-specific curriculum stages and $7\times10^{-6}$ for final joint
training.

\subsubsection{Counterfactual Refinement}
\label{app:cf_details}

Stage~2 starts from the promoted Stage~1 checkpoint. The source pool contains
20K supported/counterfactual pairs. Training interleaves counterfactual
updates, verification replay, and decomposition replay in an
$8{:}1{:}1$ update ratio.

For pair $i$, let
$m_i^{+}=m_\theta(I_i,a_i^{+})$ and
$m_i^{-}=m_\theta(I_i,a_i^{-})$
denote the verification margins from
Equation~\ref{eq:verification_margin} for the supported and counterfactual
claims, respectively. Let $m_{i,\mathrm{ref}}^{+}$ denote the positive
margin produced by the frozen pre-Stage-2 checkpoint. We define the retained
positive target as
\begin{equation}
\bar m_i^{+}
=
\max\!\left(
\tau_{+},
m_{i,\mathrm{ref}}^{+}-\delta_{\mathrm{keep}}
\right),
\label{eq:cf_positive_target}
\end{equation}
and optimize
\begin{equation}
\mathcal{L}_{\mathrm{CF}}
=
\mathbb{E}_{i}\!\left[
\lambda_{\mathrm{pair}}
\left[
\gamma_{\mathrm{cf}}-(m_i^{+}-m_i^{-})
\right]_{+}
+
\lambda_{+}
\left[
\bar m_i^{+}-m_i^{+}
\right]_{+}
+
\lambda_{-}
\left[
m_i^{-}+\tau_{-}
\right]_{+}
\right],
\label{eq:counterfactual}
\end{equation}
where $[z]_{+}=\max(0,z)$. The three terms respectively enforce separation
between the supported and counterfactual claims, preserve sufficient support
for the positive claim, and push the counterfactual below the negative
target.

We use
$\gamma_{\mathrm{cf}}=3.0$,
$\tau_{+}=3.0$,
$\tau_{-}=2.0$,
$\delta_{\mathrm{keep}}=1.0$,
$\lambda_{\mathrm{pair}}=1.0$, and
$\lambda_{+}=\lambda_{-}=0.5$.
Thus, the positive target is
$\max(3.0,m_{i,\mathrm{ref}}^{+}-1.0)$ and the negative target is $-2.0$.

Verification replay uses its original binary verification objective together
with the retention constraint, whereas decomposition replay retains the
Stage~1 token-level sequence objective. We train for one counterfactual epoch
(450 optimizer steps) with AdamW, learning rate $5\times10^{-7}$,
$4\%$ warmup, cosine decay to a $10\%$ learning-rate floor, and gradient
clipping at $1.0$.

\subsubsection{Evidence-aware GDPO}
\label{app:gdpo_details}

Stage~3 starts from the promoted counterfactual checkpoint and uses 20K
semantic-verification examples together with 20K supported causal-evidence
examples. Our final Stage~3 procedure is an evidence-aware GDPO variant:
semantic correctness supplies the group-relative rollout signal, while
causal evidence is optimized directly as an auxiliary objective. In
particular, the causal signal is not treated as a second independently
group-normalized rollout reward; this distinction is important because the
image-intervention signal is shared across rollouts of the same input and
would otherwise be attenuated by within-group centering.

\paragraph{Semantic policy optimization.}
For each semantic prompt, we sample $G=8$ completions with temperature
$0.7$, top-$p=1.0$, and a maximum of 16 generated tokens. Correct,
incorrect, and invalid decisions receive rewards $+1$, $-1$, and $-1.25$,
respectively.

The group-relative quantities in
Equation~\ref{eq:gdpo_advantage_ratio} are implemented with
\begin{equation}
A_{i,g}
=
\frac{r_{i,g}-\mu_i}
     {\sigma_i+\epsilon_A},
\qquad
\epsilon_A=10^{-4},
\label{eq:gdpo_advantage_impl}
\end{equation}
where $\mu_i$ and $\sigma_i$ are the mean and population standard deviation
of the $G$ rollout rewards for semantic example $i$. If all rewards in a
group are identical, every centered reward is zero and the group therefore
provides no relative policy signal.

For the clipped semantic objective in Equation~\ref{eq:gdpo}, we use
$\epsilon=0.2$. KL regularization uses $\beta=0.02$ with the frozen Stage~2
checkpoint as the reference policy. For a generated token, the token-level
KL penalty is
\begin{equation}
d^{\mathrm{KL}}
=
\exp(\kappa)-\kappa-1,
\qquad
\kappa
=
\log\pi_{\mathrm{ref}}
-
\log\pi_{\theta}.
\label{eq:gdpo_kl}
\end{equation}

\paragraph{Replay schedule.}
Stage~3 interleaves semantic GDPO updates, counterfactual replay, and SFT
replay in a $7{:}2{:}1$ update ratio. Replay draws from the earlier training
pools and introduces no additional training examples. Counterfactual replay
therefore retains the Stage~2 margin objective in
Equation~\ref{eq:counterfactual}, while SFT replay retains the original
sequence-supervision objective in Equation~\ref{eq:sft}.

\paragraph{Five-view causal construction.}
For causal example $i$, let $E_i$ denote the selected evidence region and
$R_i$ its matched random control. We construct five image views:
the intact image $I$,
selected-evidence removal $I\setminus E_i$,
random-control removal $I\setminus R_i$,
selected-evidence-only $I\odot E_i$, and
random-control-only $I\odot R_i$.
Here $\setminus$ masks the specified region while retaining the rest of the
image, whereas $\odot$ masks the remaining canvas and retains only the
specified region. The masking implementation uses either Gaussian blur or
local-mean filling.

Let $s_i(V)$ denote the \texttt{yes}--\texttt{no} verification margin for
claim $i$ under image view $V$. We define
\begin{equation}
\begin{aligned}
\Delta_i^{\mathrm{sel}}
&=
s_i(I)-s_i(I\setminus E_i),
&
\Delta_i^{\mathrm{rnd}}
&=
s_i(I)-s_i(I\setminus R_i),
\\
\Delta_i^{\mathrm{suff}}
&=
s_i(I\odot E_i)-s_i(I\odot R_i),
&
\Delta_i^{\mathrm{anchor}}
&=
s_i(I)-s_i(I\odot E_i).
\end{aligned}
\label{eq:causal_deltas}
\end{equation}
The first two quantities measure the effect of removing selected versus
random regions, $\Delta_i^{\mathrm{suff}}$ compares their evidence-only
sufficiency, and $\Delta_i^{\mathrm{anchor}}$ measures the drop from the
intact image to the selected-evidence-only view.

\paragraph{Causal-evidence objective.}
Using the quantities in Equation~\ref{eq:causal_deltas}, we optimize
\begin{equation}
\begin{aligned}
\mathcal{L}_{\mathrm{causal}}(u)
=
\lambda_{\mathrm{ramp}}(u)\big[
&
w_{\mathrm{comp}}
\operatorname{softplus}
\!\left(
\gamma_{\mathrm{comp}}
-
(\Delta_i^{\mathrm{sel}}-\Delta_i^{\mathrm{rnd}})
\right)
\\[-1mm]
&+
w_{\mathrm{drop}}
\operatorname{softplus}
\!\left(
\gamma_{\mathrm{drop}}
-
\Delta_i^{\mathrm{sel}}
\right)
\\[-1mm]
&+
w_{\mathrm{suff}}
\operatorname{softplus}
\!\left(
\gamma_{\mathrm{suff}}
-
\Delta_i^{\mathrm{suff}}
\right)
\\[-1mm]
&+
w_{\mathrm{suffdrop}}
\operatorname{softplus}
\!\left(
\Delta_i^{\mathrm{anchor}}
-
\gamma_{\max}
\right)
\big].
\end{aligned}
\label{eq:causal-loss}
\end{equation}
The four terms respectively encourage
(i) a larger removal effect for selected than random evidence,
(ii) sensitivity to removing selected evidence,
(iii) greater support from selected than random evidence-only views, and
(iv) a bounded degradation when verification is restricted to selected
evidence alone.

We use
$\gamma_{\mathrm{comp}}=0.25$,
$\gamma_{\mathrm{drop}}=0.25$,
$\gamma_{\mathrm{suff}}=0.75$, and
$\gamma_{\max}=1.0$, with
\[
(w_{\mathrm{comp}},
w_{\mathrm{drop}},
w_{\mathrm{suff}},
w_{\mathrm{suffdrop}})
=
(1.0,0.5,1.0,0.25).
\]
The outer causal coefficient in the Stage~3 objective is
$\lambda_{\mathrm{c}}=0.3$. The internal coefficient
$\lambda_{\mathrm{ramp}}(u)$ increases linearly from $0$ to $0.5$ over the
first 100 completed causal updates and remains at $0.5$ thereafter. The
maximum resulting multiplier on the weighted causal auxiliary is therefore
$0.3\times0.5=0.15$.

\paragraph{Eligibility and random-control matching.}
A causal example contributes to $\mathcal{L}_{\mathrm{causal}}$ only when
the source image and all five required views can be constructed. Examples
that fail this requirement are excluded from the causal auxiliary; if no
eligible causal examples remain in an update, the causal term is omitted for
that update.

Each selected evidence region $E_i$ is paired with a random control region
$R_i$ from the same image. The control is matched to $E_i$ within
$\pm15\%$ in area and $\pm20\%$ in aspect ratio while satisfying
$\operatorname{IoU}(R_i,E)<0.10$ for every evidence region $E$ in that
image. We allow at most 200 sampling attempts.

\paragraph{Joint Stage~3 objective and optimization.}
The final Stage~3 update combines the semantic objective from
Equation~\ref{eq:gdpo} with the direct causal auxiliary from
Equation~\ref{eq:causal-loss}:
\begin{equation}
\mathcal{L}_{\mathrm{Stage3}}
=
-\mathcal{J}_{\mathrm{sem}}
+
\lambda_{\mathrm c}
\mathcal{L}_{\mathrm{causal}}.
\label{eq:stage3_training}
\end{equation}
The causal term is optimized directly rather than incorporated into the
group-normalized rollout reward, preserving the image-intervention signal
while semantic correctness is learned through the rollout-based policy
objective.

Stage~3 uses AdamW with learning rate $10^{-6}$, cosine decay,
$3\%$ warmup, zero weight decay, and gradient clipping at $1.0$.
The sequential Verification-GRPO and Causal-GRPO alternatives used in the
training ablations are described separately in Appendix~\ref{app:grpo};
they branch from the Stage~2 checkpoint and are not additional stages of the
final \textsc{VisExMEM} model.

\begin{table}[t]
\centering
\small
\setlength{\tabcolsep}{3.2pt}
\renewcommand{\arraystretch}{1.08}
\caption{Training hyperparameters for the 4B and 9B \textsc{VisExMEM}
models. For SFT, $4/2$ denotes per-device batch size $4$ for the
decomposition curriculum stage and $2$ for the remaining stages; the two
learning rates correspond to the first three task-specific curriculum stages
and the final joint stage, respectively. Replay/update mixtures denote
counterfactual:verification:decomposition for Stage~2 and
semantic-GDPO:counterfactual:SFT for Stage~3.}
\label{tab:training_hyperparameters}

\resizebox{\textwidth}{!}{%
\begin{tabular}{lcccccc}
\toprule
&
\multicolumn{2}{c}{\textbf{SFT}}
&
\multicolumn{2}{c}{\textbf{Counterfactual}}
&
\multicolumn{2}{c}{\textbf{Evidence-aware GDPO}}
\\
\cmidrule(lr){2-3}
\cmidrule(lr){4-5}
\cmidrule(lr){6-7}
\textbf{Hyperparameter}
& \textbf{4B} & \textbf{9B}
& \textbf{4B} & \textbf{9B}
& \textbf{4B} & \textbf{9B}
\\
\midrule

Training duration (hours)
& 4.4 & 4.6
& 2.7 & 3.4
& 40.9 & 43.3 \\

GPUs
& $4\times$ H100 & $4\times$ H100
& $4\times$ H100 & $4\times$ H100
& $2\times$ H100 & $2\times$ H100 \\

Per-device batch size
& $4/2$ & $4/2$
& 1 & 1
& 2 & 2 \\

Gradient accumulation
& 8 & 8
& 8 & 8
& 2 & 2 \\

Learning rate
& $10^{-5}/7{\times}10^{-6}$ & $10^{-5}/7{\times}10^{-6}$
& $5{\times}10^{-7}$ & $5{\times}10^{-7}$
& $10^{-6}$ & $10^{-6}$ \\

Learning-rate scheduler
& cosine & cosine
& cosine & cosine
& cosine & cosine \\

Warmup
& $3\%$ & $3\%$
& $4\%$ & $4\%$
& $3\%$ & $3\%$ \\

Weight decay
& 0.0 & 0.0
& 0.01 & 0.01
& 0.0 & 0.0 \\

Gradient clipping
& 1.0 & 1.0
& 1.0 & 1.0
& 1.0 & 1.0 \\

Replay/update mixture
& -- & --
& $8{:}1{:}1$ & $8{:}1{:}1$
& $7{:}2{:}1$ & $7{:}2{:}1$ \\

Generations per prompt
& -- & --
& -- & --
& 8 & 8 \\

\bottomrule
\end{tabular}%
}
\end{table}

\subsection{GRPO Ablations}
\label{app:grpo}

\paragraph{Ablation strategy.}
We evaluate a sequential GRPO alternative to the final
\textsc{VisExMEM} training strategy. The ablation branch starts from the
counterfactually refined Stage~2 checkpoint:
\[
\text{Counterfactual}
\rightarrow
\text{Verification-GRPO}
\rightarrow
\text{Causal-GRPO}.
\]
In contrast, the final evidence-aware GDPO stage branches directly from the
counterfactual checkpoint. Thus, Verification-GRPO and Causal-GRPO are
controlled ablations rather than additional stages of the deployed model.

\paragraph{Verification-GRPO.}
Verification-GRPO isolates reinforcement learning for semantic correctness.
For each image--claim input, we sample $G=8$ completions and assign reward
$+1$ to a correct \texttt{yes}/\texttt{no} decision, $-1$ to an incorrect
valid decision, and $-1.25$ to an invalid output. Rewards are normalized
within each rollout group as
\begin{equation}
A_{i,g}
=
\frac{r_{i,g}-\mu_i}{\sigma_i+\epsilon_A},
\qquad
\epsilon_A=10^{-4},
\label{eq:grpo_adv}
\end{equation}
where $\mu_i$ and $\sigma_i$ are the mean and population standard deviation
of the $G$ rollout rewards. Groups with identical rewards therefore provide
no relative policy signal. We then apply the same clipped policy objective
and KL regularization described in Equation~\ref{eq:gdpo}. Unlike Stage~2,
which directly optimizes verification margins, Verification-GRPO learns from
relative differences between sampled decisions.

\paragraph{Causal-GRPO.}
Causal-GRPO continues from Verification-GRPO and tests whether visual
evidence dependence can be incorporated through the rollout reward itself.
It uses the same GRPO formulation, but for a correct supported decision adds
a causal bonus derived from the difference between masking selected evidence
and a matched random region:
\begin{equation}
r_{i,g}^{\mathrm{causal}}
=
r_{i,g}^{\mathrm{sem}}
+
\alpha\,
\operatorname{clip}(e_i,0,1),
\qquad
\alpha=0.3,
\label{eq:causal_grpo_reward}
\end{equation}
where $e_i$ denotes the selected-versus-random evidence advantage.
This ablation therefore tests a \emph{reward-based} treatment of causal
evidence, whereas the final model optimizes evidence dependence directly.

\paragraph{Replay.}
Both GRPO ablations interleave decomposition examples from the earlier SFT
stage to preserve the structured evaluation interface during policy
optimization. This differs from GRPO experience replay
\citep{yoo2026rolloutleveladvantageprioritizedexperiencereplay}: we do not
store or reuse previous policy rollouts. Instead, earlier supervised examples
are revisited with their original training objective.

The final GDPO stage uses the same retention principle more broadly,
interleaving semantic policy updates, counterfactual replay, and SFT replay.
The contribution of this replay strategy is evaluated separately in the
training ablations.

\paragraph{Relation to the final GDPO objective.}
The sequential GRPO branch separates semantic verification and causal
evidence into two successive policy-optimization stages. Our final model
instead combines them in a single Stage~3 optimization. Semantic verification
provides the rollout-based policy signal, while the five-view causal objective
from Equation~\ref{eq:causal-loss} is optimized directly:
\begin{equation}
\mathcal{L}_{\mathrm{Stage3}}
=
-\mathcal{J}_{\mathrm{sem}}
+
\lambda_{\mathrm{c}}\mathcal{L}_{\mathrm{causal}}.
\label{eq:stage3_appendix}
\end{equation}
This distinction is important because a causal signal shared across all
rollouts of the same input is largely removed by within-group centering and
therefore provides little relative GRPO signal. Direct optimization preserves
the image-intervention signal while allowing semantic correctness and
evidence dependence to be learned jointly.

\subsection{Additional Results}
\label{app:additional_results}

\subsubsection{Out-of-Distribution Generalization}
\label{app:ood_generalization}

We further evaluate \textsc{VisExMEM} on three external benchmarks that are
not used in training, covering structured visualizations, culturally grounded
image--text alignment, and prototypicality-sensitive semantic evaluation.

\textbf{VisPlotBench}~\citep{ni2025viscoder2}, introduced with VisCoder2,
provides a structured-visual distribution based on rendered visualization
tasks. We evaluate 886 examples and report pairwise accuracy (PAcc) and the
stricter Group score. We additionally include
\textsc{ChartVE}~\citep{huang2023chartve}, a visual-entailment model developed
for factual-consistency evaluation of chart captions.

\textbf{CulturalFrames}~\citep{nayak2025culturalframes} evaluates alignment
with explicit and implicit cultural expectations across countries and
socio-cultural domains. We evaluate 3,577 images and report Spearman's
$\rho$ with human alignment judgments and within-prompt pairwise accuracy
(PAcc$_{\mathrm{wp}}$). We also include the culturally targeted
AHEaD metrics~\citep{malakouti2025culture}: ALIGN measures coverage of
expected cultural descriptors, while HAL measures unsupported cultural
content. We report $1-\mathrm{HAL}$ so that higher is better throughout.

\textbf{ProtoBias}~\citep{roy2026prototypicalitybiasrevealsblindspots} tests whether evaluation metrics
prefer semantically correct images over more prototypical alternatives
containing controlled semantic errors. We evaluate its 600 human-annotated
examples, with 200 each from Animals, Demography, and Objects, and report
category-level accuracy and Cohen's $\kappa$. The GPT-5 (Judge) and
\textsc{ProtoScore} results are taken directly from the original ProtoBias
study~\citep{roy2026prototypicalitybiasrevealsblindspots} and are marked with $^\dagger$ in the table;
all remaining values are obtained under our evaluation setup.

\input{figures/app_ood_results}

\paragraph{Results.}
The OOD results show that \textsc{VisExMEM}-specific training provides
substantial gains over the corresponding untuned Qwen3.5 judges.
On VisPlotBench, \textsc{VisExMEM}-4B improves over both 4B controls and
reaches 91.3 PAcc and 74.4 Group, while \textsc{VisExMEM}-9B reaches
91.3 PAcc and the highest Group score of 78.6, compared with 74.5 for the
strongest 9B Qwen control. On CulturalFrames, the gain is clearest at 9B:
\textsc{VisExMEM}-9B reaches $\rho=.311$ and 65.8
PAcc$_{\mathrm{wp}}$, exceeding both corresponding Qwen3.5 controls and
all conventional metrics; the strongest conventional baseline,
VQAScore, obtains $\rho=.250$ and 64.4 PAcc$_{\mathrm{wp}}$.

The same pattern appears on ProtoBias. \textsc{VisExMEM}-9B substantially
improves over both 9B Qwen controls across all three categories, reaching
73.5/71.5/81.0 accuracy and .254/.205/.286 $\kappa$ on
Animals/Demography/Objects. \textsc{VisExMEM}-4B likewise improves strongly
over its Qwen baselines in category accuracy. Among the conventional
general-purpose metrics, the \textsc{VisExMEM} variants obtain the strongest
ProtoBias results. \textsc{ProtoScore} achieves higher category accuracy,
as expected for a model trained specifically for the prototypicality-bias
setting, while \textsc{VisExMEM}-9B remains stronger in $\kappa$ on Animals
and Objects despite receiving no ProtoBias-specific training.

\subsubsection{Replay Ablation}
\label{app:replay_ablation}

We examine the role of replay during progressive training using
\textsc{VisExMEM}-9B. Both variants start from the same base model and SFT
checkpoint. We then train two complete trajectories: the standard trajectory
replays earlier-stage supervision during both counterfactual refinement (CF)
and evidence-aware GDPO, whereas the no-replay trajectory disables replay in
both stages. Table~\ref{tab:replay_ablation} reports the resulting checkpoints
on the same five benchmarks used in the main training-stage ablation.

\begin{table}[t]
\centering
\caption{
\textbf{Replay ablation for \textsc{VisExMEM}-9B.}
Both paths start from the same SFT checkpoint and either retain or disable
replay throughout CF and GDPO. $\Delta$ denotes Replay $-$ No replay.
Higher is better.
}
\label{tab:replay_ablation}

\fontsize{6.3}{6.7}\selectfont
\setlength{\tabcolsep}{1.5pt}
\renewcommand{\arraystretch}{0.90}

\resizebox{0.56\linewidth}{!}{%
\begin{tabular}{@{}llccccc@{}}
\toprule

\textbf{Variant}
& \textbf{Replay}
& \textbf{WG}
& \textbf{EqB}
& \textbf{TIFA}
& \textbf{Pick}
& \textbf{GenAI}
\\
\midrule


Base (Hol.)
& --
& 40.4 & 35.7 & 50.1 & 74.2 & 50.3
\\

\quad + SFT
& --
& 61.9 & 58.9 & 59.8 & 61.8 & 57.2
\\

\midrule


\rowcolor{blue!6}
\qquad + CF
& \textcolor{blue!70!black}{\textbf{Yes}}
& 66.3
& 64.9
& 68.9
& 77.5
& 67.2
\\

\rowcolor{purple!7}
\qquad + CF
& \textcolor{purple!70!black}{\textbf{No}}
& \textbf{67.9}
& \textbf{66.2}
& \textbf{71.1}
& \textbf{81.1}
& \textbf{67.3}
\\

\rowcolor{gray!8}
\qquad\textit{$\Delta$}
& \textit{Yes $-$ No}
& \textcolor{red!70!black}{(-1.6)}
& \textcolor{red!70!black}{(-1.3)}
& \textcolor{red!70!black}{(-2.2)}
& \textcolor{red!70!black}{(-3.6)}
& \textcolor{red!70!black}{(-0.1)}
\\

\midrule


\rowcolor{blue!6}
\qquad\quad + GDPO
& \textcolor{blue!70!black}{\textbf{Yes}}
& \textbf{68.9}
& \textbf{70.1}
& \textbf{73.1}
& \textbf{85.6}
& \textbf{69.4}
\\

\rowcolor{purple!7}
\qquad\quad + GDPO
& \textcolor{purple!70!black}{\textbf{No}}
& 66.1
& 64.8
& 70.7
& 74.5
& 65.1
\\

\rowcolor{gray!8}
\qquad\quad\textit{$\Delta$}
& \textit{Yes $-$ No}
& \textcolor{green!45!black}{(+2.4)}
& \textcolor{green!45!black}{(+5.3)}
& \textcolor{green!45!black}{(+2.4)}
& \textcolor{green!45!black}{(+11.1)}
& \textcolor{green!45!black}{(+4.3)}
\\

\bottomrule
\end{tabular}%
}

\end{table}

At the CF checkpoint, disabling replay is consistently beneficial. The
no-replay variant improves over the replay variant on all five benchmarks,
raising the average from 69.0 to 70.7. The largest difference occurs on
Pick-a-Pic (81.1 vs.\ 77.5), while GenAI-Bench changes only marginally
(67.3 vs.\ 67.2). Thus, replay is not required for effective counterfactual
adaptation and slightly constrains the immediate CF checkpoint.

The behavior changes markedly after evidence-aware GDPO. Starting from the
stronger no-replay CF checkpoint, continuing training without replay reduces
the average from 70.7 to 68.2. In contrast, the replay trajectory improves
from 69.0 after CF to 73.3 after GDPO and outperforms the no-replay trajectory
on all five benchmarks. The largest final differences are observed on
Pick-a-Pic (+11.1) and EqBen (+5.3).

This reversal suggests that replay plays different roles across the two
training stages: it is less important for short-term counterfactual
adaptation, but becomes beneficial during policy optimization, where
retaining earlier structured and counterfactual supervision helps avoid
degradation while optimizing the new semantic-policy and causal-evidence
objectives. Importantly, the two GDPO checkpoints continue from their
respective replay and no-replay CF branches; therefore, this ablation
establishes the effect of the complete replay versus no-replay training
trajectories, rather than isolating replay within Stage~3 alone.

\subsubsection{Aggregation Ablation}
\label{app:aggregation}
We evaluate the sensitivity of \textsc{VisExMEM}-9B to the rule used to
aggregate claim-level support scores. Using the same claim predictions, we
compare the arithmetic mean with geometric, harmonic, and minimum aggregation.
As shown in Table~\ref{tab:aggregation_ablation}, the arithmetic mean performs
best on all five evaluated benchmarks, with the geometric mean as the closest
alternative. This result supports the arithmetic mean used in the main method,
which gives equal weight to all visually verifiable claims and requires no
learned or benchmark-specific aggregation parameters.

\begin{table}[t]
\centering
\caption{\textbf{Aggregation ablation.}
Performance of \textsc{VisExMEM}-9B under alternative fixed aggregation rules.
Higher is better.}
\label{tab:aggregation_ablation}

\setlength{\tabcolsep}{4.0pt}
\renewcommand{\arraystretch}{1.0}

\begin{tabular}{lccccc}
\toprule
\textbf{Aggregation}
& \textbf{WG}
& \textbf{EqBen}
& \textbf{TIFA160}
& \textbf{Pick-a-Pic}
& \textbf{GenAI}
\\
\midrule

\textbf{Arithmetic mean}
& \textbf{68.9}
& \textbf{70.1}
& \textbf{73.1}
& \textbf{85.6}
& \textbf{69.4}
\\

Geometric
& 68.4
& 69.6
& 71.6
& 84.4
& 68.7
\\

Harmonic
& 67.3
& 60.3
& 69.6
& 81.5
& 64.4
\\

Minimum
& 62.9
& 55.4
& 68.9
& 80.4
& 63.7
\\

\bottomrule
\end{tabular}
\end{table}

\subsection{Benchmark Evaluation Protocols}
\label{app:benchmark_protocols}

We use the same fixed \textsc{VisExMEM} inference procedure across all
benchmarks. The evaluation set and meta-evaluation statistic for each
benchmark are fixed in advance; no \textsc{VisExMEM} prompt, decision
threshold, or aggregation rule is tuned on benchmark test data. When an
evaluation statistic itself includes a calibration step, as in pairwise
accuracy with tie calibration, this is applied only at meta-evaluation time
to the already-computed metric scores and does not modify
\textsc{VisExMEM} inference.

\paragraph{Winoground and EqBen.}
Winoground~\citep{thrush2022winogroundprobingvisionlanguage} and
EqBen~\citep{wang2023equivariantsimilarityvisionlanguagefoundation} evaluate
fine-grained image--text matching using two matched image--caption pairs per
example. Following the evaluation setting used in our comparisons, we use
the complete 400-example Winoground set and the 280-example EqBen-Mini
subset. Let $(I_0,c_0)$ and $(I_1,c_1)$ denote the two matched pairs and
$S(I,c)$ the metric score. The Text condition is satisfied when each image
prefers its matched caption,
\[
S(I_0,c_0) > S(I_0,c_1)
\quad\text{and}\quad
S(I_1,c_1) > S(I_1,c_0),
\]
whereas the Image condition is satisfied when each caption prefers its
matched image,
\[
S(I_0,c_0) > S(I_1,c_0)
\quad\text{and}\quad
S(I_1,c_1) > S(I_0,c_1).
\]
The Group condition requires both the Text and Image conditions to hold.
We report all three scores where space permits and use Group for the main
comparison because it requires all four matching inequalities to be correct.

\paragraph{TIFA160 and GenAI-Bench.}
For TIFA160~\citep{hu2023tifaaccurateinterpretabletexttoimage}, we evaluate
the 800 image--text pairs formed from 160 prompts and outputs from five
text-to-image models, together with their 1--5 human alignment ratings.
For GenAI-Bench~\citep{li2024genaibenchevaluatingimprovingcompositional},
we use the 527-prompt image coreset used for human and automatic
meta-evaluation, rather than the full 1,600-prompt collection.

For both benchmarks, we report pairwise accuracy (PAcc) with tie calibration.
PAcc is computed from the vectors of human ratings and metric scores by
comparing their induced pairwise higher/lower/tie decisions; the metric's
tie threshold is calibrated according to the pairwise-accuracy protocol.
Thus, these benchmarks contain scalar human ratings rather than predefined
binary preference pairs, and PAcc measures agreement between the pairwise
orderings induced by the human and metric scores. Results are reported on
the $0$--$100$ scale.

\paragraph{Pick-a-Pic.}
For Pick-a-Pic~\citep{kirstain2023pickapicopendatasetuser}, we follow the
filtered evaluation subset used in the VQAScore protocol: 100 prompts, each
paired with two generated images, for a total of 200 images. Human-tied
examples are excluded from this subset. We report binary preference
accuracy, where a comparison is correct when the human-preferred image
receives the larger metric score for the corresponding prompt.

\paragraph{CROC$_{\mathrm{hum}}$.}
CROC$_{\mathrm{hum}}$~\citep{leiter2026crocevaluatingtrainingt2i} evaluates
contrastive robustness in four directions obtained by crossing
text-based versus image-based evaluation with forward versus inverse
evaluation. In text-based evaluation, the image is held fixed while the
original and contrastive texts are compared; in image-based evaluation, the
text is held fixed while original and contrastive images are compared.
Forward evaluation treats the original text--image pair as the matching
pair, whereas inverse evaluation treats the contrastive pair as the
matching pair. We compute all four direction scores using the benchmark
protocol and report their unweighted arithmetic mean as the single
CROC$_{\mathrm{hum}}$ score in the main comparison.

\paragraph{T2I-CompBench++.}
For
T2I-CompBench++~\citep{huang2025t2icompbenchenhancedcomprehensivebenchmark},
we use its released human-evaluation image--text examples and report
Spearman's rank correlation coefficient (SRCC) between automatic metric
scores and the corresponding human alignment judgments on the evaluation
subset used in our experiments. Spearman correlation evaluates agreement in
rank ordering and is invariant to monotonic rescaling of either score.
The identical sample set and correlation computation are used for every
method in our comparison.

\paragraph{CountBench.}
CountBench~\citep{paiss2023teachingclipcount} evaluates zero-shot visual
counting for counts from two through ten. For each image, the benchmark
constructs nine candidate captions by replacing the count in its caption
with each number in $\{\text{two},\ldots,\text{ten}\}$. We score all nine
image--caption combinations using the same metric inference procedure and
take the count appearing in the highest-scoring caption as the predicted
count. We report exact top-1 counting accuracy.

\paragraph{T2IScoreScore.}
T2IScoreScore (TS2)~\citep{saxon2024evaluatesevaluationsobjectivelyscoring}
meta-evaluates image--text metrics using \emph{semantic error graphs} (SEGs).
Each SEG associates a prompt with groups of images containing controlled
semantic-error configurations. We use the two principal TS2 statistics
reported in the main paper. The ordering score $\mathrm{rank}_m$ is the
benchmark's Spearman-based measure of whether metric scores correctly order
images along walks through an SEG according to semantic-error severity.
The separation score $\mathrm{sep}_m$ measures how well score distributions
for neighboring error nodes are distinguished using the two-sample
Kolmogorov--Smirnov statistic. Higher values indicate better performance for
both measures.

We report $\mathrm{rank}_m$ and $\mathrm{sep}_m$ over all evaluated SEGs
(\emph{All}) and separately for the \emph{Synth}, \emph{Nat}, and
\emph{Real} partitions. \emph{Synth} uses generated images to populate
predefined semantic error graphs, \emph{Nat} constructs error graphs from
natural images, and \emph{Real} contains naturally occurring semantic errors
from text-to-image generations. Here, ``Real'' refers to the errors being
naturally occurring rather than to the images being real photographs.
Reporting both ordering and separation distinguishes metrics that recover the
intended semantic ordering from those that also assign clearly separated
scores to different error populations.

\paragraph{AGITA-Bench.}
For AGITA-Bench (Sec.~\ref{sec:agita}), evaluation is performed separately at
the global and atomic levels. Global agreement is Spearman's $\rho$ between
each method's image--text score and the human global-alignment rating.
Atomic agreement is computed only for methods that can directly score the
fixed benchmark claims and is Spearman's $\rho$ between claim-level
metric scores and human atomic-support judgments; claims marked
\emph{cannot judge} are excluded. Correlations are reported separately for
the Easy, Medium, and Hard semantic-complexity subsets, with the reported
average equal to the unweighted mean of the three subset correlations.
The benchmark construction and full human-annotation protocol are provided
in Appendix~\ref{app:agita}.

\paragraph{GroundingDINO baseline.}
To test how much image-text alignment can be recovered from grounding alone,
we replace the \textsc{VisExMEM} verifier with
GroundingDINO-base~\citep{liu2024groundingdinomarryingdino}. Captions are
decomposed into atomic claims by the same frozen text-only
\textsc{VisExMEM} v2 decomposer, which receives neither images nor benchmark
labels. Each claim is converted to a lowercase, period-terminated query, and
GroundingDINO is the only component that receives image pixels. We use a box
threshold of $0.30$, text threshold of $0.25$, NMS IoU of $0.5$, and at most
32 candidate boxes per query. Claim support is based on the strongest
phrase-aligned confidence before thresholding, with the scoring rule determined
by claim type.

Object claims use the maximum confidence and negated claims use its complement.
Count claims combine the strongest detections with agreement between the
detected and expected counts. Attribute claims combine entity and attribute
confidence with box overlap, while spatial claims also use the relative
positions of subject and object boxes. Verb relations combine subject, object,
and phrase confidence, and unary actions combine subject and phrase confidence.
Depth relations and abstract claims are marked ungroundable because no reliable
2D box rule is defined. If $M$ of the $N$ claims have a defined score, the
final metric is

\[
S_{\mathrm{GDINO}}
=
\frac{M}{N}
\exp\left(
\frac{1}{M}
\sum_{i=1}^{M}
\log \max(s_i,10^{-6})
\right).
\]

The geometric mean makes low-support claims reduce the overall score, while
the coverage term $M/N$ penalizes captions for which only part of the content
can be grounded. All thresholds and constants are fixed before evaluation.
The baseline uses no training, calibration, or benchmark-specific tuning, and
GroundingDINO confidence is treated as an uncalibrated grounding support score
rather than a probability.

\subsubsection{VLM-as-a-Judge Evaluation}
\label{app:vlm_judge_protocol}

We evaluate untuned Qwen3.5-4B, Qwen3.5-9B, and GPT-5.6 Sol as
VLM-as-a-Judge controls under two fixed inference settings. Results on the
eight established benchmarks and AGITA-Bench are reported in
Tables~\ref{tab:main_results} and~\ref{tab:agita_results}, respectively.

\paragraph{Holistic (Hol.).}
The VLM receives the image and complete caption and directly assigns a single
alignment score using a fixed rubric from $0.05$ to $0.95$. No decomposition,
claim-level verification, or localized evidence is used. The complete prompt
is shown in Figure~\ref{fig:vlm_judge_holistic_prompt}.

\paragraph{Decomposed (Dec.).}
The same backbone is evaluated with a structured end-to-end
decompose--verify--aggregate procedure. The VLM receives the image and complete
caption and, in a single response, decomposes the caption into visually
verifiable atomic claims, localizes claim-level visual evidence, assigns each
claim a support score in $[0,1]$, and generates a short evidence-grounded
explanation. Possible claim types include \emph{object}, \emph{attribute},
\emph{count}, \emph{action}, \emph{relation}, \emph{spatial},
\emph{negation}, and \emph{other}. After evaluating all claims, the VLM
produces an overall score in $[0,1]$ based on the completed part-level
judgments. The overall score is generated by the VLM rather than computed
using an external aggregation formula. The complete prompt is shown in
Figure~\ref{fig:vlm_judge_decomposed_prompt}.

These controls use no \textsc{VisExMEM}-specific fine-tuning and do not use
\textsc{VisExMEM}'s learned evidence-conditioned verifier or external
grounding cues. Thus, within each backbone, Hol.\ versus Dec.\ changes the
inference structure while keeping the underlying VLM fixed.

\subsubsection{Baseline Coverage and Reproducibility}
\label{app:baseline_coverage}

REVEALER~\citep{shi2026revealerreinforcementguidedvisualreasoning} and T2I-FineEval~\citep{hosseini2025t2ifineeval} are particularly relevant
to \textsc{VisExMEM} because both perform fine-grained image--text evaluation
rather than relying only on a global similarity score. We therefore examined
whether they could be included in the controlled benchmark comparison.

\paragraph{REVEALER.}
REVEALER performs element-level alignment evaluation through a structured
grounding--reasoning--conclusion trajectory and trains its evaluator with
reinforcement learning over grounding and alignment objectives. At the time
of our evaluation, we were unable to obtain a usable public implementation
and released checkpoint sufficient to reproduce the published method on our
benchmark suite. Reimplementing the method from the paper would introduce
uncontrolled choices in model initialization, training data, reward
construction, and inference, and would therefore not constitute a faithful
baseline reproduction. We consequently do not report reconstructed REVEALER
numbers.

\paragraph{T2I-FineEval.}
T2I-FineEval decomposes both modalities: its reported configuration uses
GPT-4 with two-shot prompting to decompose the input text and generate
entity- and relation-level questions, YOLOv9 to obtain image regions, and
BLIP-VQA for question--region scoring. Although evaluation code is publicly
available, faithfully reproducing the published configuration therefore
requires the proprietary GPT-4 decomposition stage and its corresponding
prompting configuration. Replacing this stage with a different language
model would change the evaluated method and confound the comparison with
backbone and prompting differences. Because we could not reproduce the
published configuration exactly under our controlled evaluation setup, we
do not report a modified T2I-FineEval result.

We also avoid copying numbers reported in the original papers into our main
tables, since their evaluated subsets, scoring statistics, and/or underlying
models do not match our fixed protocols. All quantitative comparisons in
Tables~\ref{tab:main_results} and~\ref{tab:agita_results} therefore use
scores obtained under the evaluation settings specified in this paper.
\input{figures/prompts/hol_prompt}
\input{figures/prompts/dec_prompt}

\newpage
\raggedbottom

\setlength{\textfloatsep}{8pt plus 2pt minus 2pt}
\setlength{\floatsep}{6pt plus 2pt minus 2pt}

\subsection{AGITA-Bench}
\label{app:agita}

\subsubsection{Benchmark Construction}
\label{app:agita_construction}

AGITA-Bench contains 600 image-text pairs, split equally between I2T and
T2I evaluation. The 300 I2T cases use 100 real images each from
ADE20K~\citep{Zhou_2017_CVPR},
HICO-DET~\citep{chao2018learningdetecthumanobjectinteractions}, and
Objects365~\citep{Shao_2019_ICCV}. Their captions cover high, partial, and
low alignment. The 300 T2I cases use original generation prompts paired with
181 images from DiffusionDB~\citep{wang2023diffusiondblargescalepromptgallery}
and 119 from MJHQ-30K~\citep{li2024playgroundv25insightsenhancing}.

We construct 200 easy, 200 medium, and 200 hard cases based on
\emph{semantic complexity} rather than alignment severity. Fixed rules assign
difficulty using the number and composition of atomic claims. Easy cases
contain few simple claims. Medium cases add relations or attribute and count
assignments. Hard cases contain more constraints or require correct binding
of attributes, counts, roles, or spatial relations. The resulting benchmark
contains 8,735 atomic claims. AGITA-Bench is excluded from training,
calibration, and model selection.

\paragraph{Atomic decomposition prompt.}
AGITA-Bench captions are decomposed into typed visual semantic units using
GPT-5.6 Sol with a fixed caption-only prompt. The model does not receive the
image during decomposition. It is instructed to preserve entity and attribute
bindings, counts, actions, relations, spatial relations, and negation while
avoiding unobservable or stylistic content. The complete prompt is shown in
Figure~\ref{fig:agita_decomp_prompt}.

\subsubsection{Human Annotation Protocol}
\label{app:agita_annotation_protocol}

AGITA-Bench is annotated at two complementary levels. For each atomic claim,
annotators assign an image-support score from 0, meaning contradicted or not
supported, to 4, meaning fully supported. A separate \emph{cannot judge}
option is available when the visual evidence is insufficient for a reliable
decision. Each image-text pair is also rated for global alignment from 1,
meaning not aligned, to 5, meaning fully aligned. Decomposition quality is
judged independently of image support. A claim is \emph{reasonable as written}
when it is a clear and non-redundant atomic proposition that requires neither
splitting nor merging. This separates decomposition errors from genuine
image-text mismatches.

AGITA-Bench provides human score annotations rather than human bounding-box
annotations. Any localized evidence boxes reported by \textsc{VisExMEM} are
produced by the metric and are not part of the benchmark's human labels.

All seven annotators independently annotate the same 30 image-text pairs,
comprising 460 unique atomic claims, for inter-annotator agreement analysis.
The remaining 570 pairs are each assigned to one annotator. Pairwise
quadratic-weighted Cohen's $\kappa$ is reported in
Tables~\ref{tab:agita_iaa_global} and~\ref{tab:agita_iaa_atomic}.
Aggregate annotator demographics are reported in
Table~\ref{tab:agita_annotator_demographics}.

\subsubsection{Annotation Interface, Guidelines and Examples}
\label{app:agita_interface}

\begin{figure}[t]
    \centering

    \begin{minipage}{\linewidth}
        \centering
        \includegraphics[
            angle=270,
            width=.90\linewidth
        ]{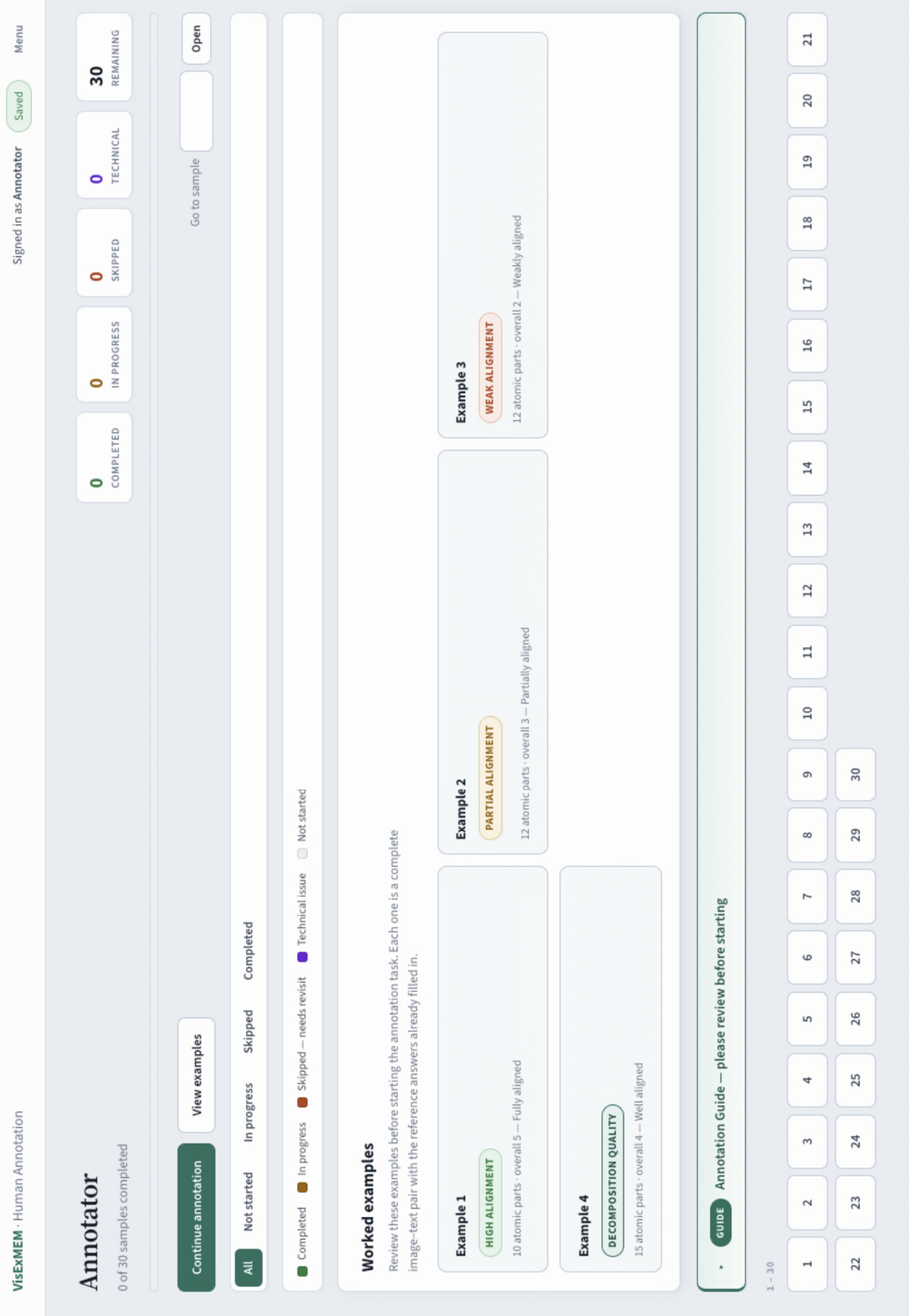}
    \end{minipage}

    \vspace{0.8em}

    \begin{minipage}{\linewidth}
        \centering
        \includegraphics[
            angle=270,
            width=.90\linewidth
        ]{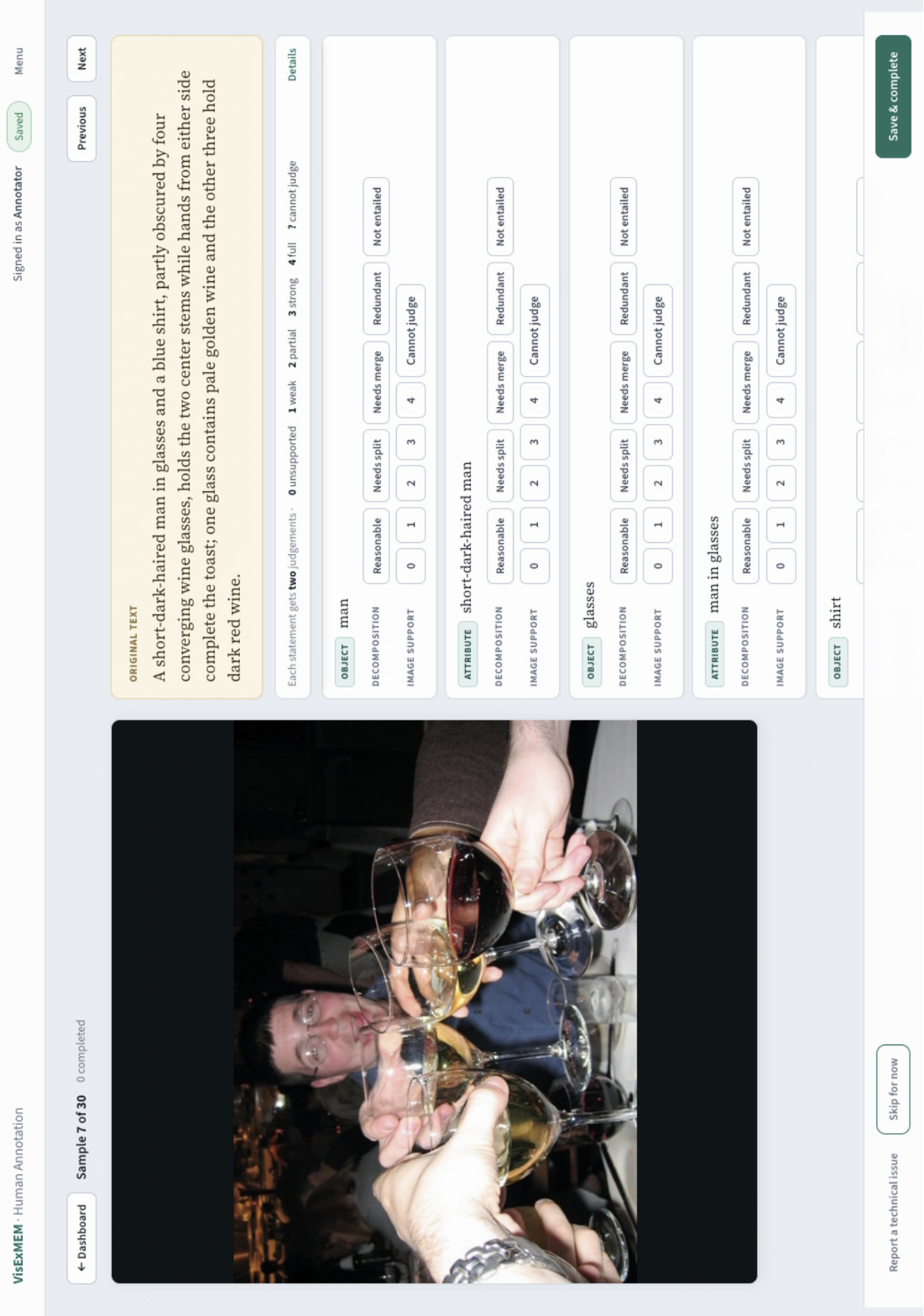}
    \end{minipage}

    \caption{\textbf{AGITA-Bench annotation interface.}
    Dashboard and sample-level interface used for human annotation.}
    \label{fig:agita_annotation_tool}
\end{figure}

\begin{figure}[p]
    \centering

    \begin{tikzpicture}[
        inner sep=0pt,
        outer sep=0pt
    ]

        \node[anchor=north] (guide1) at (0,0) {%
            \includegraphics[
                width=.95\linewidth
            ]{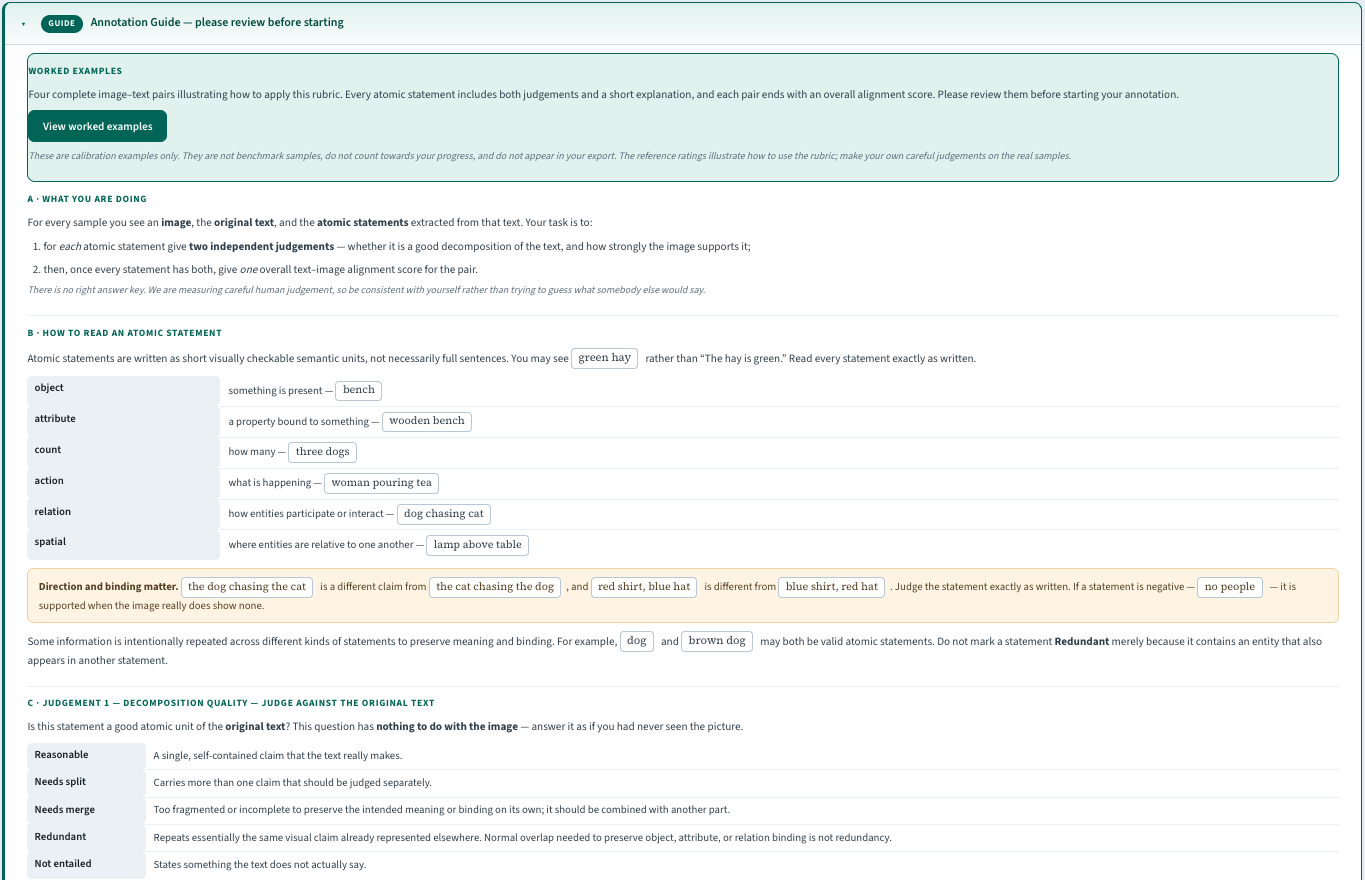}%
        };

        \node[
            anchor=north,
            yshift=0.3pt
        ] (guide2) at (guide1.south) {%
            \includegraphics[
                width=.95\linewidth
            ]{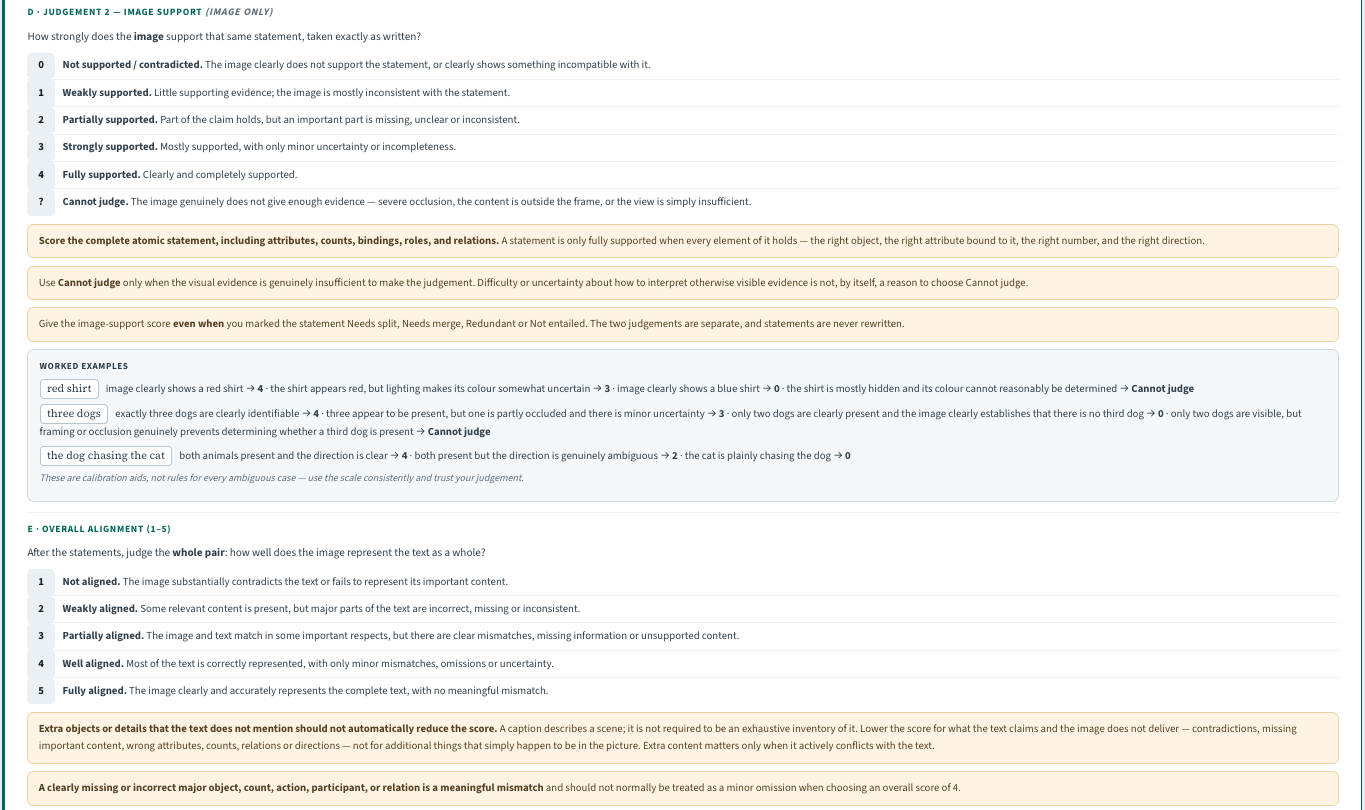}%
        };

        \node[
            anchor=north,
            yshift=0.3pt
        ] (guide3) at (guide2.south) {%
            \includegraphics[
                width=.95\linewidth
            ]{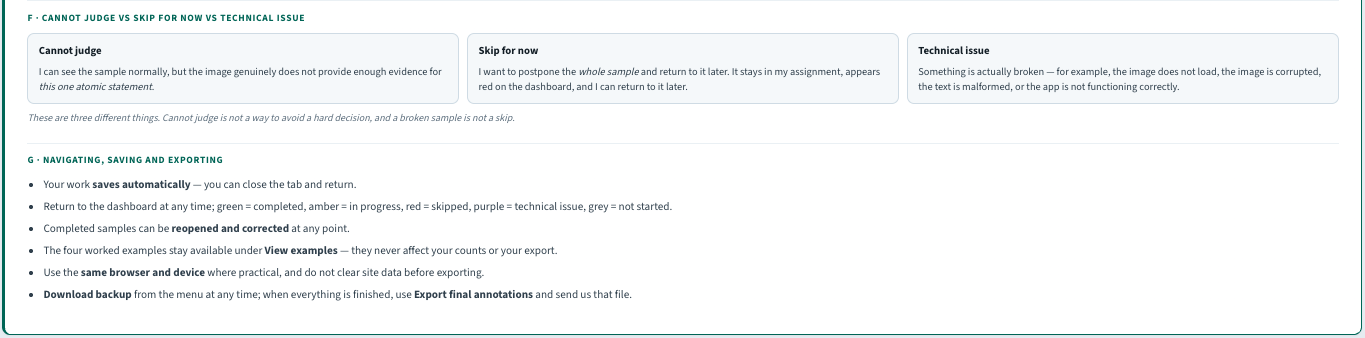}%
        };

    \end{tikzpicture}

    \caption{\textbf{AGITA-Bench annotation guidelines.}
    Complete annotator-facing guidelines.}
    \label{fig:agita_annotation_guidelines}
\end{figure}

Annotations are collected through a dedicated interface that presents the
same information and rating controls to every annotator.
Figure~\ref{fig:agita_annotation_tool} shows the dashboard and sample-level
interface. Before starting, annotators can review the annotation guide and
worked examples. For each assigned sample, they record decomposition quality
and atomic-support judgments for individual claims alongside the image and
input text. Global image-text alignment is rated once for the complete
sample. The full annotator instructions are shown in
Figure~\ref{fig:agita_annotation_guidelines}.

We also provide four worked examples covering high, partial, and weak
image-text alignment, including one example that shows the difference between
decomposition quality and visual support. Each example includes the input,
global alignment judgment, and claim-level decomposition and support
annotations.

\begin{table}[H]
\centering

\setlength{\abovecaptionskip}{2pt}
\setlength{\belowcaptionskip}{3pt}


\begin{minipage}[t]{0.485\textwidth}
\vspace{0pt}
\centering

\caption{\textbf{Inter-annotator agreement for global alignment.}
Pairwise quadratic-weighted Cohen's $\kappa$ on the 30 shared
image--text pairs; mean pairwise $\kappa=.602$. Only unique
annotator pairs are shown.}
\label{tab:agita_iaa_global}

\vspace{-2pt}

\scriptsize
\setlength{\tabcolsep}{2.8pt}
\renewcommand{\arraystretch}{0.96}

\begin{tabular}{@{}lcccccc@{}}
\toprule
& \textbf{Ann2} & \textbf{Ann3} & \textbf{Ann4}
& \textbf{Ann5} & \textbf{Ann6} & \textbf{Ann7} \\
\midrule
\textbf{Ann1} & .57 & .78 & .54 & .76 & .37 & .62 \\
\textbf{Ann2} &     & .62 & .70 & .64 & .54 & .48 \\
\textbf{Ann3} &     &     & .73 & .81 & .39 & .62 \\
\textbf{Ann4} &     &     &     & .63 & .47 & .61 \\
\textbf{Ann5} &     &     &     &     & .51 & .65 \\
\textbf{Ann6} &     &     &     &     &     & .61 \\
\bottomrule
\end{tabular}

\end{minipage}
\hfill
\begin{minipage}[t]{0.485\textwidth}
\vspace{0pt}
\centering

\caption{\textbf{Inter-annotator agreement for atomic support.}
Pairwise quadratic-weighted Cohen's $\kappa$ on shared claims;
\emph{cannot judge} responses are excluded pairwise
($N=418$--$453$). Mean pairwise $\kappa=.564$.}
\label{tab:agita_iaa_atomic}

\vspace{-2pt}

\scriptsize
\setlength{\tabcolsep}{2.8pt}
\renewcommand{\arraystretch}{0.96}

\begin{tabular}{@{}lcccccc@{}}
\toprule
& \textbf{Ann2} & \textbf{Ann3} & \textbf{Ann4}
& \textbf{Ann5} & \textbf{Ann6} & \textbf{Ann7} \\
\midrule
\textbf{Ann1} & .52 & .72 & .61 & .60 & .47 & .39 \\
\textbf{Ann2} &     & .59 & .60 & .58 & .59 & .55 \\
\textbf{Ann3} &     &     & .65 & .58 & .67 & .47 \\
\textbf{Ann4} &     &     &     & .55 & .55 & .49 \\
\textbf{Ann5} &     &     &     &     & .56 & .45 \\
\textbf{Ann6} &     &     &     &     &     & .66 \\
\bottomrule
\end{tabular}

\end{minipage}

\vspace{5pt}


\begin{minipage}[t]{0.58\textwidth}
\vspace{0pt}
\centering

\caption{\textbf{AGITA-Bench annotator demographics.}
Demographic summary of the seven human annotators ($N=7$).}
\label{tab:agita_annotator_demographics}

\vspace{-2pt}

\scriptsize
\setlength{\tabcolsep}{4pt}
\renewcommand{\arraystretch}{0.88}

\begin{tabular}{@{}llc@{}}
\toprule
\textbf{Attribute} & \textbf{Category} & \textbf{Count} \\
\midrule
Nationality
    & German  & 3 \\
    & Indian  & 1 \\
    & Chinese & 1 \\
    & UAE     & 2 \\
\midrule
Education
    & Postdoctoral researcher & 1 \\
    & PhD student             & 2 \\
    & Master's student        & 4 \\
\midrule
Age
    & 18--25 & 4 \\
    & 26--30 & 2 \\
    & $>30$  & 1 \\
\bottomrule
\end{tabular}

\end{minipage}

\end{table}

\input{figures/prompts/agita_decomp_prompt}

\input{figures/results_app_agita}


\clearpage

\input{figures/annotation_examples}

\newpage
\subsection{Explanation and Evidence Annotation}
\label{app:explanation_annotation}

\paragraph{Protocol.}
The study contains 300 image--text pairs. Each pair presents four
explanations, one each from EXPERT, VIEScore, GPT-5.6 Sol, and
\textsc{VisExMEM}, for 1,200 sample--system explanation instances.
System identities are hidden and the explanations are shown under
randomized labels \emph{A/B/C/D}. Annotation has two stages. In
Task~A, annotators rate all four explanations without seeing localized
evidence. They assess diagnostic correctness, visual grounding,
clarity, and diagnostic usefulness on a 1 to 5 scale, and flag
unsupported factual content. For systems that provide localized
evidence, Task~B is shown only after Task~A is complete. Annotators
then rate up to three claim-region pairs for localization accuracy
and evidence sufficiency, again on a 1 to 5 scale. Task~A responses
are locked before Task~B is shown so the highlighted regions cannot
affect the earlier explanation ratings. The complete questions,
scales, and instructions are reproduced in
Figure~\ref{fig:explanation_annotation_guidelines}.

\paragraph{Annotators and agreement.}
Three annotators took part in the study. Their reported countries
of nationality were Germany, India, and the UAE. Two were aged 18
to 25 and one was aged 26 to 30, and all three were Master's
students. All three annotated the same 30 image--text pairs.
Agreement is computed pairwise on items completed by both
annotators in each pair after resolving the randomized labels to
their underlying systems. For the ordinal ratings, we use
quadratic-weighted Cohen's $\kappa$, while \emph{Unsupported Content}
uses unweighted Cohen's $\kappa$. Across the four explanation-quality
criteria, mean pairwise weighted $\kappa$ is $.59$, with 89.5\% of
paired ratings within one scale point. \emph{Unsupported Content}
gives $\kappa=.42$. For localized evidence, mean weighted $\kappa$
is $.58$, with 86.9\% within-one-point agreement. Because the
evidence ratings are strongly concentrated near the top of the
scale, we report near-agreement together with $\kappa$.

\begin{table}[t]
\centering
\small
\setlength{\tabcolsep}{6pt}
\renewcommand{\arraystretch}{1.08}
\caption{\textbf{Inter-annotator agreement for explanation and
evidence evaluation.}
Values show mean pairwise agreement. Ordinal ratings use
quadratic-weighted Cohen's $\kappa$, while \emph{Unsupported Content}
uses unweighted $\kappa$.}
\label{tab:explanation_iaa}

\begin{tabular}{@{}lcc@{}}
\toprule
\textbf{Evaluation}
& \textbf{Mean $\kappa$}
& \textbf{Within $\pm1$ (\%)} \\
\midrule
Task A explanation quality
& .59 & 89.5 \\
Unsupported Content
& .42 & -- \\
Task B localized evidence
& .58 & 86.9 \\
\bottomrule
\end{tabular}
\end{table}

\input{figures/app_explanation_annotation_guideline}


%% file: figures/appendix_training_data.tex
%
%
\definecolor{TrainBlue}{HTML}{123F79}
\definecolor{TrainBG}{HTML}{F5F7FB}
\definecolor{TrainPurple}{HTML}{F8F5FB}
\definecolor{TrainGreen}{HTML}{F4F9F5}

\begin{SaveVerbatim}{StageOneLeft}
Source: coco_train2017
Sample: coco_3333

DECOMPOSITION VIEW

Caption:
"A fire truck near a yellow fire hydrant."

Target:
{"parts":[
 {"text":"fire truck",
  "type":"object"},
 {"text":"yellow fire hydrant",
  "type":"attribute"},
 {"text":"fire truck near a yellow fire hydrant",
  "type":"relation"}
]}
\end{SaveVerbatim}

\begin{SaveVerbatim}{StageOneRight}
EVIDENCE-SELECTION VIEW

Claim (attribute):
yellow fire hydrant

Candidate E1:
box=[93,279,224,482]

Target:
{"selected_evidence_ids":["E1"]}

Training signal:
supervised structured target
\end{SaveVerbatim}

\begin{SaveVerbatim}{StageTwoLeft}
Source: coco_train2017
Claim type: attribute

SUPPORTED SIDE

Caption:
"A fire truck near a yellow fire hydrant."

Claim:
yellow fire hydrant

Decision: YES
\end{SaveVerbatim}

\begin{SaveVerbatim}{StageTwoRight}
COUNTERFACTUAL SIDE

Caption:
"A fire truck near a green fire hydrant."

Claim:
green fire hydrant

Decision: NO

Change: yellow -> green
Image pixels: identical for this example

Training signal:
Yes/No decision-margin loss
Generated target text: none
\end{SaveVerbatim}

\begin{SaveVerbatim}{StageThreeLeft}
SEMANTIC EXAMPLE

Source: coco_train2017
Sample: cf_coco_9065_count

Caption:
"A dog and three people walking on sand."

Claim (count): three
Decision: NO

Rollouts per prompt: 8

Reward:
 +1.00  correct decision
 -1.00  incorrect decision
 -1.25  invalid JSON
\end{SaveVerbatim}

\begin{SaveVerbatim}{StageThreeRight}
CAUSAL-EVIDENCE EXAMPLE

Source: coco_train2017
Sample: coco_3333

Claim (attribute):
yellow fire hydrant

Decision: YES

Selected region:
[92.85,278.53,224.34,482.07]

Matched random region:
[141.57,62.35,270.76,272.01]

Image views:
full
mask_selected
mask_random
evidence_only
random_only
\end{SaveVerbatim}

\begin{figure}[p]
    \centering

    \begin{tikzpicture}

        \node[
            anchor=north,
            draw=TrainBlue,
            line width=0.9pt,
            rounded corners=3pt,
            fill=TrainBG,
            inner xsep=8pt,
            inner ysep=7pt
        ] (stage1) {%
            \begin{minipage}{0.95\textwidth}
                \vspace*{3mm}

                \begin{minipage}[t]{0.48\linewidth}
                    \vspace{0pt}
                    \UseVerbatim[
                        formatcom=\fontsize{6.2}{6.65}\selectfont,
                        breaklines=true,
                        breakanywhere=true,
                        breaksymbolleft={}
                    ]{StageOneLeft}
                \end{minipage}
                \hfill
                \begin{minipage}[t]{0.48\linewidth}
                    \vspace{0pt}
                    \UseVerbatim[
                        formatcom=\fontsize{6.2}{6.65}\selectfont,
                        breaklines=true,
                        breakanywhere=true,
                        breaksymbolleft={}
                    ]{StageOneRight}
                \end{minipage}

            \end{minipage}
        };

        \node[
            anchor=west,
            fill=TrainBlue,
            text=white,
            rounded corners=3pt,
            inner xsep=10pt,
            inner ysep=3.5pt,
            font=\bfseries\ttfamily\scriptsize
        ] at ([xshift=12pt]stage1.north west)
        {Stage 1: Structured SFT};

        \node[
            below=3mm of stage1,
            anchor=north,
            draw=TrainBlue,
            line width=0.9pt,
            rounded corners=3pt,
            fill=TrainPurple,
            inner xsep=8pt,
            inner ysep=7pt
        ] (stage2) {%
            \begin{minipage}{0.95\textwidth}
                \vspace*{3mm}

                \begin{minipage}[t]{0.48\linewidth}
                    \vspace{0pt}
                    \UseVerbatim[
                        formatcom=\fontsize{6.2}{6.65}\selectfont,
                        breaklines=true,
                        breakanywhere=true,
                        breaksymbolleft={}
                    ]{StageTwoLeft}
                \end{minipage}
                \hfill
                \begin{minipage}[t]{0.48\linewidth}
                    \vspace{0pt}
                    \UseVerbatim[
                        formatcom=\fontsize{6.2}{6.65}\selectfont,
                        breaklines=true,
                        breakanywhere=true,
                        breaksymbolleft={}
                    ]{StageTwoRight}
                \end{minipage}

            \end{minipage}
        };

        \node[
            anchor=west,
            fill=TrainBlue,
            text=white,
            rounded corners=3pt,
            inner xsep=10pt,
            inner ysep=3.5pt,
            font=\bfseries\ttfamily\scriptsize
        ] at ([xshift=12pt]stage2.north west)
        {Stage 2: Counterfactual Refinement};

        \node[
            below=3mm of stage2,
            anchor=north,
            draw=TrainBlue,
            line width=0.9pt,
            rounded corners=3pt,
            fill=TrainGreen,
            inner xsep=8pt,
            inner ysep=7pt
        ] (stage3) {%
            \begin{minipage}{0.95\textwidth}
                \vspace*{3mm}

                \begin{minipage}[t]{0.48\linewidth}
                    \vspace{0pt}
                    \UseVerbatim[
                        formatcom=\fontsize{6.2}{6.65}\selectfont,
                        breaklines=true,
                        breakanywhere=true,
                        breaksymbolleft={}
                    ]{StageThreeLeft}
                \end{minipage}
                \hfill
                \begin{minipage}[t]{0.48\linewidth}
                    \vspace{0pt}
                    \UseVerbatim[
                        formatcom=\fontsize{6.2}{6.65}\selectfont,
                        breaklines=true,
                        breakanywhere=true,
                        breaksymbolleft={}
                    ]{StageThreeRight}
                \end{minipage}

            \end{minipage}
        };

        \node[
            anchor=west,
            fill=TrainBlue,
            text=white,
            rounded corners=3pt,
            inner xsep=10pt,
            inner ysep=3.5pt,
            font=\bfseries\ttfamily\scriptsize
        ] at ([xshift=12pt]stage3.north west)
        {Stage 3: Evidence-Aware GDPO};

    \end{tikzpicture}

    \caption{\textbf{Representative processed training examples.}
    Real trainer-facing examples from the three VisExMEM training stages.
    Unused metadata are omitted and coordinates are rounded for readability.}
    \label{fig:processed_training_examples}

\end{figure}

%% file: figures/app_ood_results.tex

\begin{table*}[t]
\centering
\caption{
\textbf{Out-of-distribution generalization.}
Results on VisPlotBench, CulturalFrames, and the 600 human-annotated
ProtoBias examples. VisPlotBench reports pairwise accuracy (PAcc) and
Group score; CulturalFrames reports Spearman's $\rho$ and within-prompt
pairwise accuracy (PAcc$_{\mathrm{wp}}$); ProtoBias reports
category-level accuracy (Acc.) and Cohen's $\kappa$.
Accuracy-based metrics are reported on a 0--100 scale; correlations and
$\kappa$ are reported on their native scale. Higher is better.
Bold, underlined, and italicized values denote the best, second-best,
and third-best results in each column, respectively.
}
\label{tab:ood_results}

\scriptsize
\setlength{\tabcolsep}{2.8pt}
\renewcommand{\arraystretch}{1.08}

\resizebox{0.85\textwidth}{!}{%
\begin{tabular}{lcc|cc|cc|cc|cc}
\toprule

& \multicolumn{2}{c|}{\textbf{VisPlotBench}}
& \multicolumn{2}{c|}{\textbf{CulturalFrames}}
& \multicolumn{6}{c}{\textbf{ProtoBias}} \\

\cmidrule(lr){2-3}
\cmidrule(lr){4-5}
\cmidrule(lr){6-11}

\textbf{Method}
& \textbf{PAcc}
& \textbf{Group}
& $\boldsymbol{\rho}$
& \textbf{PAcc$_{\mathrm{wp}}$}
& \multicolumn{2}{c|}{\textbf{Animal}}
& \multicolumn{2}{c|}{\textbf{Demography}}
& \multicolumn{2}{c}{\textbf{Object}} \\

& & & & &
\textbf{Acc.} & $\boldsymbol{\kappa}$
& \textbf{Acc.} & $\boldsymbol{\kappa}$
& \textbf{Acc.} & $\boldsymbol{\kappa}$ \\

\midrule

CLIPScore
& 82.5 & 52.6
& .145 & 60.5
& 25.0 & .001
& 28.0 & -.005
& 32.0 & .058 \\


ALIGNScore
& 79.1 & 45.3
& .149 & 57.8
& 51.0 & .033
& 34.0 & -.064
& 32.0 & .011 \\

PickScore
& 77.1 & 39.5
& .206 & 62.2
& 32.0 & .001
& 34.0 & -.014
& 31.0 & .126 \\


HPSv2
& 79.7 & 37.4
& .219 & 61.5
& 25.0 & .024
& 38.0 & -.006
& 38.0 & .125 \\

VQAScore
& 85.1 & 61.4
& .250 & 64.4
& 51.0 & .028
& 47.0 & -.007
& 55.0 & .065 \\

ChartVE
& 69.2 & 31.3
& -- & --
& -- & --
& -- & --
& -- & -- \\

GPT-5 (Judge)
& -- & --
& -- & --
& 56.0 & .108
& 51.0 & \textit{.176}
& 54.0 & \textit{.265} \\

ProtoScore
& -- & --
& -- & --
& \textbf{95.0} & \textit{.182}
& \textbf{91.0} & \textbf{.232}
& \textbf{98.0} & .178 \\

AHEaD-ALIGN
& -- & --
& .114 & 55.6
& 35.5 & .064
& 49.5 & .083
& 55.5 & -.030 \\

AHEaD-HAL ($1-\mathrm{HAL}$)
& -- & --
& .131 & 55.0
& -- & --
& -- & --
& -- & -- \\

\midrule

Qwen3.5-4B (Hol.)
& 84.4 & 70.2
& .242 & 56.7
& 39.8 & .131
& 45.8 & .075
& 44.5 & .161 \\

Qwen3.5-4B (Dec.)
& \underline{85.8} & 73.8
& \underline{.296} & 49.4
& 43.5 & .121
& 39.0 & .102
& 60.0 & \underline{.269} \\

Qwen3.5-9B (Hol.)
& 81.5 & 72.4
& \textit{.252} & 15.2
& 32.0 & .049
& 26.0 & .142
& 40.5 & .102 \\

Qwen3.5-9B (Dec.)
& 83.4 & \underline{74.5}
& .210 & 51.4
& 42.9 & .056
& 54.0 & .167
& 55.6 & .195 \\

\midrule

\rowcolor{green!6}
\textbf{VisExMEM-4B}
& \textbf{91.3} & \textit{74.4}
& .198 & 60.5
& \underline{76.0} & \underline{.189}
& \textit{67.0} & .106
& \textit{73.5} & .256 \\

\rowcolor{green!6}
\textbf{VisExMEM-9B}
& \textbf{91.3} & \textbf{78.6}
& \textbf{.311} & \textbf{65.8}
& \textit{73.5} & \textbf{.254}
& \underline{71.5} & \underline{.205}
& \underline{81.0} & \textbf{.286} \\

\bottomrule
\end{tabular}%
}

\end{table*}

%% file: figures/prompts/hol_prompt.tex

\begin{SaveVerbatim}{HolisticJudgePrompt}
----- HOLISTIC system -----
You are a strict visual alignment evaluator.
You judge how well a caption matches an image, using only visible image evidence.

Do not assume something is true because it sounds plausible.
Do not use world knowledge when the image does not show it.
Do not reward fluency or aesthetic quality.

Return only the required JSON schema.

----- HOLISTIC user (image + full caption; no decomposition) -----
Judge how well the caption below matches the supplied image.

Consider every visually verifiable element the caption expresses: objects, attributes, actions, relations, counts, and negation.
A meaningful contradiction must lower the score even when most of the caption is correct.

alignment_score rubric (from 0.05 to 0.95):
  0.90-0.95  clearly and directly supported by visible evidence
  0.75-0.85  strongly supported, with minor visual ambiguity
  0.60-0.70  more supported than not, but incomplete/weak evidence
  0.45-0.55  genuinely uncertain / ambiguous / not enough visible evidence
  0.30-0.40  more unsupported than supported
  0.15-0.25  strongly unsupported / likely contradicted
  0.05-0.10  clearly contradicted, or required visual evidence is clearly absent

Return valid JSON only, in exactly this shape:
{"alignment_score":0.95}

Caption: {caption}
\end{SaveVerbatim}

\begin{figure}[p]
    \centering

    \begin{tikzpicture}

        \node[
            anchor=north,
            draw=PromptBlue,
            line width=1.1pt,
            rounded corners=3pt,
            fill=PromptBG,
            inner xsep=12pt,
            inner ysep=13pt
        ] (promptbox) {%
            \begin{minipage}{0.94\linewidth}

                \vspace*{5mm}

                \UseVerbatim[
                    fontsize=\small,
                    baselinestretch=0.98,
                    breaklines=true,
                    breakanywhere=true,
                    breaksymbolleft={}
                ]{HolisticJudgePrompt}

                \vspace*{1mm}

            \end{minipage}%
        };

        \node[
            anchor=west,
            fill=PromptBlue,
            text=white,
            rounded corners=3pt,
            inner xsep=12pt,
            inner ysep=5pt,
            font=\bfseries\ttfamily\small
        ] at ([xshift=16pt]promptbox.north west)
        {VLM-as-a-Judge (Holistic)};

    \end{tikzpicture}

    \caption{
        Prompt used for the holistic VLM-as-a-Judge setting.
        The model directly evaluates the complete image--caption pair
        and returns a single alignment score without atomic decomposition.
    }
    \label{fig:vlm_judge_holistic_prompt}

\end{figure}

%% file: figures/prompts/dec_prompt.tex

%


\begin{SaveVerbatim}{DecompPromptColOne}
You are an expert visual evaluator.

Your task is to evaluate how well the provided image supports the provided
caption.

You must perform the COMPLETE evaluation in this single response.

CAPTION:
{caption}

Follow these steps internally in order:

1. Decompose the caption into visually verifiable atomic claims.
2. For every atomic claim, locate the visual evidence needed to judge it.
3. Assign an evidence-grounded support score to every atomic claim.
4. Give one short explanation for each score based on the visible evidence.
5. Only after evaluating all atomic claims, produce one overall
   image-caption support score based on those part-level judgments.

Do not skip any visually meaningful component of the caption.

--------------------------------
ATOMIC DECOMPOSITION
--------------------------------

Split the caption into the smallest useful independently verifiable visual
claims.

Possible claim types include:

- object
- attribute
- count
- action
- relation
- spatial
- negation
- other

Examples of information that should normally become separate claims when
present:

- an object exists;
- an object has an attribute;
- a specific number of objects exists;
- an entity performs an action;
- one entity acts on another;
- two entities have a relation;
- an entity is left/right/above/below/in-front-of/behind another;
- a stated object is absent or a condition is negated.

Preserve binding.

For example, do not reduce:

"the man in a red shirt is holding a blue umbrella"

to unrelated statements such as:

"red exists"
"blue exists"
"man exists"
"umbrella exists"

The atomic claims must preserve which property/action/relation belongs to
which entity.

Do not invent claims not contained in the caption.

Do not evaluate non-visual linguistic details that cannot reasonably be
verified from the image.

--------------------------------
BOUNDING BOXES / VISUAL EVIDENCE
--------------------------------

For each atomic claim identify only the visual entities or regions needed
to verify that claim.

Represent every bounding box using normalized coordinates:

[x_min, y_min, x_max, y_max]

with each coordinate an integer from 0 to 1000.

Coordinate system:

- (0,0) = top-left of the image
- (1000,1000) = bottom-right of the image

Each evidence item must contain:

- label
- role
- box

Example:

{
  "label": "dog",
  "role": "subject",
  "box": [120, 310, 430, 820]
}

Use multiple boxes when necessary.
\end{SaveVerbatim}


\begin{SaveVerbatim}{DecompPromptColTwo}
Examples:

COUNT:
box all clearly visible relevant instances needed to judge the count.

RELATION / SPATIAL:
box all relevant participating entities.

ACTION:
box the actor and other important participant/object where applicable.

ATTRIBUTE:
box the entity to which the attribute is bound.

INCORRECT CLAIM:
show the visible counter-evidence when possible.

For example, if the caption claims three dogs but only two are visible,
return the boxes of those two visible dogs.

NEGATION / ABSENCE:
if the claim is supported specifically because something is absent and
there is no meaningful region to box, `evidence_boxes` may be an empty
list.

Do not invent bounding boxes merely to satisfy the schema.

--------------------------------
PART-LEVEL SUPPORT SCORE
--------------------------------

For every atomic claim output a support score in:

[0.0, 1.0]

Interpretation:

1.00
= clearly and fully supported by the image

0.75
= mostly supported, with minor ambiguity or uncertainty

0.50
= genuinely ambiguous, partially supported, or visually uncertain

0.25
= mostly unsupported

0.00
= clearly contradicted or absent

Use the full continuous range when appropriate.

The score must measure SUPPORT FOR THE SPECIFIC ATOMIC CLAIM, not generic
image-caption similarity.

Pay particular attention to:

- entity identity;
- attribute binding;
- subject/object roles;
- action participants;
- spatial ordering;
- counts;
- negation;
- same-class entity binding.

Do not give a high score merely because all relevant words or objects
appear somewhere in the image.

--------------------------------
SHORT EXPLANATION
--------------------------------

For each atomic claim provide ONE concise evidence-grounded sentence.

The explanation should state the visually relevant fact that justifies the
score.

Good examples:

"Two dogs are visible, while the claim requires three."

"The red shirt is worn by the man, matching the claim."

"The dog is to the right of the cat, not to its left."

"The woman is holding the umbrella rather than the man."

Avoid:

- generic descriptions;
- unnecessary prose;
- hidden chain-of-thought;
- repeating the numerical score;
- mentioning coordinate values in the explanation.
\end{SaveVerbatim}


\begin{SaveVerbatim}{DecompPromptColThree}
The explanation must agree with both the score and the returned evidence
boxes.

--------------------------------
OVERALL SCORE
--------------------------------

After ALL atomic parts have been evaluated, return:

overall_score

in:

[0.0, 1.0]

The overall score must be derived from your completed part-level
judgments.

IMPORTANT:

Do NOT independently ignore the part scores and make an unrelated holistic
judgment.

The overall score should reflect how well the caption as a whole is
supported by the image given the atomic-part scores.

A strongly contradicted essential claim --- especially an incorrect count,
binding, action, relation, spatial relation, or negation --- should
meaningfully lower the overall score.

At the same time, do not invent an external formula or expose
chain-of-thought.

The final overall score is YOUR end-to-end assessment based on the
part-level results you produced above.

--------------------------------
OUTPUT FORMAT
--------------------------------

Return ONLY valid JSON.

No Markdown.
No code fences.
No text before or after the JSON.

Use exactly this top-level structure:

{
  "atomic_parts": [
    {
      "id": "p1",
      "claim": "atomic visual claim",
      "type": "object",
      "evidence_boxes": [
        {
          "label": "entity name",
          "role": "target",
          "box": [0, 0, 1000, 1000]
        }
      ],
      "score": 0.95,
      "explanation": "One short evidence-grounded sentence."
    }
  ],
  "overall_score": 0.90
}

Requirements:

- `atomic_parts` must contain at least one item.
- IDs must be p1, p2, p3, ... in order.
- Every claim must have exactly one score.
- Every claim must have exactly one explanation.
- Every score must be within [0,1].
- Every box coordinate must be within [0,1000].
- `evidence_boxes` may be [] only when no meaningful localized visual
  evidence exists.
- `overall_score` must be within [0,1].
- `overall_score` must be produced after considering all atomic-part
  judgments.

Return only the JSON object.
\end{SaveVerbatim}


\begin{figure}[p]
    \centering
    \captionsetup{skip=2pt}

    \begin{tikzpicture}
        \node[
            anchor=north,
            draw=PromptBlue,
            line width=1.0pt,
            rounded corners=3pt,
            fill=PromptBG,
            inner xsep=8pt,
            inner ysep=6pt
        ] (promptbox) {%
            \begin{minipage}{0.95\textwidth}

                \vspace*{3.2mm}

                \begin{minipage}[t]{0.323\linewidth}
                    \vspace{0pt}
                    \UseVerbatim[
                        formatcom=\fontsize{4.95}{5.20}\selectfont,
                        baselinestretch=0.89,
                        breaklines=true,
                        breakanywhere=true,
                        breaksymbolleft={}
                    ]{DecompPromptColOne}
                \end{minipage}
                \hfill
                \begin{minipage}[t]{0.323\linewidth}
                    \vspace{0pt}
                    \UseVerbatim[
                        formatcom=\fontsize{4.95}{5.20}\selectfont,
                        baselinestretch=0.89,
                        breaklines=true,
                        breakanywhere=true,
                        breaksymbolleft={}
                    ]{DecompPromptColTwo}
                \end{minipage}
                \hfill
                \begin{minipage}[t]{0.323\linewidth}
                    \vspace{0pt}
                    \UseVerbatim[
                        formatcom=\fontsize{4.95}{5.20}\selectfont,
                        baselinestretch=0.89,
                        breaklines=true,
                        breakanywhere=true,
                        breaksymbolleft={}
                    ]{DecompPromptColThree}
                \end{minipage}

                \vspace*{1.2mm}

            \end{minipage}
        };

        \node[
            anchor=west,
            fill=PromptBlue,
            text=white,
            rounded corners=3pt,
            inner xsep=12pt,
            inner ysep=4.5pt,
            font=\bfseries\ttfamily\scriptsize
        ] at ([xshift=15pt]promptbox.north west)
        {VLM-as-a-Judge (Decomposed, End-to-End)};
    \end{tikzpicture}

    \caption{
        Prompt used for the decomposed end-to-end
        VLM-as-a-Judge setting.
        The model performs atomic decomposition, evidence localization,
        claim-level scoring, short explanation generation, and final
        overall scoring in a single response.
    }
    \label{fig:vlm_judge_decomposed_prompt}
\end{figure}

%% file: figures/prompts/agita_decomp_prompt.tex

\begin{SaveVerbatim}{AGITADecompPromptLeft}
You split a caption into VISUAL SEMANTIC UNITS for a compositional benchmark.

You receive ONLY the caption text. You have not seen any image. Never add anything the caption does not state, and never judge whether it is true.

THE UNITS ARE PHRASES, NOT SENTENCES.

Stay close to the caption's own wording. Do not rewrite meaning into full propositions. This is the single most important rule.

caption: "A person is moving green hay towards an elephant that is inside the back of a white truck."

  CORRECT                                              type
    person                                             object
    hay                                                object
    green hay                                          attribute
    elephant                                           object
    truck                                              object
    white truck                                        attribute
    moving                                             action
    green hay towards an elephant                      relation
    elephant inside the back of a white truck          spatial

  WRONG -- do not produce this
    A person exists.
    Hay exists.
    The hay is green.
    An elephant exists.
    A truck exists.
    The truck is white.

caption: "Dozens of mostly red apples hang from the branches of a leafy green fruit tree."

  CORRECT
    apples                                             object
    dozens of apples                                   count
    mostly red apples                                  attribute
    branches                                           object
    fruit tree                                         object
    leafy fruit tree                                   attribute
    green fruit tree                                   attribute
    hang                                               action
    apples hang from the branches of a fruit tree      relation

  WRONG
    There are dozens of apples.
    Most of the apples are red.
    The fruit tree is leafy.
    The fruit tree is green.
\end{SaveVerbatim}

\begin{SaveVerbatim}{AGITADecompPromptRight}
BINDING IS THE POINT. An attribute, count or role must stay attached to the entity the caption gives it.

  "The woman wears blue and the man wears red."
     woman / man / woman wears blue / man wears red
     NOT: woman / man / blue / red

  "The dog chases the cat." -> dog chases cat
     (dog is the agent; never "cat chases dog")

  "one black dog and one white dog" ->
     dogs / black dog / white dog

  "A red cup is to the left of a blue bowl." ->
     red cup / blue bowl / red cup left of blue bowl

  "A man without a hat" ->
     man / man without a hat
     (keep the negation)

TYPES -- use exactly these:

  object
    a visible entity, as a compact phrase:
    person, dog, fruit tree, branches

  attribute
    a property bound to its entity:
    green hay, white truck, mostly red apples

  count
    a quantity bound to its entity:
    three dogs, dozens of apples

  action
    what is happening, compactly:
    moving, hang, chases, holding

  relation
    who/what relates to whom/what:
    apples hang from the branches

  spatial
    an arrangement, direction preserved:
    cup left of bowl, cat under table

  negation
    an absence or denial:
    no dog, man without a hat

  other
    visually checkable, none of the above

GRANULARITY -- the smallest USEFUL unit, not every lexical combination.

  "leafy green fruit tree" ->
     fruit tree, leafy fruit tree, green fruit tree

  NOT also:
     tree, leafy tree, green tree,
     leafy green tree, leafy green fruit tree

  But "dog" AND "black dog" are BOTH wanted:
     presence and attribute are different checks.

  Likewise "dogs" AND "three dogs".

OMIT entirely: style, medium, quality and render words
(8k, cinematic, unreal engine, masterpiece, trending on artstation,
photorealistic, octane render, artist names), camera and lens jargon,
aspect ratios, mood words with no visual test (epic, dreamy), and
unobservable facts about a subject (their nationality, their profession
when not visible).

Return STRICT JSON, no prose:

{"parts": [{"text": "...", "type": "object"},
           {"text": "...", "type": "attribute"}]}

Do not number the parts and do not invent ids.
\end{SaveVerbatim}

\begin{figure}[p]
    \centering
    \captionsetup{skip=3pt}

    \begin{tikzpicture}

        \node[
            anchor=north,
            draw=PromptBlue,
            line width=1.0pt,
            rounded corners=3pt,
            fill=PromptBG,
            inner xsep=9pt,
            inner ysep=8pt
        ] (promptbox) {%
            \begin{minipage}{0.96\textwidth}

                \vspace*{3mm}

                \begin{minipage}[t]{0.485\linewidth}
                    \vspace{0pt}

                    \UseVerbatim[
                        fontsize=\tiny,
                        baselinestretch=0.92,
                        breaklines=true,
                        breakanywhere=true,
                        breaksymbolleft={}
                    ]{AGITADecompPromptLeft}

                \end{minipage}
                \hfill
                \begin{minipage}[t]{0.485\linewidth}
                    \vspace{0pt}

                    \UseVerbatim[
                        fontsize=\tiny,
                        baselinestretch=0.92,
                        breaklines=true,
                        breakanywhere=true,
                        breaksymbolleft={}
                    ]{AGITADecompPromptRight}

                \end{minipage}

            \end{minipage}
        };

        \node[
            anchor=west,
            fill=PromptBlue,
            text=white,
            rounded corners=3pt,
            inner xsep=12pt,
            inner ysep=4pt,
            font=\bfseries\ttfamily\small
        ] at ([xshift=16pt]promptbox.north west)
        {AGITA-Bench Atomic Decomposition};

    \end{tikzpicture}

    \caption{
        Prompt used for AGITA-Bench atomic decomposition.
        GPT-5.6 Sol receives only the caption and decomposes it into
        typed visual semantic units without access to the image or
        making any image-based correctness judgment.
    }
    \label{fig:agita_decomp_prompt}

\end{figure}

%% file: figures/annotation_examples.tex

\definecolor{agitaBlue}{HTML}{254E7B}
\definecolor{agitaBlueLight}{HTML}{F2F6FB}
\definecolor{agitaBlueBorder}{HTML}{A9C5E6}

\definecolor{agitaGreen}{HTML}{155D34}
\definecolor{agitaGreenLight}{HTML}{E9F5E9}

\definecolor{agitaGold}{HTML}{A86F00}
\definecolor{agitaGoldLight}{HTML}{FFF8E8}
\definecolor{agitaGoldBorder}{HTML}{EBC15E}

\definecolor{agitaTableBlue}{HTML}{E8F1FB}
\definecolor{agitaTableGreen}{HTML}{E5F3E5}
\definecolor{agitaSupportBlue}{HTML}{EEF5FD}

\newtcolorbox{agitaexamplebox}{
    enhanced,
    colback=white,
    colframe=agitaBlueBorder,
    boxrule=0.7pt,
    arc=2mm,
    left=7pt,
    right=7pt,
    top=7pt,
    bottom=7pt,
    boxsep=0pt,
    before skip=4pt,
    after skip=4pt
}

\newtcolorbox{agitainputbox}{
    colback=agitaBlueLight,
    colframe=agitaBlueBorder,
    boxrule=0.45pt,
    arc=1.5mm,
    left=7pt,
    right=7pt,
    top=6pt,
    bottom=6pt,
    boxsep=0pt
}

\newtcolorbox{agitascorebox}{
    colback=agitaGreenLight,
    colframe=agitaGreenLight,
    boxrule=0pt,
    arc=1.5mm,
    left=7pt,
    right=7pt,
    top=7pt,
    bottom=7pt,
    boxsep=0pt
}

\newtcolorbox{agitarationalebox}{
    colback=agitaGoldLight,
    colframe=agitaGoldBorder,
    boxrule=0.5pt,
    arc=1.5mm,
    left=8pt,
    right=8pt,
    top=6pt,
    bottom=6pt,
    boxsep=0pt
}

\newcommand{\agitabiasbadge}[1]{%
    \tcbox[
        on line,
        colback=agitaGreenLight,
        colframe=agitaGreenLight,
        boxrule=0pt,
        arc=3mm,
        left=7pt,
        right=7pt,
        top=3pt,
        bottom=3pt,
        boxsep=0pt
    ]{\textbf{\color{agitaGreen}#1}}%
}


\clearpage

\phantomsection
\label{example:worked-example-1}

\vspace*{\fill}

\begin{agitaexamplebox}


\noindent
\begin{minipage}[c]{0.70\linewidth}
    {\large\bfseries
    \color{agitaBlue}
    Example 1:
    \color{black}
    High alignment}
\end{minipage}%
\hfill
\begin{minipage}[c]{0.27\linewidth}
    \raggedleft
    \agitabiasbadge{High alignment}
\end{minipage}

\vspace{0.65em}


\noindent
\begin{minipage}[t]{0.40\linewidth}
    \vspace{0pt}
    \includegraphics[
        width=\linewidth,
        height=0.245\textheight,
        keepaspectratio
    ]{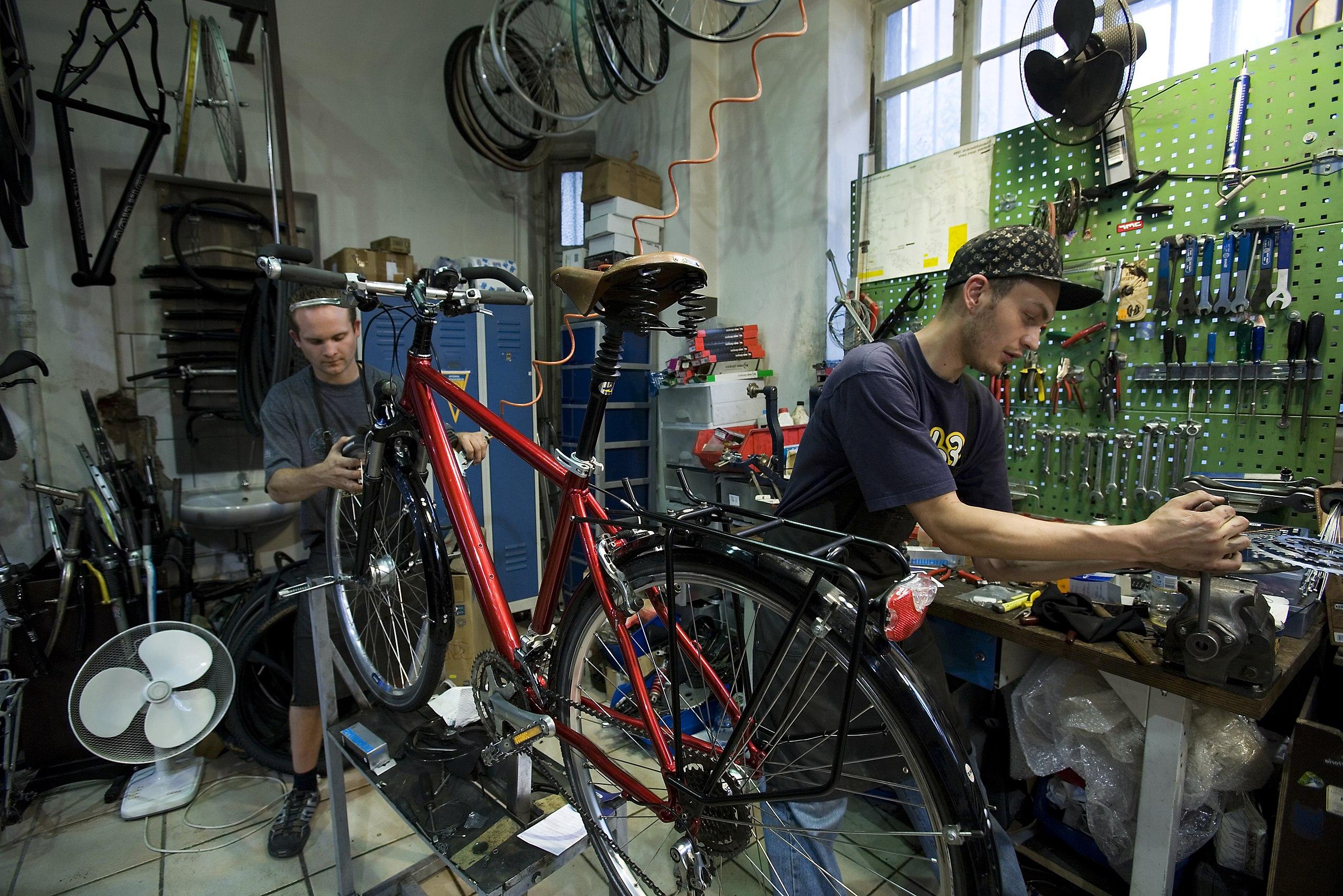}
\end{minipage}%
\hfill
\begin{minipage}[t]{0.57\linewidth}
    \vspace{0pt}

    \begin{agitainputbox}
        {\bfseries\color{agitaBlue}Input text}

        \vspace{0.35em}

        A man works on a red bicycle in a workshop with
        tools hanging on a green pegboard.
    \end{agitainputbox}

    \vspace{0.55em}

    \begin{agitascorebox}
        \noindent
        \begin{minipage}[c]{0.68\linewidth}
            {\bfseries\color{agitaGreen}
            Global alignment score}
        \end{minipage}%
        \begin{minipage}[c]{0.30\linewidth}
            \raggedleft
            {\Large\bfseries\color{agitaGreen}5 / 5}
        \end{minipage}
    \end{agitascorebox}

\end{minipage}

\vspace{0.75em}


\footnotesize

\setlength{\tabcolsep}{5pt}
\renewcommand{\arraystretch}{1.16}

\begin{tabularx}{\linewidth}{
    >{\centering\arraybackslash}p{0.035\linewidth}
    >{\raggedright\arraybackslash}p{0.12\linewidth}
    >{\raggedright\arraybackslash}X
    >{\centering\arraybackslash}p{0.19\linewidth}
    >{\centering\arraybackslash}p{0.14\linewidth}
}

\arrayrulecolor{agitaBlueBorder}

\rowcolor{agitaTableBlue}
\textbf{\#} &
\textbf{Type} &
\textbf{Atomic claim} &
\textbf{Decomp.\ quality} &
\textbf{Atomic support} \\
\hline

1 &
object &
man &
\cellcolor{agitaTableGreen}Reasonable &
\cellcolor{agitaSupportBlue}\textbf{4} \\

2 &
object &
bicycle &
\cellcolor{agitaTableGreen}Reasonable &
\cellcolor{agitaSupportBlue}\textbf{4} \\

3 &
attribute &
red bicycle &
\cellcolor{agitaTableGreen}Reasonable &
\cellcolor{agitaSupportBlue}\textbf{4} \\

4 &
object &
workshop &
\cellcolor{agitaTableGreen}Reasonable &
\cellcolor{agitaSupportBlue}\textbf{4} \\

5 &
object &
tools &
\cellcolor{agitaTableGreen}Reasonable &
\cellcolor{agitaSupportBlue}\textbf{4} \\

6 &
object &
pegboard &
\cellcolor{agitaTableGreen}Reasonable &
\cellcolor{agitaSupportBlue}\textbf{4} \\

7 &
attribute &
green pegboard &
\cellcolor{agitaTableGreen}Reasonable &
\cellcolor{agitaSupportBlue}\textbf{4} \\

8 &
action &
man works on a red bicycle &
\cellcolor{agitaTableGreen}Reasonable &
\cellcolor{agitaSupportBlue}\textbf{4} \\

9 &
spatial &
man in a workshop &
\cellcolor{agitaTableGreen}Reasonable &
\cellcolor{agitaSupportBlue}\textbf{4} \\

10 &
relation &
tools hanging on a green pegboard &
\cellcolor{agitaTableGreen}Reasonable &
\cellcolor{agitaSupportBlue}\textbf{4} \\

\end{tabularx}

\arrayrulecolor{black}
\normalsize

\vspace{0.75em}


\begin{agitarationalebox}

\noindent
\begin{minipage}[t]{0.06\linewidth}
    \centering
    {\Large\color{agitaGold}$\bullet$}
\end{minipage}%
\begin{minipage}[t]{0.91\linewidth}

    \textbf{Overall rationale}

    \vspace{0.25em}

    The image clearly supports the man, red bicycle,
    workshop, green pegboard, tools, and the described
    actions and relations.

\end{minipage}

\end{agitarationalebox}

\end{agitaexamplebox}

\vspace*{\fill}

\clearpage


\clearpage

\phantomsection
\label{example:worked-example-2}

\vspace*{\fill}

\begin{agitaexamplebox}


\noindent
\begin{minipage}[c]{0.70\linewidth}
    {\large\bfseries
    \color{agitaBlue}
    Example 2:
    \color{black}
    Partial alignment}
\end{minipage}%
\hfill
\begin{minipage}[c]{0.27\linewidth}
    \raggedleft
    \agitabiasbadge{Partial alignment}
\end{minipage}

\vspace{0.65em}


\noindent
\begin{minipage}[t]{0.40\linewidth}
    \vspace{0pt}
    \includegraphics[
        width=\linewidth,
        height=0.22\textheight,
        keepaspectratio
    ]{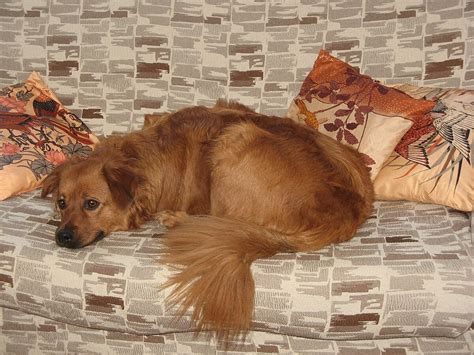}
\end{minipage}%
\hfill
\begin{minipage}[t]{0.57\linewidth}
    \vspace{0pt}

    \begin{agitainputbox}
        {\bfseries\color{agitaBlue}Input text}

        \vspace{0.35em}

        A black dog lies on a patterned sofa beside several pillows,
        with a red ball near its front paws.
    \end{agitainputbox}

    \vspace{0.55em}

    \begin{agitascorebox}
        \noindent
        \begin{minipage}[c]{0.68\linewidth}
            {\bfseries\color{agitaGreen}
            Global alignment score}
        \end{minipage}%
        \begin{minipage}[c]{0.30\linewidth}
            \raggedleft
            {\Large\bfseries\color{agitaGreen}3 / 5}
        \end{minipage}
    \end{agitascorebox}

\end{minipage}

\vspace{0.65em}


\footnotesize
\setlength{\tabcolsep}{4.5pt}
\renewcommand{\arraystretch}{1.08}

\begin{tabularx}{\linewidth}{
    >{\centering\arraybackslash}p{0.035\linewidth}
    >{\raggedright\arraybackslash}p{0.115\linewidth}
    >{\raggedright\arraybackslash}X
    >{\centering\arraybackslash}p{0.185\linewidth}
    >{\centering\arraybackslash}p{0.145\linewidth}
}

\arrayrulecolor{agitaBlueBorder}

\rowcolor{agitaTableBlue}
\textbf{\#} &
\textbf{Type} &
\textbf{Atomic claim} &
\textbf{Decomp.\ quality} &
\textbf{Atomic support} \\
\hline

1 & object & dog
& \cellcolor{agitaTableGreen}Reasonable
& \cellcolor{agitaSupportBlue}\textbf{4} \\

2 & attribute & black dog
& \cellcolor{agitaTableGreen}Reasonable
& \cellcolor{agitaSupportBlue}\textbf{0} \\

3 & object & sofa
& \cellcolor{agitaTableGreen}Reasonable
& \cellcolor{agitaSupportBlue}\textbf{4} \\

4 & attribute & patterned sofa
& \cellcolor{agitaTableGreen}Reasonable
& \cellcolor{agitaSupportBlue}\textbf{4} \\

5 & object & pillows
& \cellcolor{agitaTableGreen}Reasonable
& \cellcolor{agitaSupportBlue}\textbf{4} \\

6 & count & several pillows
& \cellcolor{agitaTableGreen}Reasonable
& \cellcolor{agitaSupportBlue}\textbf{4} \\

7 & object & ball
& \cellcolor{agitaTableGreen}Reasonable
& \cellcolor{agitaSupportBlue}\textbf{0} \\

8 & attribute & red ball
& \cellcolor{agitaTableGreen}Reasonable
& \cellcolor{agitaSupportBlue}\textbf{0} \\

9 & object & front paws
& \cellcolor{agitaTableGreen}Reasonable
& \cellcolor{agitaSupportBlue}{\scriptsize Cannot judge} \\

10 & action & dog lies on a patterned sofa
& \cellcolor{agitaTableGreen}Reasonable
& \cellcolor{agitaSupportBlue}\textbf{4} \\

11 & spatial & dog beside several pillows
& \cellcolor{agitaTableGreen}Reasonable
& \cellcolor{agitaSupportBlue}\textbf{4} \\

12 & spatial & red ball near the dog's front paws
& \cellcolor{agitaTableGreen}Reasonable
& \cellcolor{agitaSupportBlue}\textbf{0} \\

\end{tabularx}

\arrayrulecolor{black}
\normalsize

\vspace{0.55em}


\begin{agitainputbox}
\footnotesize
\textbf{\color{agitaBlue}Annotation note.}
\textbf{p9 (front paws):}
The dog's front paws are sufficiently obscured that this standalone part
cannot be confidently judged from visible evidence.
\normalsize
\end{agitainputbox}

\vspace{0.45em}


\begin{agitarationalebox}
\textbf{Overall rationale}

\vspace{0.2em}

The dog, patterned sofa, pillows, and lying position are well supported,
but the dog is brown rather than black and no red ball is visible.
\end{agitarationalebox}

\end{agitaexamplebox}

\vspace*{\fill}

\clearpage


\clearpage

\phantomsection
\label{example:worked-example-3}

\vspace*{\fill}

\begin{agitaexamplebox}


\noindent
\begin{minipage}[c]{0.70\linewidth}
    {\large\bfseries
    \color{agitaBlue}
    Example 3:
    \color{black}
    Weak alignment}
\end{minipage}%
\hfill
\begin{minipage}[c]{0.27\linewidth}
    \raggedleft
    \agitabiasbadge{Weak alignment}
\end{minipage}

\vspace{0.6em}


\noindent
\begin{minipage}[t]{0.40\linewidth}
    \vspace{0pt}
    \includegraphics[
        width=\linewidth,
        height=0.20\textheight,
        keepaspectratio
    ]{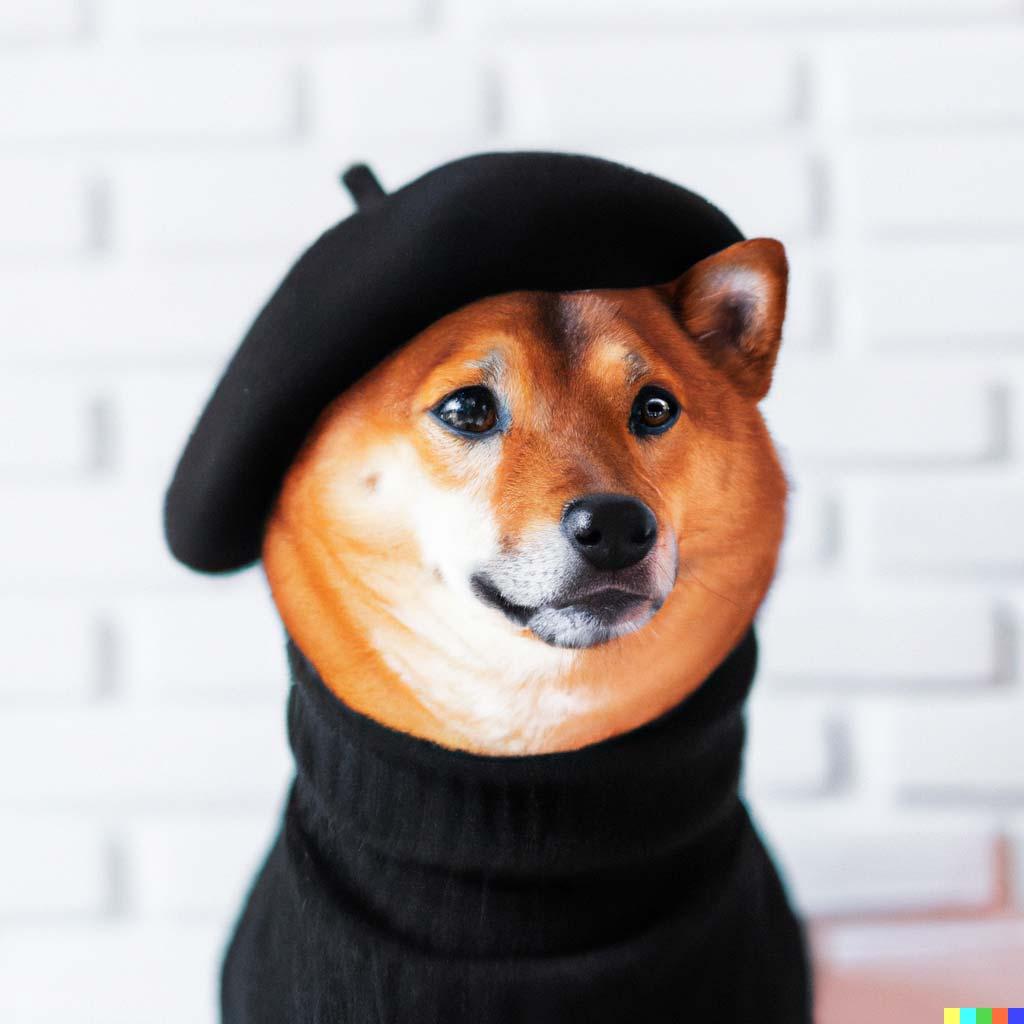}
\end{minipage}%
\hfill
\begin{minipage}[t]{0.57\linewidth}
    \vspace{0pt}

    \begin{agitainputbox}
        {\bfseries\color{agitaBlue}Input text}

        \vspace{0.3em}

        A Shiba Inu wearing a red beret and a blue turtleneck
        sits in front of a green wooden wall.
    \end{agitainputbox}

    \vspace{0.45em}

    \begin{agitascorebox}
        \noindent
        \begin{minipage}[c]{0.68\linewidth}
            {\bfseries\color{agitaGreen}
            Global alignment score}
        \end{minipage}%
        \begin{minipage}[c]{0.30\linewidth}
            \raggedleft
            {\Large\bfseries\color{agitaGreen}2 / 5}
        \end{minipage}
    \end{agitascorebox}

\end{minipage}

\vspace{0.55em}


\footnotesize
\setlength{\tabcolsep}{4.5pt}
\renewcommand{\arraystretch}{1.05}

\begin{tabularx}{\linewidth}{
    >{\centering\arraybackslash}p{0.035\linewidth}
    >{\raggedright\arraybackslash}p{0.115\linewidth}
    >{\raggedright\arraybackslash}X
    >{\centering\arraybackslash}p{0.185\linewidth}
    >{\centering\arraybackslash}p{0.145\linewidth}
}

\arrayrulecolor{agitaBlueBorder}

\rowcolor{agitaTableBlue}
\textbf{\#} &
\textbf{Type} &
\textbf{Atomic claim} &
\textbf{Decomp.\ quality} &
\textbf{Atomic support} \\
\hline

1 & object & Shiba Inu
& \cellcolor{agitaTableGreen}Reasonable
& \cellcolor{agitaSupportBlue}\textbf{4} \\

2 & object & beret
& \cellcolor{agitaTableGreen}Reasonable
& \cellcolor{agitaSupportBlue}\textbf{4} \\

3 & attribute & red beret
& \cellcolor{agitaTableGreen}Reasonable
& \cellcolor{agitaSupportBlue}\textbf{0} \\

4 & object & turtleneck
& \cellcolor{agitaTableGreen}Reasonable
& \cellcolor{agitaSupportBlue}\textbf{4} \\

5 & attribute & blue turtleneck
& \cellcolor{agitaTableGreen}Reasonable
& \cellcolor{agitaSupportBlue}\textbf{0} \\

6 & object & wall
& \cellcolor{agitaTableGreen}Reasonable
& \cellcolor{agitaSupportBlue}\textbf{4} \\

7 & attribute & green wall
& \cellcolor{agitaTableGreen}Reasonable
& \cellcolor{agitaSupportBlue}\textbf{0} \\

8 & attribute & wooden wall
& \cellcolor{agitaTableGreen}Reasonable
& \cellcolor{agitaSupportBlue}\textbf{0} \\

9 & relation & Shiba Inu wearing a red beret
& \cellcolor{agitaTableGreen}Reasonable
& \cellcolor{agitaSupportBlue}\textbf{2} \\

10 & relation & Shiba Inu wearing a blue turtleneck
& \cellcolor{agitaTableGreen}Reasonable
& \cellcolor{agitaSupportBlue}\textbf{2} \\

11 & action & Shiba Inu sits
& \cellcolor{agitaTableGreen}Reasonable
& \cellcolor{agitaSupportBlue}{\scriptsize Cannot judge} \\

12 & spatial & Shiba Inu in front of a wall
& \cellcolor{agitaTableGreen}Reasonable
& \cellcolor{agitaSupportBlue}\textbf{4} \\

\end{tabularx}

\arrayrulecolor{black}
\normalsize

\vspace{0.45em}


\begin{agitainputbox}
\scriptsize
\textbf{\color{agitaBlue}Annotation notes.}

\textbf{p9 (Shiba Inu wearing a red beret):}
Partially supported: the dog IS wearing a beret, but the required red
attribute is wrong. A composite relation that is partly correct but contains
an important mismatch scores 2, not 0.

\smallskip
\textbf{p10 (Shiba Inu wearing a blue turtleneck):}
Partially supported for the same reason: the turtleneck is present but black
rather than blue.

\smallskip
\textbf{p11 (Shiba Inu sits):}
The lower body is outside the frame, so the sitting posture cannot be judged
confidently.
\normalsize
\end{agitainputbox}

\vspace{0.4em}


\begin{agitarationalebox}
\footnotesize
\textbf{Overall rationale}

\vspace{0.2em}

The Shiba Inu, beret, turtleneck, and wall are present, but the beret is
black rather than red, the turtleneck is black rather than blue, and the
wall is light-colored brick rather than green wood. The lower body is
outside the frame, so the sitting posture cannot be judged confidently.
\normalsize
\end{agitarationalebox}

\end{agitaexamplebox}

\vspace*{\fill}

\clearpage


\clearpage

\phantomsection
\label{example:worked-example-4}

\vspace*{\fill}

\begin{agitaexamplebox}


\noindent
\begin{minipage}[c]{0.68\linewidth}
    {\large\bfseries
    \color{agitaBlue}
    Example 4:
    \color{black}
    Decomposition quality}
\end{minipage}%
\hfill
\begin{minipage}[c]{0.29\linewidth}
    \raggedleft
    \agitabiasbadge{Decomposition quality}
\end{minipage}

\vspace{0.4em}


\begin{agitarationalebox}
\footnotesize
\textbf{Teaching point.}
Decomposition quality and image support are independent: this example pairs
every decomposition label with a high support score, and a Reasonable
decomposition with a support score of 0.
\normalsize
\end{agitarationalebox}

\vspace{0.4em}


\noindent
\begin{minipage}[t]{0.37\linewidth}
    \vspace{0pt}
    \includegraphics[
        width=\linewidth,
        height=0.16\textheight,
        keepaspectratio
    ]{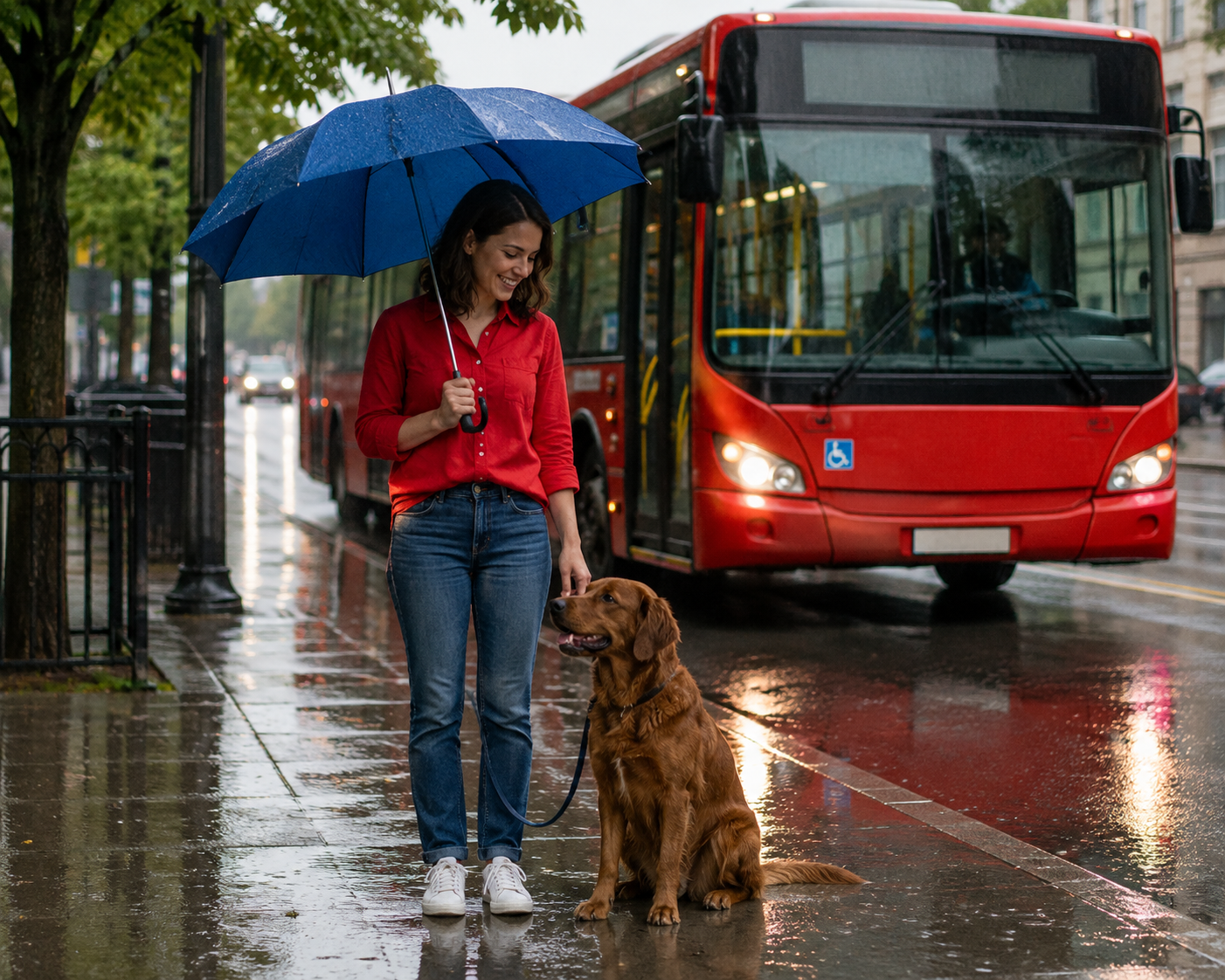}
\end{minipage}%
\hfill
\begin{minipage}[t]{0.60\linewidth}
    \vspace{0pt}

    \begin{agitainputbox}
        \footnotesize
        {\bfseries\color{agitaBlue}Input text}

        \vspace{0.2em}

        A woman in a red shirt is holding a blue umbrella beside a brown dog,
        while standing on a wet sidewalk in front of a yellow bus.
        \normalsize
    \end{agitainputbox}

    \vspace{0.35em}

    \begin{agitascorebox}
        \noindent
        \begin{minipage}[c]{0.68\linewidth}
            {\bfseries\color{agitaGreen}
            Global alignment score}
        \end{minipage}%
        \begin{minipage}[c]{0.30\linewidth}
            \raggedleft
            {\Large\bfseries\color{agitaGreen}4 / 5}
        \end{minipage}
    \end{agitascorebox}

\end{minipage}

\vspace{0.4em}


\scriptsize
\setlength{\tabcolsep}{3.7pt}
\renewcommand{\arraystretch}{1.00}

\begin{tabularx}{\linewidth}{
    >{\centering\arraybackslash}p{0.03\linewidth}
    >{\raggedright\arraybackslash}p{0.10\linewidth}
    >{\raggedright\arraybackslash}X
    >{\centering\arraybackslash}p{0.18\linewidth}
    >{\centering\arraybackslash}p{0.13\linewidth}
}

\arrayrulecolor{agitaBlueBorder}

\rowcolor{agitaTableBlue}
\textbf{\#} &
\textbf{Type} &
\textbf{Atomic claim} &
\textbf{Decomp.\ quality} &
\textbf{Atomic support} \\
\hline

1 & object & woman
& \cellcolor{agitaTableGreen}Reasonable
& \cellcolor{agitaSupportBlue}\textbf{4} \\

2 & attribute & red shirt
& \cellcolor{agitaTableGreen}Reasonable
& \cellcolor{agitaSupportBlue}\textbf{4} \\

3 & attribute & blue umbrella
& \cellcolor{agitaTableGreen}Reasonable
& \cellcolor{agitaSupportBlue}\textbf{4} \\

4 & attribute & brown dog
& \cellcolor{agitaTableGreen}Reasonable
& \cellcolor{agitaSupportBlue}\textbf{4} \\

5 & attribute & wet sidewalk
& \cellcolor{agitaTableGreen}Reasonable
& \cellcolor{agitaSupportBlue}\textbf{4} \\

6 & object & bus
& \cellcolor{agitaTableGreen}Reasonable
& \cellcolor{agitaSupportBlue}\textbf{4} \\

7 & attribute & yellow bus
& \cellcolor{agitaTableGreen}Reasonable
& \cellcolor{agitaSupportBlue}\textbf{0} \\

8 & relation & woman holding a blue umbrella
& \cellcolor{agitaTableGreen}Reasonable
& \cellcolor{agitaSupportBlue}\textbf{4} \\

9 & spatial & woman beside a brown dog
& \cellcolor{agitaTableGreen}Reasonable
& \cellcolor{agitaSupportBlue}\textbf{4} \\

10 & relation & woman standing on a wet sidewalk
& \cellcolor{agitaTableGreen}Reasonable
& \cellcolor{agitaSupportBlue}\textbf{4} \\

11 & spatial & woman in front of a yellow bus
& \cellcolor{agitaTableGreen}Reasonable
& \cellcolor{agitaSupportBlue}\textbf{2} \\

12 & relation &
woman in a red shirt holding a blue umbrella beside a brown dog
& \cellcolor{agitaGoldLight}\textbf{Needs split}
& \cellcolor{agitaSupportBlue}\textbf{4} \\

13 & action & holding
& \cellcolor{agitaGoldLight}\textbf{Needs merge}
& \cellcolor{agitaSupportBlue}\textbf{4} \\

14 & attribute & shirt that is red
& \cellcolor{agitaGoldLight}\textbf{Redundant}
& \cellcolor{agitaSupportBlue}\textbf{4} \\

15 & action & smiling woman
& \cellcolor{agitaGoldLight}\textbf{Not entailed}
& \cellcolor{agitaSupportBlue}\textbf{4} \\

\end{tabularx}

\arrayrulecolor{black}
\normalsize

\vspace{0.35em}


\begin{agitainputbox}
\scriptsize
\textbf{\color{agitaBlue}Annotation notes.}

\textbf{p7 (yellow bus):}
The text correctly decomposes to `yellow bus', but the visible bus is red.

\smallskip
\textbf{p11 (woman in front of a yellow bus):}
The woman is in front of a bus, but the required yellow attribute is
incorrect.

\smallskip
\textbf{p12 (woman in a red shirt holding a blue umbrella beside a brown dog):}
It combines several independently checkable claims, although the complete
statement is visually supported.

\smallskip
\textbf{p13 (holding):}
The action is visually present, but as a decomposition it loses who is
holding what.

\smallskip
\textbf{p14 (shirt that is red):}
It repeats the visual claim already represented by `red shirt'.

\smallskip
\textbf{p15 (smiling woman):}
The original text never says the woman is smiling, but the image clearly
shows her smiling.
\normalsize
\end{agitainputbox}

\vspace{0.35em}


\begin{agitarationalebox}
\scriptsize
\textbf{Overall rationale}

\vspace{0.15em}

The woman, red shirt, blue umbrella, brown dog, wet sidewalk, and described
actions and spatial relations are clearly supported. The meaningful mismatch
is the bus colour: the text specifies a yellow bus, while the image shows a
red bus.
\normalsize
\end{agitarationalebox}

\end{agitaexamplebox}

\vspace*{\fill}

\clearpage

%% file: figures/app_explanation_annotation_guideline.tex

\definecolor{guideblue}{HTML}{294E87}
\definecolor{guidefill}{HTML}{F5F7FB}
\definecolor{guideline}{HTML}{D8DEE9}
\definecolor{guidetext}{HTML}{273142}

\clearpage

\begin{figure*}[p]
\centering

\begin{tikzpicture}

\node[
    draw=guideblue,
    line width=0.8pt,
    rounded corners=2pt,
    fill=guidefill,
    inner sep=10pt,
    text width=0.96\textwidth,
    text=guidetext
] (guidebox) {%

\scriptsize
\setlength{\parindent}{0pt}
\setlength{\parskip}{2.5pt}


\begin{minipage}[t]{0.485\linewidth}

{\color{guideblue}\bfseries
A $\boldsymbol{\cdot}$ WHAT YOU ARE DOING}

\vspace{2pt}

For every sample you see an \textbf{image} and the \textbf{caption} being
evaluated. Four different automatic systems each wrote a short explanation
of whether the image supports the caption. The explanations are shown only as
\textbf{Explanation A / B / C / D}; the system identity is hidden and the order
varies across samples.

\vspace{2pt}

\begin{enumerate}
    \setlength{\itemsep}{1pt}
    \setlength{\parskip}{0pt}
    \setlength{\topsep}{1pt}

    \item \textbf{Task A} --- rate each of the four explanations, one at a
    time. No highlighted regions are shown yet.

    \item \textbf{Task B} --- after all four explanations are rated, a
    second panel opens only for explanation(s) whose system provides a
    highlighted image region for a claim. Depending on the sample, this may
    apply to none, one, or multiple explanations. Up to three highlighted
    regions are rated. Once Task B is revealed, the corresponding
    \textbf{Task A answers are locked and can no longer be changed}, keeping
    the two judgments independent.
\end{enumerate}

\textit{Not every explanation has a Task B. Some systems do not produce
localized image regions; this is expected and should not be flagged.}

\vspace{5pt}
\hrule
\vspace{5pt}

{\color{guideblue}\bfseries
B $\boldsymbol{\cdot}$ TASK A --- RATING ONE EXPLANATION}

\vspace{3pt}

\renewcommand{\arraystretch}{1.16}
\setlength{\tabcolsep}{3pt}

\begin{tabularx}{\linewidth}{
    @{}
    >{\bfseries\raggedright\arraybackslash}p{0.29\linewidth}
    X
    @{}
}
\hline
\multicolumn{2}{@{}l@{}}{\rule{0pt}{1.5ex}}\\[-7pt]

Diagnostic Correctness &
``How accurately does the explanation identify the relevant matches or
mismatches between the image and caption?''

\textit{1 Completely incorrect $\cdot$ 2 Mostly incorrect $\cdot$
3 Partially correct $\cdot$ 4 Mostly correct $\cdot$
5 Completely correct}
\\[3pt]
\hline

Visual Grounding &
``How well are the statements in the explanation supported by what is visibly
present in the image and stated in the caption?''

\textit{1 Not supported $\cdot$ 2 Mostly unsupported $\cdot$
3 Partially supported $\cdot$ 4 Mostly supported $\cdot$
5 Fully supported}
\\[3pt]
\hline

Clarity &
``How easy is the explanation to understand, without unnecessary or confusing
information?''

\textit{1 Very unclear $\cdot$ 2 Somewhat unclear $\cdot$
3 Neither clear nor unclear $\cdot$ 4 Clear $\cdot$ 5 Very clear}
\\[3pt]
\hline

Diagnostic Usefulness &
\textbf{Primary question.}
``How useful is the explanation for understanding why the image and caption
align or fail to align?''

\textit{1 Not useful $\cdot$ 2 Slightly useful $\cdot$
3 Moderately useful $\cdot$ 4 Very useful $\cdot$
5 Extremely useful}
\\[3pt]
\hline

Unsupported Content &
``Does this explanation contain any specific factual statement that is
contradicted by, or unsupported by, the image/caption?''

\textit{Yes / No / Unsure}
\\[2pt]
\hline
\end{tabularx}

\vspace{3pt}

\textit{Judge every explanation against the image yourself.}

\end{minipage}
\hfill
\begin{minipage}[t]{0.485\linewidth}

{\color{guideblue}\bfseries
C $\boldsymbol{\cdot}$ TASK B --- RATING HIGHLIGHTED EVIDENCE}

\vspace{2pt}

For each qualifying explanation, annotators evaluate up to three claims,
each accompanied by its own highlighted image region and short claim text.

\vspace{3pt}

\renewcommand{\arraystretch}{1.16}
\setlength{\tabcolsep}{3pt}

\begin{tabularx}{\linewidth}{
    @{}
    >{\bfseries\raggedright\arraybackslash}p{0.31\linewidth}
    X
    @{}
}
\hline
\multicolumn{2}{@{}l@{}}{\rule{0pt}{1.5ex}}\\[-7pt]

Evidence Localization Accuracy &
``How accurately do the highlighted region(s) identify the visual evidence
relevant to this claim?''

\textit{1 Completely incorrect $\cdot$ 2 Mostly incorrect $\cdot$
3 Partially correct $\cdot$ 4 Mostly correct $\cdot$
5 Completely correct}
\\[3pt]
\hline

Evidence Sufficiency &
``How sufficient are the highlighted region(s) for deciding whether this claim
is supported by the image?''

\textit{1 Completely insufficient $\cdot$ 2 Mostly insufficient $\cdot$
3 Partially sufficient $\cdot$ 4 Mostly sufficient $\cdot$
5 Fully sufficient}
\\[2pt]
\hline
\end{tabularx}

\vspace{6pt}
\hrule
\vspace{5pt}

{\color{guideblue}\bfseries
D $\boldsymbol{\cdot}$ SKIP FOR NOW / TECHNICAL ISSUE}

\vspace{3pt}

\begin{tabularx}{\linewidth}{
    @{}
    >{\bfseries\raggedright\arraybackslash}p{0.27\linewidth}
    X
    @{}
}
Skip for now &
Postpone the whole sample. It remains in the annotator's assignment and is
marked as unfinished on the dashboard until completed.
\\[4pt]

Technical issue &
Use only when something is actually broken, such as an image failing to load,
garbled text, or the annotation interface malfunctioning.
\\
\end{tabularx}

\vspace{6pt}
\hrule
\vspace{5pt}

{\color{guideblue}\bfseries
E $\boldsymbol{\cdot}$ NAVIGATING, SAVING AND EXPORTING}

\vspace{2pt}

\begin{itemize}
    \setlength{\itemsep}{2pt}
    \setlength{\parskip}{0pt}
    \setlength{\topsep}{1pt}

    \item Work \textbf{saves automatically}; annotators may close the tab and
    return later.

    \item Completed samples can be \textbf{reopened}, but after Task B has
    been revealed for a sample, its Task A answers remain locked. Task B
    ratings may still be revised.

    \item Annotators are asked to use the \textbf{same browser and device}
    where practical and not clear site data before exporting.

    \item A \textbf{backup} can be downloaded from the menu at any time.
    After completing all assigned samples, annotators use
    \textbf{Export final annotations} to produce the final annotation file.
\end{itemize}

\vspace{6pt}

{\color{guideblue}\bfseries Annotation order}

\vspace{2pt}

\centering
\textbf{Image + Caption}
$\;\longrightarrow\;$
\textbf{Task A: Explanation}
$\;\longrightarrow\;$
\textbf{Lock Task A}
$\;\longrightarrow\;$
\textbf{Task B: Evidence}

\end{minipage}

}; 

\node[
    anchor=west,
    fill=guideblue,
    text=white,
    rounded corners=1.5pt,
    font=\bfseries\footnotesize,
    inner xsep=7pt,
    inner ysep=3.5pt
] at ([xshift=10pt]guidebox.north west)
{\textsc{Human Explanation \& Evidence Annotation Guidelines}};

\end{tikzpicture}

\caption{\textbf{Human explanation and evidence annotation guidelines.}
Complete instructions provided to annotators.}
\label{fig:explanation_annotation_guidelines}

\end{figure*}

\clearpage